\pdfoutput=1

\documentclass[11pt]{article}

\usepackage[preprint]{acl}

\usepackage{times}
\usepackage{latexsym}

\usepackage[T1]{fontenc}

\usepackage[utf8]{inputenc}

\usepackage{microtype}

\usepackage{inconsolata}

\usepackage{graphicx}
\usepackage{booktabs}
\usepackage{multirow}
\usepackage{mathtools}
\usepackage{array}
\usepackage{amssymb}
\usepackage{subcaption}
\usepackage{algorithm}
\usepackage{algorithmic}
\usepackage{cleveref}

\crefname{section}{Sec.}{Secs.}
\Crefname{section}{Section}{Sections}
\Crefname{table}{Table}{Tables}
\crefname{table}{Tab.}{Tabs.}
\Crefname{figure}{Figure}{Figures}
\crefname{figure}{Fig.}{Figs.}
\Crefname{equation}{Equation}{Equations}
\crefname{equation}{Eq.}{Eqs.}

\title{Temporal Scaling Law for Large Language Models}

\author{Yizhe Xiong$^{1,2}$\quad Xiansheng Chen$^{3}$\quad Xin Ye$^{3}$\quad Hui Chen$^{2}$\thanks{Corresponding Authors.}\quad Zijia Lin$^{3*}$\quad \textbf{Haoran Lian}$^{4}$\\
\textbf{Zhenpeng Su}$^{5}$\quad \textbf{Wei Huang}$^6$\quad \textbf{Jianwei Niu}$^{4}$\quad \textbf{Jungong Han}$^7$\quad \textbf{Guiguang Ding}$^{1,2*}$\\
$^{1}$School of Software, Tsinghua University\quad $^{2}$BNRist, Tsinghua University\\
$^{3}$Kuaishou Technology\quad $^{4}$Beihang University\quad $^{5}$University of Chinese Academy of Sciences\\
$^{6}$School of Computer Science, BUPT\quad $^{7}$ Department of Automation, Tsinghua University \\
{\tt\small \{xiongyizhe2001,  jichenhui2012\}@gmail.com linzijia07@tsinghua.org.cn dinggg@tsinghua.edu.cn}
}

\begin{document}
\maketitle
\begin{abstract}
Recently, Large Language Models (LLMs) have been widely adopted in a wide range of tasks, leading to increasing attention towards the research on how scaling LLMs affects their performance. 
Existing works, termed Scaling Laws, have discovered that the final test loss of LLMs scales as power-laws with model size, computational budget, and dataset size. 
However, the temporal change of the test loss of an LLM throughout its pretraining process remains unexplored, though it is valuable in many aspects, such as selecting better hyperparameters \textit{directly} on the target LLM.
In this paper, we propose the novel concept of Temporal Scaling Law, studying how the test loss of an LLM evolves as the training steps scale up.
In contrast to modeling the test loss as a whole in a coarse-grained manner, we break it down and dive into the fine-grained test loss of each token position, and further develop a dynamic hyperbolic-law.
Afterwards, we derive the much more precise temporal scaling law by studying the temporal patterns of the parameters in the dynamic hyperbolic-law.
Results on both in-distribution (ID) and out-of-distribution (OOD) validation datasets demonstrate that our temporal scaling law accurately predicts the test loss of LLMs across training steps.
Our temporal scaling law has broad practical applications. 
First, it enables direct and efficient hyperparameter selection on the target LLM, such as data mixture proportions.
Secondly, viewing the LLM pretraining dynamics from the token position granularity provides some insights to enhance the understanding of LLM pretraining.
\end{abstract}

\section{Introduction}
\label{sec:1_intro}
Large Language Model (LLM) marks a paradigm shift in the scope of natural language processing, demonstrating unprecedented capabilities in accomplishing complicated tasks
of natural language understanding and generation~\cite{DBLP:conf/naacl/DevlinCLT19,radford2018improving}. 
A cornerstone of this remarkable progress lies in the scalability of the Transformer architecture~\cite{DBLP:conf/nips/VaswaniSPUJGKP17}, which has facilitated the development of increasingly large models. 
Moreover, the accessibility of large-scale training is significantly enhanced by the abundance of data gathered from the Internet, which enables the construction of giant training datasets to improve model performance~\cite{radford2019language}. 
Due to the scalability of both the model size and training data scale, LLMs with billions of parameters~\cite{DBLP:conf/nips/BrownMRSKDNSSAA20} and even trillions of parameters~\cite{DBLP:journals/jmlr/FedusZS22} are widely proposed and applied in various tasks.

The scalability of LLMs has been extensively studied in terms of variables like model size, computational budget, and dataset size \cite{DBLP:journals/corr/abs-2001-08361}.
Prior works in this domain, which are commonly referred to as ``scaling laws''\cite{DBLP:journals/corr/abs-2001-08361,DBLP:journals/corr/abs-2010-14701}, have proposed empirical principles to characterize the power-law (i.e., exponential patterns) among those variables. 
Those scaling laws demonstrate that the final test loss\footnote{Following \cite{DBLP:journals/corr/abs-2001-08361}, "test loss" refers to the loss on unseen data, which can include both the test and validation sets.} of an LLM improves exponentially with scalable factors, i.e.,  model size, computational budget, and dataset size. 
Recently, similar explorations have expanded to multi-modal training~\cite{DBLP:conf/cvpr/ChertiBWWIGSSJ23,DBLP:conf/icml/AghajanyanYCHHZ23} and transfer learning~\cite{DBLP:journals/corr/abs-2102-01293},
enabling researchers to predict the performance of LLMs in different scenarios.

\begin{table*}
\centering
\scalebox{0.8}{\begin{tabular}{l|cc}
\hline
& Temporal Scaling Law & Prior Scaling Laws \\
\hline
Predict objective & \textbf{Evolution} of test loss during pretraining & \textbf{Final} test loss after pretraining \\
\hline
\multirow{2}{*}{Fitting \& Predicting manners} & Fit with light early pretrain on \textbf{target LLM}, & Fit with fully pretrained \textbf{smaller models}, \\
& predict on the \textbf{target LLM} & predict on the \textbf{target LLM}  \\
\hline
Granularity & Loss on each token position & Loss as a whole \\
\hline
\multirow{2}{*}{Application} & Select hyperparameters directly on target LLMs, & Decide a target LLM size \&  \\
& Provide insight w.r.t. pretraining dynamics, etc. & data scale after given a budget \\
\hline
\end{tabular}}
\caption{Comparison between our temporal scaling law and prior scaling laws.}
\label{tab:teaser_diff}
\end{table*}

Prior works of scaling laws typically predict the final test loss of \textit{fully-trained LLMs with a given computational budget} after completing pretraining \textit{with mostly fixed hyperparameters} (like data mixture proportions, weight decay, etc.)~\cite{DBLP:journals/corr/abs-2001-08361,DBLP:journals/corr/abs-2010-14701,DBLP:conf/cvpr/ChertiBWWIGSSJ23,DBLP:conf/icml/AghajanyanYCHHZ23}.
However, variations in training hyperparameters also significantly influence the final test loss in a complicated way \cite{DBLP:journals/corr/abs-2203-03466,DBLP:journals/corr/abs-2305-10429}. That underscores the need for a more fine-grained scaling law, which is to \textit{predict the \underline{temporal} evolution of the test loss as training steps scale up under different training hyperparameters} on a fixed model size. 
Such a temporal scaling law is supposed to be complementary to prior works, because it further enables the direct identification of better training hyperparameters on the target LLM after the model size and data scale are determined based on previous research.
Furthermore, such a temporal scaling law can also provide a theoretical basis for studying more training dynamics of LLMs. 
Despite broad applications, prior works have not explored practical scaling laws from the temporal perspective.

In this paper, we propose the novel concept of \textbf{Temporal Scaling Law} for LLMs. 
Specifically, we first try to model the test loss as a whole in a coarse-grained manner but find it to be not precise enough. Then we further break the test loss into the losses for tokens in different positions. By carefully investigating multiple functions, we discovered that utilizing dynamic hyperbolic-law accurately portrays the pattern of losses on different token positions across different training steps.
We further examine the evolution of the curve parameters for the dynamic hyperbolic-law and identify their temporal patterns. We term this phenomenon as the temporal scaling law, i.e., \textit{how the test loss of an LLM evolves as the training steps scale up}. 
Our temporal scaling law accurately predicts the subsequent test losses using the data from an early training period. 
Prediction results on both the in-distribution (ID) and the out-of-distribution (OOD) validation datasets show that our methodology significantly improves over baseline approaches. 

Our temporal scaling law has broad practical applications for LLM pretraining. In this paper, we provide two use cases as examples:

(a) Our temporal scaling law presents a novel and practical approach to selecting the hyperparameters \textit{directly on the target to-be-pretrained LLM}.
Taking the data mixture proportion as an example, current works generally tune data proportions on a small-scale model and directly apply the tuned weights to pretraining the much larger target LLM~\cite{DBLP:journals/corr/abs-2305-10429}.
However, optimal hyperparameters for a small-scale model probably are not the optimal ones for the much larger target LLM \cite{DBLP:journals/corr/abs-2203-03466}. With the accurate loss prediction from our proposed temporal scaling law, we can select the target LLM's hyperparameters directly by choosing the best one which can achieve the lowest-predicted test loss after training the target LLM with a small-scale of training data.

(b) Our temporal scaling law reveals some learning dynamics of LLMs at the token granularity.
Specifically, we theoretically and experimentally discovered that the loss decrease rate for tokens on different positions remains uniform after an early training period.
Through experiments on various position-based weighting strategies, we verify the effectiveness of the default pretraining practice, in which no weighting strategies on different token positions are applied, though they are imbalanced in terms of learning difficulty.


\textbf{Differences from Existing Scaling Laws.}
We summarize the differences between our temporal scaling law and prior scaling laws \cite{DBLP:journals/corr/abs-2001-08361,DBLP:journals/corr/abs-2010-14701,DBLP:conf/cvpr/ChertiBWWIGSSJ23,DBLP:conf/icml/AghajanyanYCHHZ23} in \cref{tab:teaser_diff}. Our temporal scaling law primarily focuses on the evolution of test loss during the pretraining process, while prior scaling laws model the relations of the final test loss and the computational budget (generally decided by model size and training data scale). Our temporal scaling law enables a direct selection of training hyperparameters on a target LLM after the model size and training data scale are decided by prior scaling laws.
\textit{We underscore that our temporal scaling law is a generic pattern that decoder-based LLMs follow during the pretraining process.}

Overall, our contributions are threefold:
\begin{itemize}
    \item We propose the Temporal Scaling Law, modeling the evolution of test loss during LLM pretraining.
    \item Based on our proposed temporal scaling law, we provide a method for precisely predicting the subsequent test losses after deriving the parameters of the temporal scaling law from an early training period. 
    \item We present two scenarios to illustrate the broad practical applications of our temporal scaling law. For hyperparameter tuning, our temporal scaling law enables the direct selection of better hyperparameter values based on predicted LLM performance on the target LLM. For learning dynamics, our temporal scaling law provides theoretical and experimental verifications for the default pretraining practice that puts no weight on each token loss at different positions. 
\end{itemize}

\section{Temporal Scaling Law}
\label{sec:3_temporal}

\subsection{Preliminaries}
\textbf{Pretraining Language Models.}
We mainly discuss the generation process of decoder-based generative language models. A generative language model is trained via next-token prediction. During pretraining, the text corpus is commonly segmented into token sequences. The token sequences are then fed to the model for calculating the prediction loss on each token position $i$. Formally, given a training text sequence $T_n$ consisting of $n$ tokens, i.e., $T_n=[t_1,\cdots,t_{i-1},t_i,\cdots,t_n]$, when predicting the token $t_i$, the language model takes the previous tokens $T_{i-1}=[t_1,\cdots,t_{i-1}]$ as input, and generates the probability distribution $\mathbf{p}_i$ for the token $t_i$. The loss $\mathcal{L}_i$ w.r.t. the token $t_i$ is commonly calculated via the cross-entropy loss function.
In decoder-based transformers~\cite{DBLP:conf/nips/VaswaniSPUJGKP17}, a look-ahead mask is applied in the multi-head attention (MHA) module to ensure that each token can only attend to previous tokens, preserving the causal order of the generated sequence.
Then, all tokens' $\mathbf{p}_i$ can be calculated in a single forward pass in parallel during pretraining.
The loss $\mathcal{L}_i$ on each token's $\mathbf{p}_i$ is then averaged as the loss for sequence $T_n$:
\begin{equation}
    \mathcal{L}_{T_n} = \frac{1}{n}\sum^{n}_{i=1}\mathcal{L}_i
\end{equation}

\subsection{Experiment Setup for Deriving Temporal Scaling Law}
\label{sec:3_2_exp_setup}
We use experiment results to derive the temporal scaling law and conduct predictions after deriving parameters of the temporal scaling law from an early training period. 
Please refer to \cref{sec:more_exp_setting} for more experiment details.

\textbf{Train Dataset.}
We train LLMs on the Pile~\cite{DBLP:journals/corr/abs-2101-00027}, consisting of 22 English text domains. 
For all experiments, we tokenize it using the LLaMA tokenizer~\cite{DBLP:journals/corr/abs-2302-13971} with a 32k vocabulary. 
We apply the domain weights in \cite{DBLP:journals/corr/abs-2310-19531} for LLM pretraining.

\textbf{Validation Dataset.}
We apply two validation datasets for test loss calculation. For the in-distribution (ID) dataset, we randomly sample 800 sequences (1024 consecutive tokens for each) for each text domain from the validation set in the Pile. For the out-of-distribution (OOD) dataset, we simply adopt the validation split of the RealNews-Like domain in the C4 dataset~\cite{DBLP:journals/jmlr/RaffelSRLNMZLL20}, as it is claimed to be ``distinct from the Pile''~\cite{DBLP:journals/corr/abs-2101-00027}. 
We refer to both validation datasets as ID-Val and OOD-Val, respectively. 
When calculating test loss, we forward all sequences in the validation dataset and average the results.
It is worth noting that calculating test loss is equivalent to calculating the perplexity (PPL) for the validation dataset since PPL can be directly acquired via test loss:
\begin{equation}
    \text{PPL}(T)=\exp\{-\frac{1}{n}\sum_i^n\log p_\theta(t_i|T_{i-1})\}
\end{equation}

\begin{table}
\centering
\scalebox{0.8}{\begin{tabular}{l|cccc}
\hline
 & \multicolumn{2}{c}{ID-Val} & \multicolumn{2}{c}{OOD-Val} \\
 Model & Power-law & Ours & Power-law & Ours \\
\hline
\textit{9.8M} & 0.6319 & \textbf{0.9858} & 0.7052 & \textbf{0.9795} \\
\textit{58M} & 0.7606 & \textbf{0.9961} & 0.7684 & \textbf{0.9954} \\
\hline
\end{tabular}}
\caption{$R^2$ results for fitting the test loss with the power-law (treat test loss as a whole) and our temporal scaling law (token-level granularity). \textbf{Bold} represents the best result. }
\label{tab:temporal_fit}
\vspace{-0.1in}
\end{table}

\textbf{Models.}
We train two LLaMA-structure \cite{DBLP:journals/corr/abs-2302-13971} generative language models with 9.8M and 58M parameters to illustrate our temporal scaling law. The architectures of both are identical to the 14M and 70M parameter models in~\cite{DBLP:conf/icml/BidermanSABOHKP23}. The differences between model parameters can be ascribed to the different vocabulary sizes and different activation functions (i.e., SwiGLU v.s. GeLU). When utilizing the temporal scaling law for predictions and further applications, we adopt the architectures of the 410M and 1.0B parameter models in~\cite{DBLP:conf/icml/BidermanSABOHKP23}, which leads to 468M and 1.2B models, respectively.
To further \textit{scale up and generalize} the validation of temporal scaling law, we also pre-train: (1) a 6.7B model with the same architecture of LLaMA-7B, (2) a 468M GPT-NeoX model, and (3) an 2B MoE model. Please refer to \cref{sec:append_structure} in the Appendix for training and fitting results on (2) and (3).

\textbf{Training.}
Following LLaMA~\cite{DBLP:journals/corr/abs-2302-13971}, we apply the AdamW optimizer~\cite{DBLP:conf/iclr/LoshchilovH19} with a learning rate of 3$e$-4. Following most open-source LLMs~\cite{DBLP:journals/corr/abs-2302-13971,DBLP:conf/icml/BidermanSABOHKP23}, we use the cosine learning rate decay schedule~\cite{DBLP:conf/iclr/LoshchilovH17} for all experiments to guarantee the broad applicability of our work. 
All models are pretrained with 400B tokens and $1k$ total warmup steps. 

\textbf{Metric for Fitting Model Evaluation.}
We propose the temporal scaling law to fit the test loss during pretraining.
Following the common practical choice \cite{CHENG2014137}, we choose the coefficient of determination ($R^2$) to evaluate the quality of the fit. $R^2$ demonstrates the proportion of the variability in the ground truth that the proposed fit could explain. Formally, we have:
\begin{equation}
    R^2=1-\frac{\sum_i(y_i-f_i)^2}{\sum_i(y_i-\overline{y})^2},
\end{equation}
where $y_i$ is the ground-truth test loss w.r.t. the $i$-th step, and $\overline{y}$ is the average of $y_i$. $f_i$ is the fit test loss w.r.t. the $i$-th step.
$R^2\in(-\infty,1]$ and a perfect fit has $R^2=1$. A larger $R^2$ indicates a better fit.

\subsection{Temporal Scaling Law}
\label{sec:3_3_temporal_law}
\begin{figure*}
  \centering
  \begin{tabular}{ccc}
  \begin{subfigure}{0.31\linewidth}
    \includegraphics[width=1.0\linewidth, trim={0cm 0cm 1.3cm 0cm}, clip]{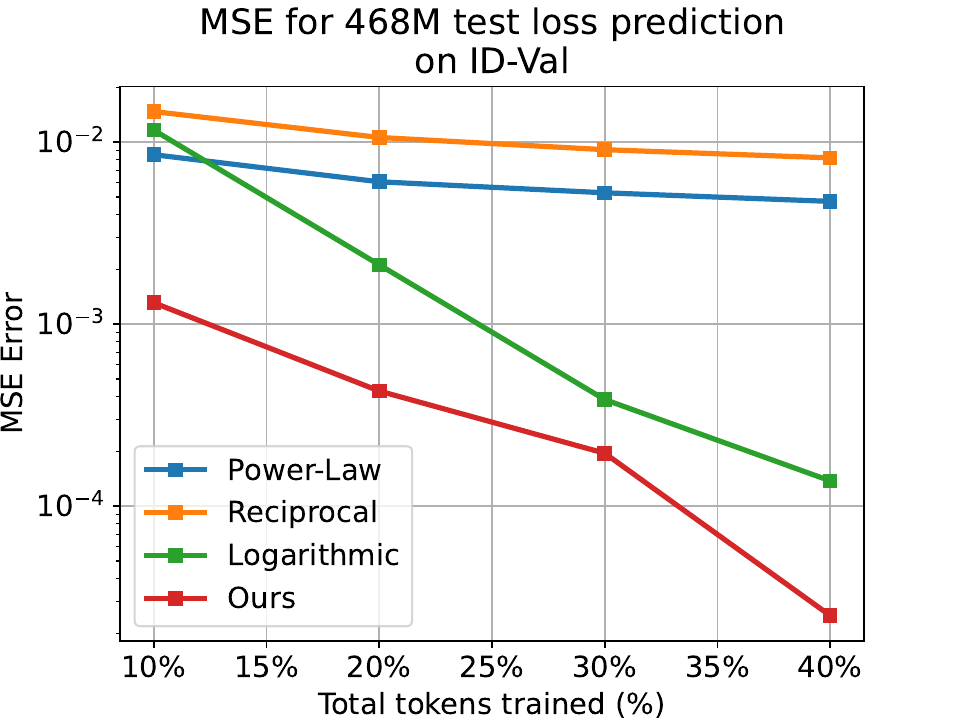}
  \end{subfigure} &
  \begin{subfigure}{0.31\linewidth}
    \includegraphics[width=1.0\linewidth, trim={0cm 0cm 1.3cm 0cm}, clip]{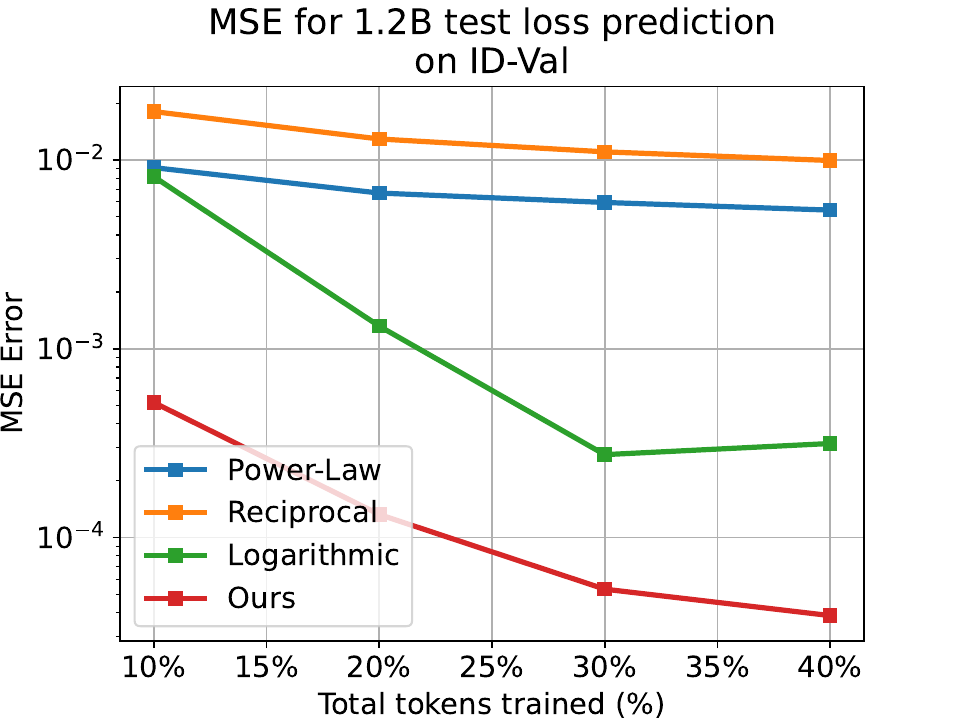}
  \end{subfigure} &
  \begin{subfigure}{0.31\linewidth}
    \includegraphics[width=1.0\linewidth, trim={0cm 0cm 1.3cm 0cm}, clip]{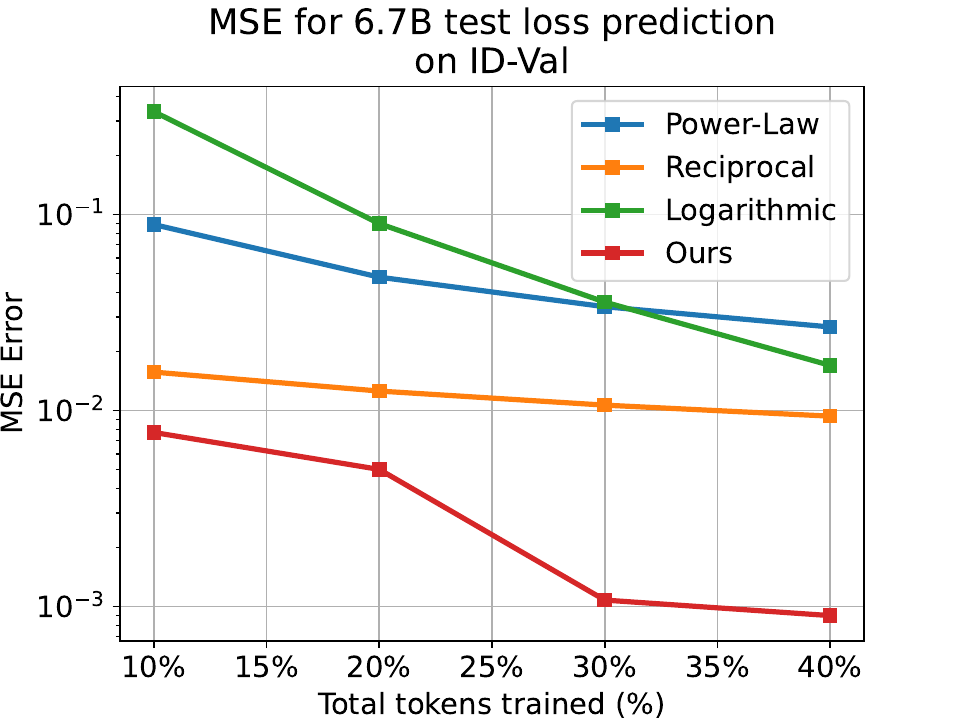}
  \end{subfigure} \\


  \end{tabular}
  \caption{MSE results for predicting the subsequent test loss via the temporal scaling law after completing different proportions of the training process. For the 468M, 1.2B, and 6.7B models, $N_{sep}$ lies approximately at 40\% of the entire training schedule, and all presented results require predicting through both fitting stages (i.e., $N<N_{sep}$ and $N\ge N_{sep}$). Note that the y-axis representing the MSE error is in \underline{\textbf{log scale}}.}
  \label{fig:fit_exp_result}
  \vspace{-0.2in}
\end{figure*}

The original scaling law~\cite{DBLP:journals/corr/abs-2001-08361} has validated that the \textit{final} test loss for training with different model sizes and data sizes follows a power-law. 
To validate the effectiveness of the power-law for the \textit{temporal evolution} of test loss, we pretrain 9.8M and 70M models following \cref{sec:3_2_exp_setup} and fit the power-law function ($\mathcal{L}=(p_1x)^{p_2}+p_3$, $\{p_i\}$ are fitting parameters) to the test loss evolution curve (i.e., treating test loss as a whole to fit in this case) using non-linear least squares. 
As shown in \cref{tab:temporal_fit}, though somehow effective, applying the power-law is not precise enough to portray the temporal evolution for the test loss (follow-up experiments in \cref{fig:fit_exp_result} also demonstrate that directly using the power-law in predicting the future test loss results in larger errors).  Such results motivate us to find a better method to understand the test loss evolution during LLM pretraining.

\textbf{The Dynamic Hyperbolic-Law of Loss on Different Token Positions.} 
To fit and predict the test loss more precisely, we propose to break it down and investigate the temporal scaling law from a finer granularity and look into the inherent patterns of the loss on every token position. Language models are statistical models trained to model the probabilistic process of next-token prediction. Formally, for a partial sequence of $n$ tokens $T_n=(t_1,t_2,\cdots,t_n)$, the probability of the following token being $t_{n+1}$ given the preceding tokens is denoted as $P(t_{n+1}|T_n)$. Intuitively and statistically, in a consecutive sequence:
\begin{equation}
\label{eq:context}
    \mathbb{E}[P(t_{n+1}|T_n)]\gtrsim \mathbb{E}[P(t_{n}|T_{n-1})],
\end{equation}
since $t_{n+1}$ has a longer context than $t_n$.
To illustrate this pattern, we plot the results of test loss for tokens in different positions, as validated on our ID-Val, using both our 9.8M and 58M models after training for 100B, 200B, 300B, and 400B tokens.
We show the results for 400B tokens and leave the rest in the Appendix.
As shown in ~\cref{fig:token_loss}, both models, although varying in scales, follow the same pattern that test loss for tokens with shorter context is commonly higher than that for tokens with longer context. For both models, the loss values for tokens near the end of the sequence seem to converge at a certain value. We find that this trend can be well fit with a hyperbolic relation\footnote{Note that this hyperbolic relation is actually a special case of the power-law \cite{DBLP:journals/corr/abs-2001-08361,team2024gemini} with an exponent of $-1$. Compared to directly applying the power-law, the hyperbolic relation leads to better fitting results. We compare and elaborate this in \cref{sec:theoretical_hyperbolic} of the Appendix.}:
\begin{equation}
\label{eq:hyperbolic}
    \mathcal{L}_i=\frac{a_0}{1+a_1i}+a_2,
\end{equation}
where $\mathcal{L}_i$ is the loss on token position $i(1\le i\le n)$, and $a_0$, $a_1$, and $a_2$ are fitting parameters solved by non-linear least squares.
The fit for data from our 9.8M and 58M model are presented in~\cref{fig:token_loss}. By applying this fit to test loss results of all checkpoints across the training phase, we find that over 99\% of them can be fit with such hyperbolic relation with $R^2>0.95$, well demonstrating its generality. We term this phenomenon as the dynamic hyperbolic-law. See \cref{sec:theoretical_hyperbolic} for more insights.

We also tested Kaplan's power-law \cite{DBLP:journals/corr/abs-2001-08361} and other functions that conform to \cref{eq:context} to fit the curve, but the results are suboptimal. For example, using the power-law yields $R^2<0.75$ for over 95\% checkpoints.
The conclusion still holds on the OOD-Val and on larger-scale models. See detailed results in \cref{sec:more_fitting}.

\begin{figure}[t]
  \centering
  \begin{subfigure}{0.49\columnwidth}
    \centering
      \includegraphics[width=1.0\linewidth]{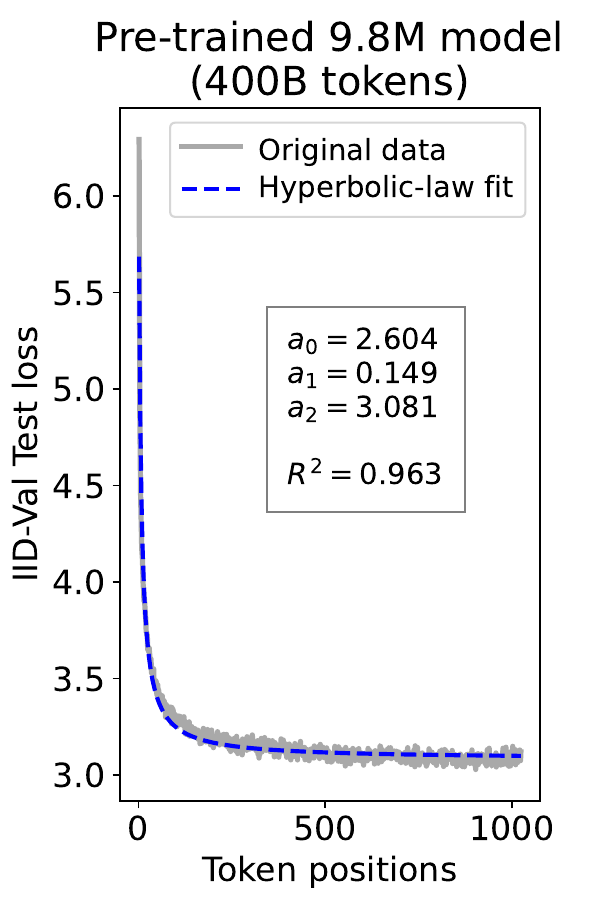}
  \end{subfigure}
  \begin{subfigure}{0.49\columnwidth}
    \centering
      \includegraphics[width=1.0\linewidth]{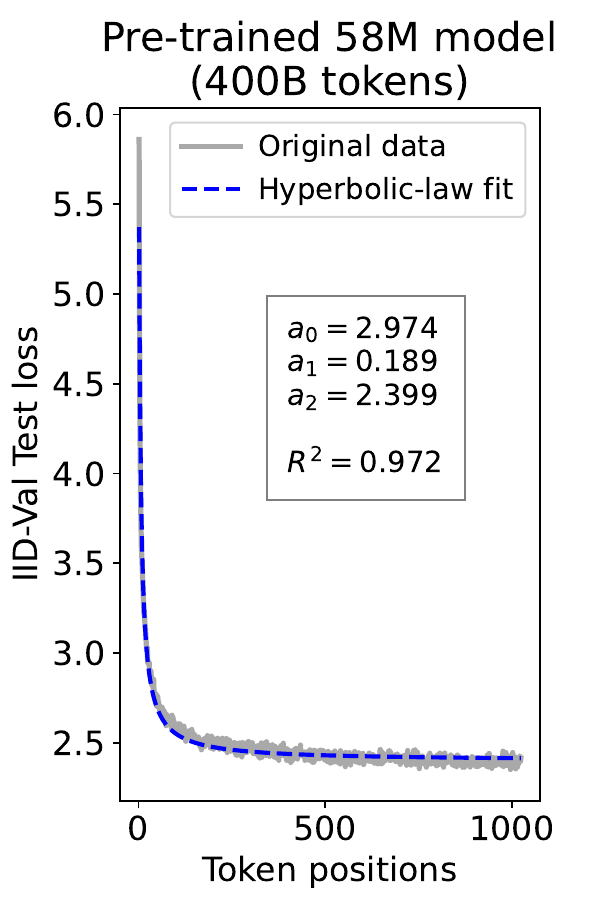}
  \end{subfigure}
   \caption{ Actual loss and fitting curve across different token positions for the 9.8M and 58M models on ID-Val. The loss follows a dynamic hyperbolic-law.}
   \label{fig:token_loss}
\end{figure}

\begin{figure*}
  \centering
  \begin{subfigure}{0.32\linewidth}
    \includegraphics[width=1.0\linewidth, trim={0cm 0cm 1.3cm 0cm}, clip]{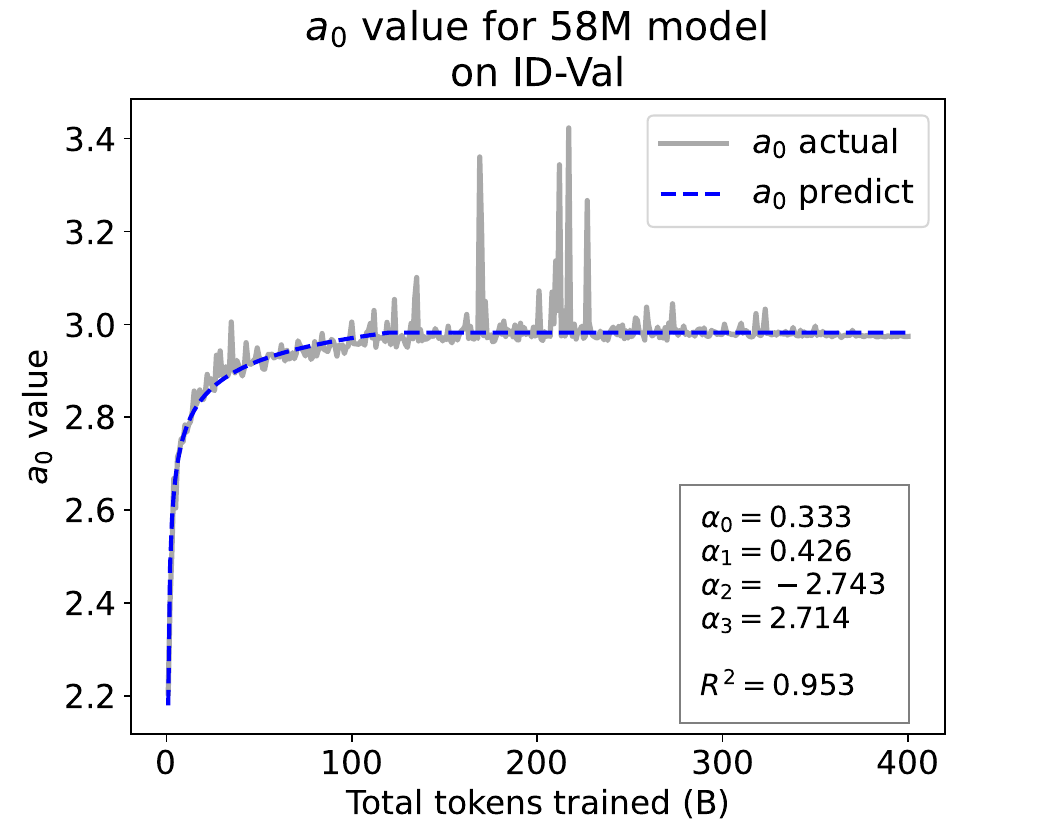}
  \end{subfigure}
  \begin{subfigure}{0.32\linewidth}
    \includegraphics[width=1.0\linewidth, trim={0cm 0cm 1.3cm 0cm}, clip]{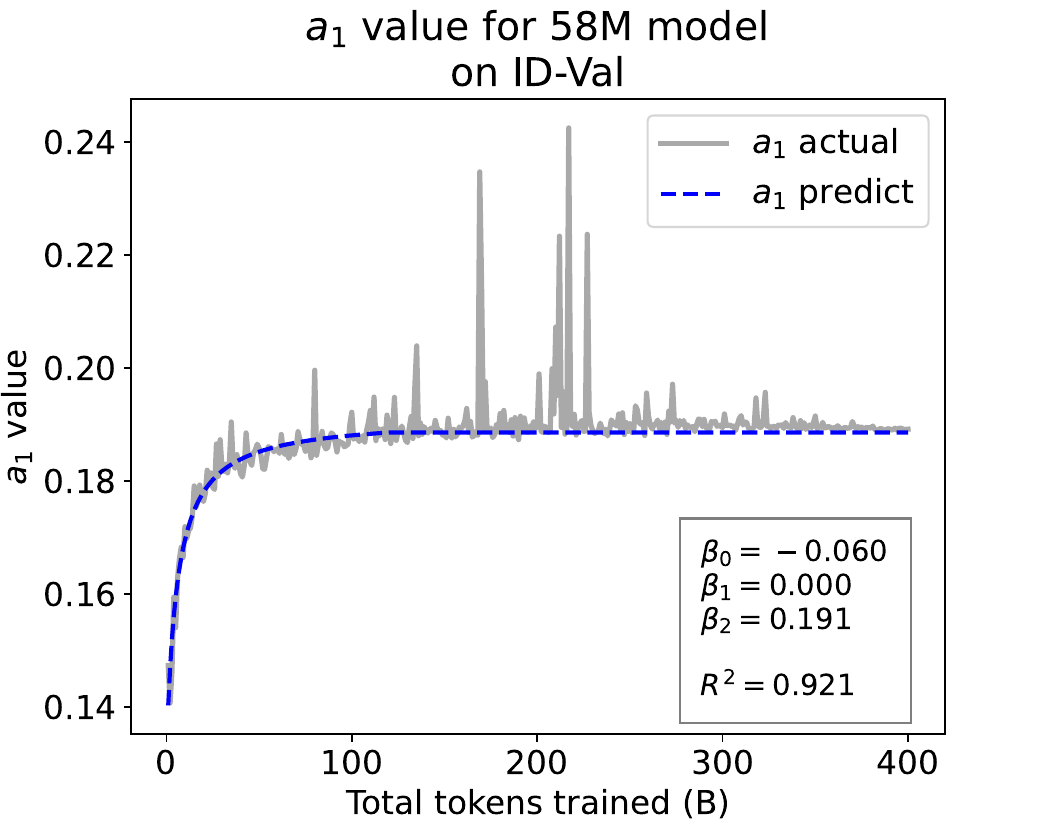}
  \end{subfigure}
  \begin{subfigure}{0.32\linewidth}
    \includegraphics[width=1.0\linewidth, trim={0cm 0cm 1.3cm 0cm}, clip]{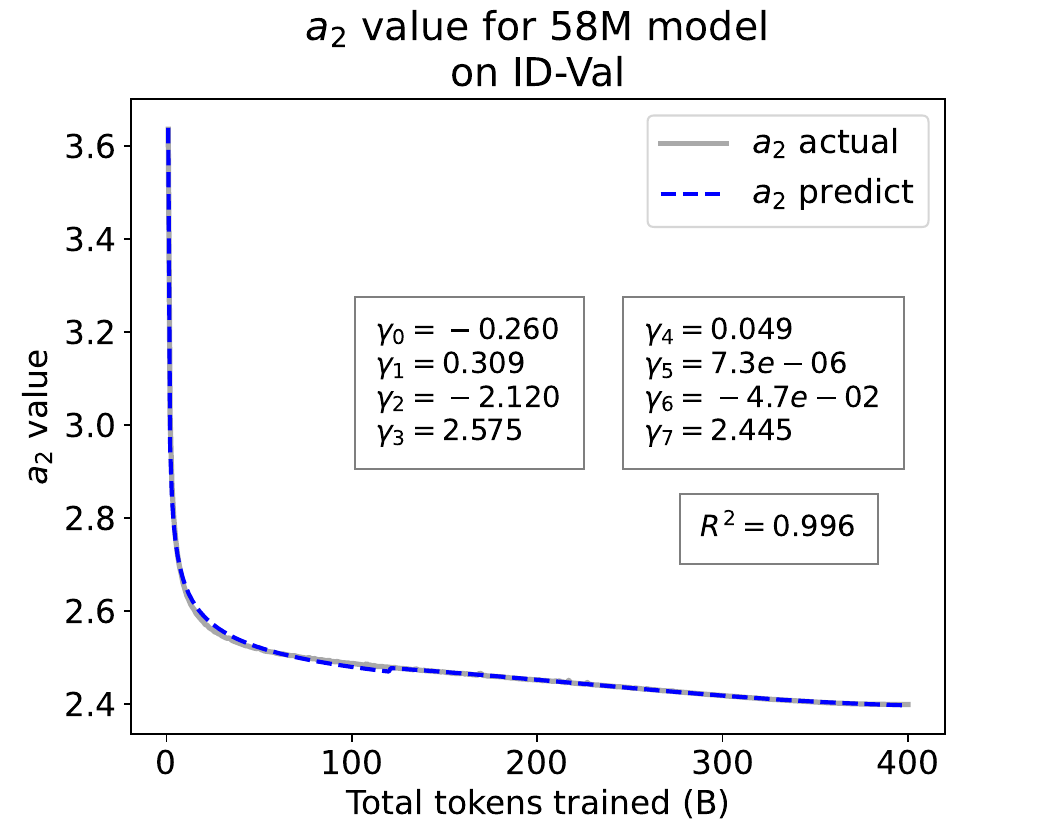}
  \end{subfigure}
  \caption{Fitting curve for $a_0$, $a_1$, and $a_2$ of the 58M model on ID-Val. $R^2$ for $a_0$ and $a_1$ is affected by fluctuations.}
  \label{fig:temporal_law}
\end{figure*}

\textbf{From the Dynamic Hyperbolic-law to the Overall Temporal Pattern.} 
In the dynamic hyperbolic-law, $a_2$ is the converging factor denoting the converged value of loss on each tail token as context length increases. $a_0$ denotes the loss gap between the first token and the last token, and $a_1$ is the scaling factor along sequence length.
To analyze how the derived values of $a_0$, $a_1$, and $a_2$ vary as training steps increase, we collect the derived value tuples of $(a_0,a_1,a_2)$ for all evaluated small model checkpoints during pretraining: $[(a_0^{N_1},a_1^{N_1},a_2^{N_1}),\cdots,(a_0^{N_{tot}},a_1^{N_{tot}},a_2^{N_{tot}})]$, where $N_i$ is the number of trained tokens at the $i$-th checkpoint, and $N_{tot}$ is the total number of tokens used for pretraining. Then we plot the collected $(a_0,a_1,a_2)$ values on ID-Val w.r.t. the number of tokens trained (denoted as $N$) for the 58M model in~\cref{fig:temporal_law}.  
It can be seen that the evolution of $a_0$, $a_1$, and $a_2$ presents an evident pattern that can be fit by common functions. Therefore, it is possible for us to firstly fit the temporal evolution of $a_0$, $a_1$, and $a_2$, and then fill them into \cref{eq:hyperbolic} to depict the temporal evolution of the overall test loss.

First, for $a_0$ and $a_1$, we use a series of common functions to fit their temporal evolution, and eventually choose the following functions that yield the lowest fitting errors:
\begin{equation}
\begin{aligned}
    a_0^{N}&=\alpha_0\log(\alpha_1\log(N)+\alpha_2)+\alpha_3, \\
    a_1^{N}&=\frac{\beta_0}{1+\beta_1N}+\beta_2,
\end{aligned}
\label{eq:a0a1_before_sep}
\end{equation}
where the $\{\alpha_i\}$ and $\{\beta_i\}$ fitting parameters are solved by non-linear least squares.
Observed from the original data in \cref{fig:temporal_law}, we find that the value of $a_0$ and $a_1$ generally converges after a period of training, but suffer from fluctuations after converging due to the uncertainty of the first prediction position in a sequence, which has no previous context to make prediction. To mitigate the fluctuations, we define a separation point ${N}_{sep}$:
\begin{equation}
\begin{aligned}
    {N}_{sep}\coloneq\min N,\ \ s.t.\ \ \nabla a_0^N < \epsilon,\ \ \nabla a_1^N < \epsilon.
\end{aligned}
\label{eq:sep_point}
\end{equation}
And we manually stabilize $a_0$ and $a_1$ after ${N}_{sep}$:
\begin{equation}
    a_0^{N}=a_0^{{N}_{sep}},a_1^{N}=a_1^{{N}_{sep}},(N\ge{N}_{sep}).
\label{eq:a0a1_post_sep}
\end{equation}
Here we empirically set $\epsilon=10^{-4}/\text{1B tokens}$.


For parameter $a_2$, we observe that ${N}_{sep}$ is also a separation point as it marks the shift of its decreasing pattern. Before ${N}_{sep}$, we apply the same function form of the loss gap factor $a_0$ as in \cref{eq:a0a1_before_sep}. After ${N}_{sep}$, we find that its temporal pattern holds strong correlations with the cosine learning rate scheduler, and in turn apply a cosine relation. Similar to \cref{eq:a0a1_before_sep}, we search the common functions and choose the best fitting function among them for $a_2$: 
\begin{equation}
a_2^{N}=
\begin{cases}
\gamma_0\log(\gamma_1\log(N)&\!\!\!\!\!\!+\gamma_2) + \gamma_3,\\ &\!\!\!\!\!\!\!(N<{N}_{sep}) \\
\gamma_4\cos(\gamma_5N+\gamma_6&\!\!\!\!\!\!) + \gamma_7,\\ &\!\!\!\!\!\!\!(N\ge{N}_{sep})
\end{cases}
\label{eq:a2}
\end{equation}
where $\{\gamma_i\}$ are fitting parameters solved by non-linear least squares.
Interestingly, from the fit for $a_2$, we find that $\gamma_5\approx\frac{\pi}{{N}_{tot}}$, indicating that the training schedule resembles half of a cosine period, consistent with the cosine scheduler. It further validates our speculation that $a_2$ has a strong correlation with the cosine learning rate decay.

We plot the fit of $a_0$, $a_1$, and $a_2$ in \cref{fig:temporal_law}. Our fit captures the primary patterns of parameter evolution and ignores the insignificant fluctuations. The conclusion holds on the OOD-Val, and for larger models and models with other structures. See detailed results in \cref{sec:more_fitting,sec:append_structure}.

\begin{table}
\centering
\scalebox{0.85}{\begin{tabular}{l|cc}
\hline
 Model Scale & LAMBADA & Wikitext \\
\hline
\textit{9.8M} & 0.9887 & 0.9809 \\
\textit{58M} & 0.9805 & 0.9899 \\
\textit{468M} & 0.9831 & 0.9920 \\
\textit{1.2B} & 0.9859 & 0.9930 \\
\textit{6.7B} & 0.9872 & 0.9927 \\
\hline
\end{tabular}}
\caption{$R^2$ results for fitting the test loss with our temporal scaling law on the LAMBADA and the Wikitext validation datasets.}
\label{tab:temporal_fit_more_ood}
\end{table}

After acquiring $a_0^N$, $a_1^N$, and $a_2^N$, we could in turn measure the pattern of the total test loss $\mathcal{L}_N$ by averaging the loss on each token:
\begin{equation}
    \mathcal{L}^{N}=\frac{1}{n}\sum_{i=1}^n\frac{a_0^N}{1+a_1^N\cdot i}+a_2^N.
\label{eq:aggregate}
\end{equation}
As listed in~\cref{tab:temporal_fit}, the fit for all ground-truth test losses during the whole training process achieves $R^2>0.99$ across different settings, demonstrating the validity of our temporal scaling law.
Following \cite{DBLP:journals/corr/abs-2101-00027}, We also validate the fitting results on the LAMBADA \cite{DBLP:conf/acl/PapernoKLPBPBBF16} and the WikiText \cite{DBLP:conf/iclr/MerityX0S17} datasets.
Specifically, we use the validation and the test splits for the LAMBADA dataset, and use the wikitext-2-v1 domain for the WikiText dataset.
As shown in \cref{tab:temporal_fit_more_ood}, all results achieve $R^2>0.98$, well demonstrating the effectiveness of our temporal scaling law.
Please refer to \cref{sec:more_fitting} for more fitting results. 

\textbf{Summary.} Overall, we summarize our temporal scaling law as follows:
\begin{itemize}
    \item For LLM pretraining, in each training step, the loss for tokens in different positions follows a dynamic hyperbolic-law, in which the values of parameters $(a_0, a_1, a_2)$ will change as training steps increase.
    \item The temporal patterns (i.e., values at different training steps) of dynamic hyperbolic-law parameters (i.e., $a_0$, $a_1$, and $a_2$) can be captured individually with their corresponding coefficients (i.e., $\{\alpha_i\}$, $\{\beta_i\}$, and $\{\gamma_i\}$) in the fitting functions of fixed mathematical forms. 
    \item The temporal pattern of the overall test loss is derived by averaging losses at all token positions acquired by the dynamic hyperbolic-law, i.e., filling Eq. (6-9) into Eq. (10).
\end{itemize}


\subsection{Test Loss Prediction}
\label{sec:test_loss_pred}
The primary value of scaling laws lies in their capability of predicting training outcomes~\cite{DBLP:journals/corr/abs-2001-08361}. From the temporal perspective, accurate predictions of the upcoming training periods can assist in a direct selection of hyperparameters based on the target LLM's performance, early stopping, etc. In this section, we derive practical methods from our temporal scaling law to accurately predict the overall loss in the upcoming training periods, using data from an early training period to derive parameters of the temporal scaling law, during which only $N_{train}$ tokens are trained.


To predict the overall loss, we first predict the temporal trajectory of $a_0^N$, $a_1^N$, and $a_2^N$. We could fit for $a_0^N$ and $a_1^N$ with the functions in \cref{eq:a0a1_before_sep}:
\begin{equation}
\begin{aligned}
    \Tilde{a}_0^{N}&=\Tilde{\alpha}_0\log(\Tilde{\alpha}_1\log(N)+\Tilde{\alpha}_2)+\Tilde{\alpha}_3, \\
    \Tilde{a}_1^{N}&=\frac{\Tilde{\beta}_0}{1+\Tilde{\beta}_1N}+\Tilde{\beta}_2,
\end{aligned}
\label{eq:a0a1_predict_before_sep}
\end{equation}
where $\{\Tilde{\alpha}_i\}$ and $\{\Tilde{\beta}_i\}$ are fit using all collected $\{\mathcal{L}_i^N\}(N<N_{train})$ with non-linear least squares. 
Following \cref{eq:sep_point}, we could estimate the separation point as $\Tilde{N}_{sep}$. Based on the relationship of $N_{train}$ and $\Tilde{N}_{sep}$, the test loss is predicted as:

\textbf{Situation \#1} (In most occasions): $N_{train}\le\Tilde{N}_{sep}$, i.e., in an early training stage where the separate point is not reached yet.
For $\Tilde{a}_0^N$ and $\Tilde{a}_1^N$, we apply the fit \cref{eq:a0a1_predict_before_sep} for $N\le\Tilde{N}_{sep}$. For $\Tilde{a}_2^N$, we first fit the first segment in \cref{eq:a2}:
\begin{equation}
    \Tilde{a}_2^{N}=\Tilde{\gamma}_0\log(\Tilde{\gamma}_1\log(N)+\Tilde{\gamma}_2)+\Tilde{\gamma}_3,
\label{eq:a2_pred_fit_func}
\end{equation}
where $\{\gamma_i\}$ are fit using non-linear least squares.
Due to the absence of test loss data for $N_{train}>\Tilde{N}_{sep}$, we could not acquire the value for $\Tilde{a}_0^N$, $\Tilde{a}_1^N$, and $\Tilde{a}_2^N$ at that period directly.
To estimate the patterns after the separation point, we propose general boundary conditions for estimation:
\begin{equation}
\begin{aligned}
    {\frac{d\Tilde{a}_i^N}{dN}}_{N\rightarrow\Tilde{N}_{sep}^+}&={\frac{d\Tilde{a}_i^N}{dN}}_{N\rightarrow\Tilde{N}_{sep}^-}, \\
    \Tilde{a}_i^{N\rightarrow\Tilde{N}_{sep}^+}&=\Tilde{a}_i^{N\rightarrow\Tilde{N}_{sep}^-}.
\end{aligned}
\label{eq:general_estimate_principle}
\end{equation}
According to the definition in \cref{eq:sep_point}, we simply stabilize the values for $\Tilde{a}_0^N$ and $\Tilde{a}_1^N$ as $\Tilde{a}_0^{\Tilde{N}_{sep}}$ and $\Tilde{a}_1^{\Tilde{N}_{sep}}$ correspondingly for $N>\Tilde{N}_{sep}$, fulfilling the derivative condition in \cref{eq:general_estimate_principle} approximately.
For $\Tilde{a}_2^N$
at $N\ge \Tilde{N}_{sep}$, to eliminate the number of unknown variables, we directly adopt the discovery from fitting \cref{eq:a2} that $a_2$ resembles half of a cosine period, and directly set $\Tilde{\gamma}_5=\frac{\pi}{N_{tot}}$. We set $\Tilde{\gamma}_6=-\frac{\pi}{N_{tot}}N_w$, where $N_w$ is the number of tokens used for warmup training, to ensure that $N=N_w$ marks the start point of a cosine period. We in turn solve for the rest parameters $\Tilde{\gamma}_4$ and $\Tilde{\gamma}_7$ in the following function based on \cref{eq:general_estimate_principle}:
\begin{equation}
    \Tilde{a}_2^{N}=\Tilde{\gamma}_4\cos[\frac{\pi}{N_{tot}}(N-N_w)]+\Tilde{\gamma}_7.
\label{eq:a2_pred_after_sep}
\end{equation}

\textbf{Situation \#2} (Otherwise): $N_{train}>\Tilde{N}_{sep}$, i.e., in a middle stage where the separate point is already reached: 
Apart from steps in Situation \#1, we additionally calibrate $\Tilde{\gamma}_4$ and $\Tilde{\gamma}_7$ with the $N_{train}>\Tilde{N}_{sep}$ data by estimating $\epsilon_4$ and $\epsilon_7$:
\begin{equation}
    \Tilde{a}_2^{N}=(\Tilde{\gamma}_4+\epsilon_4)\cos[\frac{\pi}{N_{tot}}(N-N_w)]+(\Tilde{\gamma}_7+\epsilon_7).
\label{eq:a2_pred_calibrate}
\end{equation}

After acquiring the complete estimations of $\Tilde{a}_0^{N}$, $\Tilde{a}_1^{N}$, and $\Tilde{a}_2^{N}$, we could calculate the test loss prediction $\Tilde{\mathcal{L}}^N$ for each training step with \cref{eq:aggregate}.


\textbf{Experiments.}
Predicting test losses in upcoming training periods is a time series modeling problem.
Following standard practice \cite{DBLP:conf/nips/WuXWL21,DBLP:conf/nips/LiuWWL22}, we choose MSE for evaluation:
\begin{equation}
    \text{MSE}=\frac{1}{n}\sum_{i=1}^{n}(\mathcal{L}^N-\Tilde{\mathcal{L}}^N)^2.
\end{equation} 
Apart from the power-law, we also choose the reciprocal ($\mathcal{L}=\frac{a_0}{1+a_1 N}+a_2$) and the logarithmic ($\mathcal{L}=\log(a_1+a_2 N)+a_3$) relations as baselines that model the test loss as a whole.
We predict the subsequent test loss after deriving parameters of the temporal scaling law from, respectively, 10\%, 20\%, 30\%, and 40\% of the training process. 
Note that here we apply the larger models with 468M, 1.2B, and 6.7B parameters for validation.
As shown in~\cref{{fig:fit_exp_result}}, on different model scales and proportions of training, while the baselines hardly generate reliable results, our method always yields accurate results with $\text{MSE}<10^{-2}$. 
Our method yields far better $R^2$ as well. For example, when using 10\% data for 1.2B model on ID-Val, our method yields $R^2=0.87$, while the power-law, reciprocal, and logarithmic baselines yield $R^2=-1.26,-3.47,-1.02$, correspondingly.
Those prediction results provide strong evidence for the reliability of our temporal scaling law.
The conclusion holds on the OOD-Val. See \cref{sec:pred_ood} for detailed results.

\begin{table*}[!t]
\centering
\scalebox{0.7}{
\begin{tabular}
{ll|p{1.1cm}<{\centering}p{1.65cm}<{\centering}p{1.65cm}<{\centering}p{1.65cm}<{\centering}p{1.65cm}<{\centering}p{1.65cm}<{\centering}p{1.7cm}<{\centering}}
\toprule
&  & BoolQ & HellaSwag & OpenBookQA & PIQA & SIQA & StoryCloze & Winogrande \\
 \midrule
\multirow{2}{*}{\textit{468M}} & Small Model / 10B Test Loss & \textbf{58.70} & 45.38 & 31.25 & 68.60 & 43.39 & 64.78 & \textbf{54.93} \\
 & Our pipeline & 57.07 & \textbf{45.98} & \textbf{33.68} & \textbf{69.36} & \textbf{44.01} & \textbf{64.88} & 54.02 \\
\midrule
\multirow{2}{*}{\textit{1.2B}} & Small Model / 10B Test Loss & 56.25 & 54.13 & 34.25 & 72.20 & 42.11 & 69.05 & 59.83 \\
 & Our pipeline  & \textbf{58.70} & \textbf{54.51} & \textbf{34.90} & \textbf{72.36} & \textbf{46.73} & \textbf{69.11} & \textbf{60.32} \\
\bottomrule
\end{tabular}
}
\caption{Average benchmark performance (0/1/5-shot) of models pretrained on different data proportions. ``Small Model'' denotes the Top-1 on the smaller model. ``10B Test Loss'' denotes the Top-1 selected by \textit{real} test loss on ID-Val and OOD-Val validation sets after training 10B tokens. \textbf{Bold} represents the best result.}
\label{tab:data_proportion_benchmark}
\vspace{-0.15in}
\end{table*}

\begin{table}[!t]
\centering
\scalebox{0.7}{\begin{tabular}
{ll|cc}
\toprule
 & & ID-Val & OOD-Val \\
\midrule
\multirow{2}{*}{\textit{468M}} & Small Model / 10B Test Loss & 8.65 & 11.92 \\
& Our pipeline & \textbf{8.58} & \textbf{11.77} \\
\midrule
\multirow{2}{*}{\textit{1.2B}} & Small Model / 10B Test Loss & 7.55 & 10.29 \\
& Our pipeline & \textbf{7.50} & \textbf{10.17} \\
\bottomrule
\end{tabular}}
\caption{Pretrained model PPL results. ``Small Model'' denotes the Top-1 on the smaller model. ``10B Test Loss'' denotes the Top-1 selected by \textit{real} test loss after training 10B tokens. \textbf{Bold} is the best result.}
\label{tab:data_proportion_perplexity}
\vspace{-0.2in}
\end{table}

\section{Use Case \#1 of Temporal Scaling Law: Hyperparameter Selection}
\label{sec:4_hyperparameter}
Our temporal scaling law can be applied to enable a direct hyperparameter selection for LLMs.
With our temporal scaling law, for hyperparameters that are completely un-transferrable such as the weight decay \cite{DBLP:journals/corr/abs-2203-03466}, it is applicable to directly search them on the target large model from a candidate pool. As for hyperparameters that are "partially transferrable", e.g., data mixtures, small proxy models can help to identify a smaller range of effective candidates for the target large model, though not necessarily optimal. In such cases, our method is compatible with existing small-model-to-large-model hyperparameter searching methods, and here we could use a retrieval-rerank pipeline to achieve more precise results.


For example, data mixture proportions greatly affect the performance of LLMs~\cite{DBLP:journals/corr/abs-2302-13971,DBLP:journals/corr/abs-2307-09288,DBLP:journals/corr/abs-2305-10429}. 
Due to the huge computation cost of training with multiple data proportions directly on target LLMs, prior art simply applies the proportion values tuned on a much smaller model~\cite{DBLP:journals/corr/abs-2305-10429}. 
Using the temporal scaling law, we could accurately predict the final performance (in terms of test losses on validation sets) of different data proportions without completing the extensive pretraining.

Specifically, to select the best data mixture proportion for training a larger 468M/1.2B model,
in the retrieval stage, following the prior art \cite{DBLP:journals/corr/abs-2305-10429,feng2024maximize}, we use a small-scale model to locate a group of data mixture proportion candidates. 
In the rerank stage, we train all located data proportions on the target LLM for acceptable training tokens, and locate the best one by predicting their final losses with our temporal scaling law.

\textbf{Experiment Setup.}
In the retrieval stage, we conduct a grid search on a 58M model and locate the Top-5 candidates. 
In the rerank stage, we train each candidate for 10B tokens before making the test loss prediction with Eq.11-14 on validation datasets.
To validate the data mixture choices by temporal scaling law on the target 468M/1.2B models, we (1) primarily use the perplexity metric on the ID-Val and the OOD-Val, and also (2) follow \cite{DBLP:journals/corr/abs-2302-13971,DBLP:journals/corr/abs-2112-11446} and employ 7 popular common sense reasoning benchmarks, including BoolQ~\cite{DBLP:conf/naacl/ClarkLCK0T19}, HellaSwag~\cite{DBLP:conf/acl/ZellersHBFC19}, OpenBookQA~\cite{DBLP:conf/emnlp/MihaylovCKS18}, PIQA~\cite{DBLP:conf/aaai/BiskZLGC20}, SIQA~\cite{DBLP:journals/corr/abs-1904-09728}, StoryCloze~\cite{DBLP:journals/corr/MostafazadehCHP16}, and Winogrande~\cite{DBLP:journals/cacm/SakaguchiBBC21} for evaluating the final fully trained models. See the \cref{sec:more_case_1} for more experimental details.

\textbf{Experiment Result.}
We compare the data mixture proportion selected by our retrieval-rerank pipeline with the following baselines: (1) the Top-1 selected directly on the much smaller model and (2) the Top-1 selected by comparing the \textit{real} test losses on ID-Val and OOD-Val validation sets after training 10B tokens on 468M/1.2B model. We find that both (1) and (2) make the same choice from the Top-5 candidates, and our pipeline locates a different choice from them.
As shown in \cref{tab:data_proportion_perplexity}, our choice achieves lower perplexity across all settings after completing pretraining (i.e., 400B tokens).
Additionally, as shown in \cref{tab:data_proportion_benchmark}, our choice yields superior benchmark performance in 12 of 14 metrics.
Consequently, our approach of directly selecting hyperparameters on the target LLM is more effective and incurs an acceptable cost, which does not necessitates fully training the model.

\section{Use Case \#2 of Temporal Scaling Law: Revisit the Training Strategy}
\label{sec:5_token_position_loss}
The temporal scaling law has afforded the possibility to study LLM pretraining in a finer granularity. 
As noted in~\cref{sec:3_3_temporal_law}, tokens on different positions possess a fundamental bias in learning difficulty. Specifically, head tokens are generally harder to predict than tail tokens. However, as suggested by \cref{eq:a0a1_before_sep,eq:a0a1_post_sep}, $a_0$ and $a_1$ remain constant after the separation point. Therefore, for $N>N_{sep}$:
\begin{equation}
    \frac{\partial\mathcal{L}_i^N}{\partial N}=\frac{\partial a_2}{\partial N},
\label{eq:learning_bias}
\end{equation}
which is unrelated to token position $i$, suggesting that LLMs learn equally on different token positions.
To validate this observation, we trace the actual loss decrease observed during training along different token positions. It is observed that in a later training stage where the separation point conditions in \cref{eq:sep_point} are already fulfilled, test loss decrease among all token positions tends to be uniform across all settings (see \cref{fig:token_loss_diff} for results). Such a consistency between theoretical prediction and experimental observation strongly validates the reasonableness and effectiveness of our temporal scaling law.
Experiments also show that simply averaging losses on all token positions is an effective strategy for training LLMs.
Refer to \cref{sec:more_case_2} for more experimental settings and results.


\section{Conclusion}
Our research introduces the novel concept of Temporal Scaling Law for Large Language Models (LLMs), studying how the loss of an LLM evolves as the training steps scale up. 
We analyze the loss patterns on different token positions and discover that these patterns conform to a dynamic hyperbolic-law.
By studying the temporal evolution of parameters of the dynamic hyperbolic-law, we could properly fit and precisely predict the evolution of the LLM's test loss, marking a significant improvement over baseline methods. 
Such capability is crucial for numerous possible applications, like better hyperparameter selection directly on target LLM and deeper understanding pretraining dynamics of LLM, etc.

\section*{Limitations}
Our work termed Temporal Scaling Law, pioneers the modeling of LLM pre-training from a temporal perspective. 
Although our temporal scaling law exhibits significant potential for both LLM research and application, limitations still exist.
Our research primarily focuses on the pre-training stage. The temporal patterns in other scenarios, such as transfer learning, are not covered. 
We leave further investigations of the above topics to our future research.

\section*{Acknowledgment}
This work was supported by National Natural Science Foundation of China (Nos. 62525103, 62271281, 62441235) and Beijing Natural Science Foundation (Nos. L247026).

\bibliography{custom}

\appendix
\newpage

\section{Related Works}
\label{sec:2_related}


\subsection{Large Language Models}
Language models are statistical models designed to model the probabilistic correlation in natural language sequences~\cite{DBLP:journals/corr/abs-2302-13971}.
The introduction of the Transformer architecture~\cite{DBLP:conf/nips/VaswaniSPUJGKP17} led to the development of large language models. GPT-3~\cite{DBLP:conf/nips/BrownMRSKDNSSAA20} marks the beginning of the LLM era, as decoder-based generative models are capable of completing various tasks via in-context learning.
Further advancements in LLMs include PaLM~\cite{DBLP:journals/jmlr/ChowdheryNDBMRBCSGSSTMRBTSPRDHPBAI23}, Pythia~\cite{DBLP:conf/icml/BidermanSABOHKP23}, LLaMA~\cite{DBLP:journals/corr/abs-2302-13971}, etc. Currently, GPT-4~\cite{DBLP:journals/corr/abs-2303-08774} has pushed the boundaries of LLMs further in terms of scale and capability.


\subsection{Scaling Laws for Language Models}


The concept of scaling laws for language models was proposed by~\cite{DBLP:journals/corr/abs-2001-08361}. Their study revealed that the test loss for generative transformer models scales as a power-law with model size, dataset size, and the amount of computation used for training. 
Building upon that foundational study, further research has expanded the concept of scaling laws to diverse problem settings~\cite{DBLP:journals/corr/abs-2102-01293} and model architectures~\cite{DBLP:conf/cvpr/ChertiBWWIGSSJ23,DBLP:conf/icml/AghajanyanYCHHZ23}. For instance, ~\cite{DBLP:journals/corr/abs-2102-01293} has investigated scaling laws for transfer learning. 
\cite{DBLP:conf/cvpr/ChertiBWWIGSSJ23} and \cite{DBLP:conf/icml/AghajanyanYCHHZ23} observed scaling laws for multi-modal model pertaining.
Furthermore, some recent works discovered other patterns beside the power-law for different problem settings, such as continual pertaining \cite{que2024d}, data mixtures \cite{ye2024data}, and training instabilities \cite{wortsman2023small}. These works follow a \textit{small-model-to-large-model} perspective and predict the final performance of large models by fitting the final training results on smaller models.
Recently, researchers have also investigated scaling laws in the context of transfer learning \cite{DBLP:journals/corr/abs-2402-04177}, where model performance is influenced by both the size of the training dataset and the degree of downstream task alignment.

Despite previous advancements, a critical point that remains underexplored is the temporal trajectory of LLM performance throughout training. 
Previous works~\cite{DBLP:journals/corr/abs-2001-08361} estimate that 
the final test loss after training individually with different dataset sizes follows a power-law. However, such a coarse-grained estimation is not accurate enough in portraying the test loss evolution during a single pretrain process. 
By studying the loss behavior on different token positions, we introduce the temporal scaling law, allowing for more precise tracking and prediction of LLM test loss during pretraining. 
Our temporal scaling law focuses on the test loss evolution in a single pretrain process, while prior scaling laws model the relations of final test loss and the computational budget (model size, dataset size, etc.).


\begin{table*}[!h]
\centering
\scalebox{0.7}{\begin{tabular}
{ccccccccccc}
\toprule
 model size & dimension & intermediate & $n$ heads & $n$ layers & learning rate & scheduler & warmup & optimizer & batch size & seq length \\
\midrule
9.8M & 128 & 512 & 4 & 6 & $3.0e^{-4}$ & cosine & $1k$ steps & AdamW & 1024 & 1024 \\
58M & 512 & 2048 & 8 & 6 & $3.0e^{-4}$ & cosine & $1k$ steps & AdamW & 1024 & 1024 \\
468M & 1024 & 4096 & 16 & 24 & $3.0e^{-4}$ & cosine & $1k$ steps & AdamW & 1024 & 1024 \\
1.2B & 2048 & 8192 & 8 & 16 & $3.0e^{-4}$ & cosine & $1k$ steps & AdamW & 1024 & 1024 \\
6.7B & 4096 & 11008 & 32 & 32 & $3.0e^{-4}$ & cosine & $2k$ steps & AdamW & 2048 & 2048 \\
\bottomrule
\end{tabular}}
\caption{Model sizes, architectures, and optimization hyperparameters.}
\label{tab:model_hyperparameters}
\end{table*}

\begin{table*}[!h]
\centering
\scalebox{0.65}{\begin{tabular}
{ccccccccccc}
\toprule
 model name & dimension & intermediate & $n$ heads & $n$ layers & learning rate & scheduler & warmup & optimizer & batch size & seq length \\
\midrule
468M GPT-NeoX & 1024 & 4096 & 16 & 24 & $3.0e^{-4}$ & cosine & $1k$ steps & AdamW & 1024 & 1024 \\
1.4B MoE (act: 240M) & 768 & 3072 & 12 & 12 & $1.5e^{-4}$ & cosine & $2k$ steps & AdamW & 1024 & 1024 \\
\bottomrule
\end{tabular}}
\caption{Model sizes, architectures, and optimization hyperparameters for generalizing to more architectures.}
\label{tab:model_hyperparameters_generalize}
\end{table*}

\begin{table}[!h]
\centering
\scalebox{0.7}{\begin{tabular}
{lc|lc}
\toprule
 Package & Version & Package & Version \\
\midrule
PyTorch & 2.1.0 & transformers & 4.32.0 \\
deepspeed & 0.10.0 & tokenizers & 0.13.3 \\
flash-attn & 2.3.6 & lm-evaluation-harness & 0.3.0 \\
datasets & 2.14.3 &  &  \\
\bottomrule
\end{tabular}}
\caption{Versions of used packages.}
\label{tab:package_version}
\end{table}

\begin{table}[!t]
\centering
\scalebox{0.6}{\begin{tabular}
{l|ccccc}
\toprule
 & Hyperbolic & Logarithmic & Power-law & Log-log & Exponential \\
\midrule
\cref{eq:hyperbolic} & \textbf{>99\%} & 5\% & 67\% & 31\% & 0\% \\
\cref{eq:a0a1_before_sep} & 0.690/\textbf{0.921} & 0.788/0.659 & 0.342/0.441 & \textbf{0.953}/0.731 & -1.231/-5.002 \\
\bottomrule
\end{tabular}}
\caption{Comparison of functional form choices for \cref{eq:hyperbolic,eq:a0a1_before_sep} on ID-Val for the 58M model. For \cref{eq:hyperbolic}, we report the percentage of checkpoints that could achieve $R^2>0.95$ for each functional form. For \cref{eq:a0a1_before_sep}, we report the $R^2$ values fitting with each functional form ($a_0$/$a_1$). \textbf{Bold} represents the best result.}
\label{tab:functional_form_comparison}
\end{table}

\section{Further Experiment Details}
\label{sec:more_exp_setting}
In this section, we provide further details toward the experiment settings in the article.

\begin{algorithm*}[t]
\caption{Predict Pipeline Using the Temporal Scaling Law}
\begin{algorithmic}[1]
\REQUIRE LLM $M$, Tokenized Train Corpus $\mathcal{T}$, Tokenized Evaluation Corpus $\mathcal{E}$, Prediction Point $N_{pred}$ Tokens, Total Training Tokens $N_{tot}$, Batch Size $N_{batch}$, Evaluation Interval $N_{eval}$ Tokens
\STATE Initilize $M$ for training. Split $\mathcal{T}$ and $\mathcal{E}$ to sequences of $l$ consecutive tokens.
\STATE Initialize the number of currently trained tokens $N_{cur}=0$
\STATE Initialize the list of token position loss and the corresponding number of trained tokens $L=[\quad]$
\STATE /****** Train $M$ for an early pretraining period  ******/
\WHILE{$N_{cur} < N_{pred}$}
    \STATE Update model $M$ with the current batch data
    \STATE $N_{cur} \leftarrow N_{cur} + N_{batch}$
    \IF {$N_{cur} \% N_{eval} == 0$} 
        \STATE // evaluate $M$ for a fitting data point
        \STATE Evaluate $M$ on $\mathcal{E}$, record the average loss on every token position as $\mathbf{l}$, $len(\mathbf{l})=l$
        \STATE $L.\text{append}([\mathbf{l}, N_{cur}])$
    \ENDIF 
    \IF {$N_{cur} \ge N_{pred}$} 
        \STATE \textbf{break}
    \ENDIF 
\ENDWHILE
\STATE /****** Predict the test loss curve of $M$  ******/
\STATE Initialize $a_0$ list $A_0=[\quad]$, $a_1$ list $A_1=[\quad]$, $a_2$ list $A_2=[\quad]$
\FOR{each $\mathbf{l}_{cur}, N_{cur}$ in $L$}
    \STATE Fit $\mathbf{l}_{cur}$ for $a_0^{cur}$, $a_1^{cur}$, and $a_2^{cur}$ with Eq. 5.
    \STATE Store fitting results $A_0.\text{append}(a_0^{cur})$, $A_1.\text{append}(a_1^{cur})$, $A_2.\text{append}(a_2^{cur})$
\ENDFOR
\STATE Calculate through $A_0$ and $A_1$, verify whether $\Tilde{N}_{sep}$ has been reached according to Eq. 7.
\STATE Apply the fit in Eq. 11 for $\Tilde{a}_0^N$ and $\Tilde{a}_1^N$
\STATE Based on the distance of the fit $\Tilde{a}_0^N$, $\Tilde{a}_1^N$ and the actual ${a}_0^N$, ${a}_1^N$, filter out outlier points from $A_0$, $A_1$
\STATE Re-fit for $\Tilde{\alpha}_i$ and $\Tilde{\beta}_i$ according to Eq. 11 
\STATE Fit for $\Tilde{a}_2^N$ with Eq. 12
\STATE Apply the boundary conditions on $\Tilde{a}_0^N$, $\Tilde{a}_1^N$, and $\Tilde{a}_2^N$ for $N> \Tilde{N}_{sep}$ according to Eq. 13
\IF {$N_{pred}\le \Tilde{N}_{sep}$}
    \STATE Use $N_{pred}\le \Tilde{N}_{sep}$ data points in $A_2$ to calibrate $\Tilde{a}_2^N$ following Eq. 15.
\ENDIF
\STATE Substitute $\Tilde{a}_0^N$, $\Tilde{a}_1^N$, and $\Tilde{a}_2^N$ in Eq. 10 for the loss outcome prediction $\Tilde{L}^N$
\RETURN $\Tilde{L}^N$
\end{algorithmic}
\label{alg:predict_pipeline}
\end{algorithm*}

\textbf{License for Scientific Artifacts.}
The Pile \cite{DBLP:journals/corr/abs-2101-00027} is subject to the MIT license\footnote{https://arxiv.org/pdf/2201.07311}. The C4 dataset \cite{DBLP:journals/jmlr/RaffelSRLNMZLL20} is licensed under Open Data Commons License Attribution family\footnote{https://huggingface.co/datasets/allenai/c4}. 
The LAMBADA dataset \cite{DBLP:conf/acl/PapernoKLPBPBBF16} is licensed under Creative Commons Attribution 4.0 International license\footnote{https://huggingface.co/datasets/cimec/lambada}. 
The Wikitext dataset \cite{DBLP:conf/iclr/MerityX0S17} is licensed under the Creative Commons Attribution 4.0 International licence\footnote{https://zenodo.org/records/2630551}.
The evaluation benchmarks \cite{eval-harness} are subject to the MIT license. All usages of scientific artifacts in this paper obey the corresponding licenses.

\textbf{Hyperparameters of Models Used.}
We report the details of model hyperparameters and training hyperparameters in \cref{tab:model_hyperparameters}. Note that we use the 9.8M and the 58M models for illustrating our temporal scaling law. Meanwhile, we apply the predictions and further applications of our temporal scaling law to the larger 468M and the 1.2B models.

\textbf{Parameters for Packages.}
We report the version numbers of used packages in \cref{tab:package_version}.

\textbf{Evaluation Pipeline.}
For all benchmark evaluations, we utilize the open-source LLM evaluation tool \texttt{lm-evaluation-harness}\footnote{https://github.com/EleutherAI/lm-evaluation-harness} \cite{eval-harness}, following \cite{DBLP:journals/corr/abs-2310-19531,DBLP:conf/icml/BidermanSABOHKP23}. For all numerical results, we report the average result tested on the last 5 checkpoints.

\section{Theoretical Insights for the Dynamic Hyperbolic Law}
\label{sec:theoretical_hyperbolic}
Following prior work on validating scaling laws \cite{DBLP:journals/corr/abs-2001-08361,DBLP:journals/corr/abs-2203-15556}, we conducted comprehensive experiments and empirically confirmed the effectiveness of the hyperbolic pattern in the main article.
In the main article, the main functional forms that we have considered are the logarithmic function ($\mathcal{L}_i=a_1\log(a_2i+a_3)+a_4$), power-law ($\mathcal{L}_i=a_1i^{a_2}+a_3$), log-log function ($\mathcal{L}_i=a_1\log(a_2\log(i)+a_3)+a_4$), and exponential function ($\mathcal{L}_i=a_1\cdot a_2^{i}+a_3$) (see \cref{sec:test_loss_pred} for the experimental comparisons with those functions).
In addition, we provide a theoretical justification for selecting the hyperbolic function among candidate forms. 
Specifically, the loss with respect to token positions should approach a lower bound, since the cross-entropy loss has a minimum of zero. This constraint limits the valid choices to the power-law, exponential, and hyperbolic functions.
We then analyze the simplified forms of the power-law and exponential functions: for the power-law, $a_1>0$, $a_2<0$, $a_3=0$; for the exponential, $a_1>0$, $0<a_2<1$, $a_3=0$. It is straightforward to show that the exponential function decays faster than the power-law. Formally, for all valid parameters, there exists an $i_0 \in \mathbb{N}$ such that $\mathcal{L}^{power}_i > \mathcal{L}^{exp}_i$ for all $i > i_0$. Given that real-world corpus data typically does not yield losses approaching zero, the exponential function is impractical. 
Lastly, we note that the hyperbolic function can be rewritten in a power-law-like form as $\mathcal{L}_i = a_0(a_1 i + 1)^{-1} + a_2$. Since this formulation not only satisfies theoretical constraints but also yields better empirical performance, we adopt it as the underlying structure for modeling token position loss.

\section{Comparison with Other Functional Form Choices}
\label{sec:append_functional_comparison}

In this article, for fitting general ``increasing'' and ``decreasing'' patterns (i.e., \cref{eq:hyperbolic,eq:a0a1_before_sep}), we consider using the basic functional forms described in \cref{sec:theoretical_hyperbolic}.
We compare the fitting results of different functional forms on the 58M model.
Specifically, for \cref{eq:hyperbolic}, we report the percentage of checkpoints that could achieve $R^2>0.95$ for each functional form. For \cref{eq:a0a1_before_sep}, we report the $R^2$ values fitting with each functional form. 
The results are presented in \cref{tab:functional_form_comparison}. Based on these results, we selected the hyperbolic function for fitting the token position loss, the log-log function for $a_0$, and the hyperbolic function for $a_1$, as these provided the best fit.

\section{Pipeline for Predicting Training Outcomes Using the Temporal Scaling Law}
We provide the pipeline for predicting the training outcomes using the temporal scaling law in \cref{alg:predict_pipeline}. Note that we use non-linear least squares to solve for all fitting parameters. Normally, the fitting would converge in $<10^3$ steps, which takes <1 second on a CPU. We conduct a re-fit for $a_0$ and $a_1$ to mitigate the effect of fluctuations (as observed in \cref{fig:temporal_law}).

\section{Complete Fitting Results}
\label{sec:more_fitting}
In this section, we present complete fitting figures illustrating our proposed temporal scaling law.

\subsection{More Results of Dynamic Hyperbolic-Law}
\begin{figure*}
  \centering
  \begin{tabular}{cccc}
  
  \begin{subfigure}{0.22\linewidth}
    \includegraphics[width=1.0\linewidth]{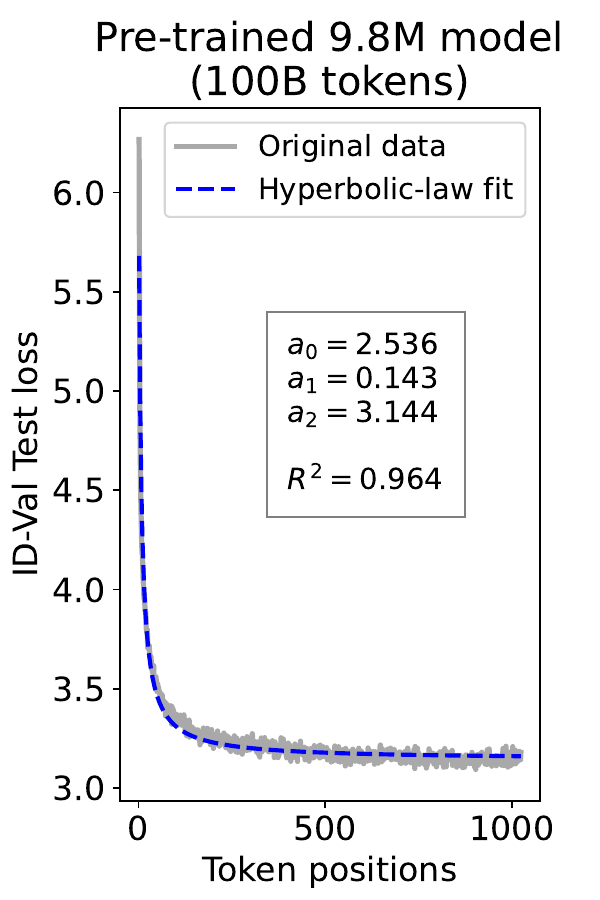}
  \end{subfigure} & 
  \begin{subfigure}{0.22\linewidth}
    \includegraphics[width=1.0\linewidth]{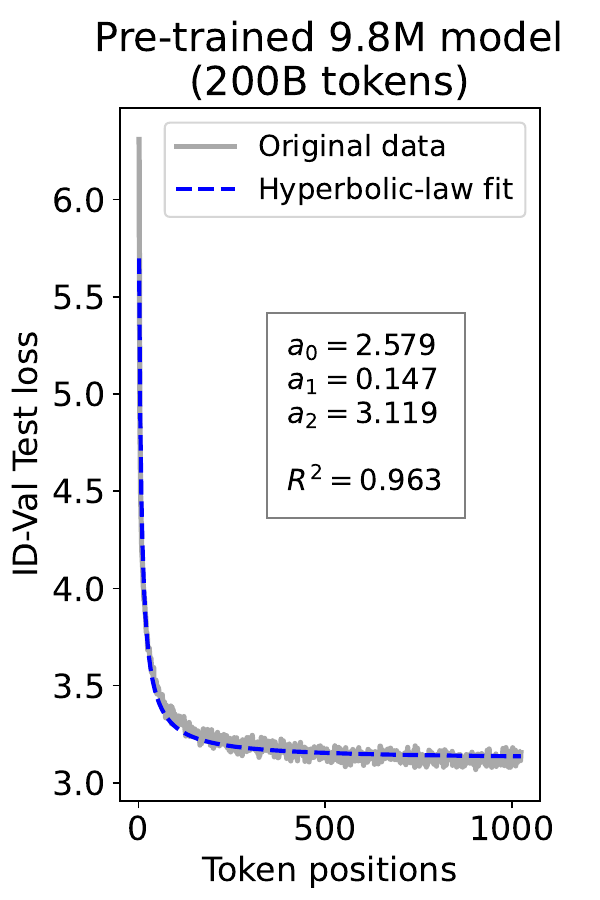}
  \end{subfigure} & 
  \begin{subfigure}{0.22\linewidth}
    \includegraphics[width=1.0\linewidth]{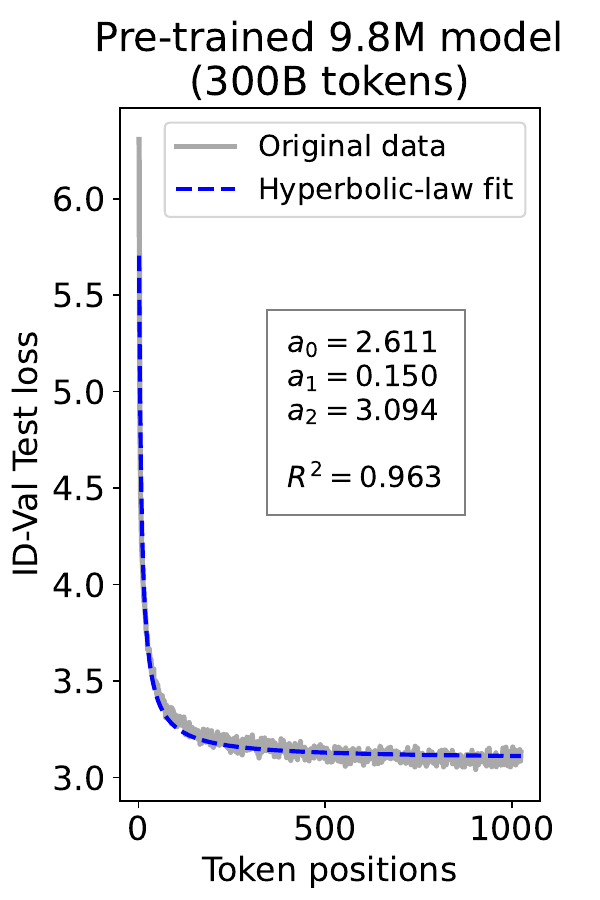}
  \end{subfigure}
  \begin{subfigure}{0.22\linewidth}
    \includegraphics[width=1.0\linewidth]{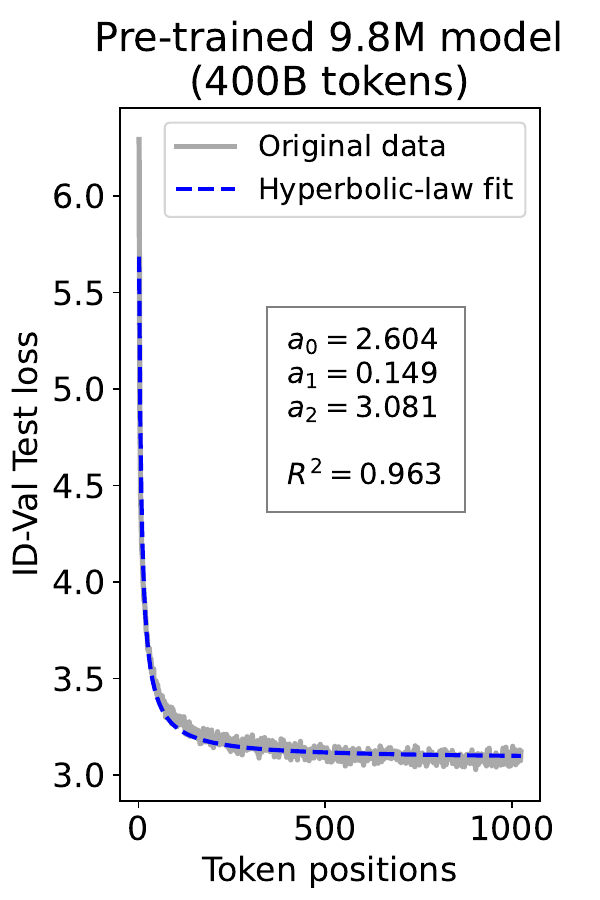}
  \end{subfigure} \\

  \begin{subfigure}{0.22\linewidth}
    \includegraphics[width=1.0\linewidth]{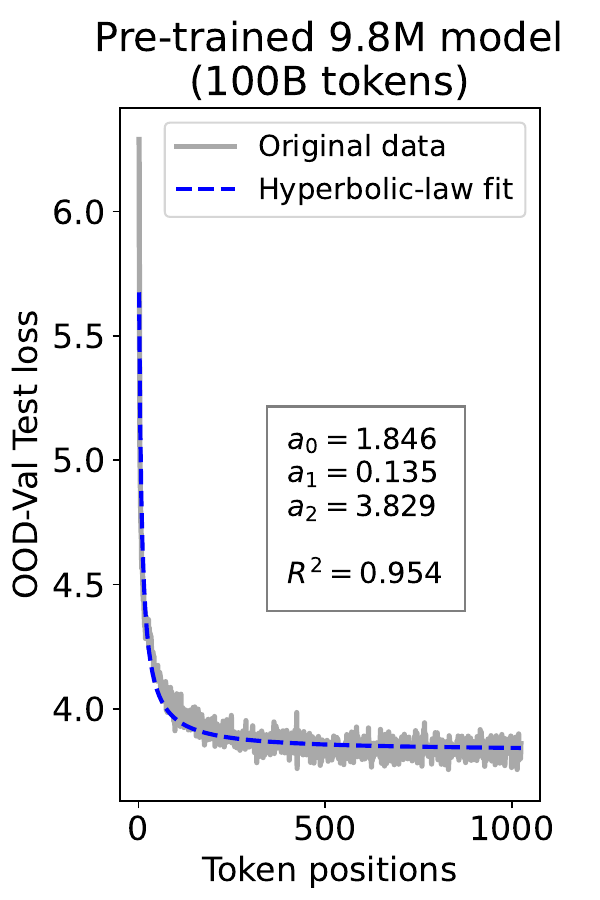}
  \end{subfigure} & 
  \begin{subfigure}{0.22\linewidth}
    \includegraphics[width=1.0\linewidth]{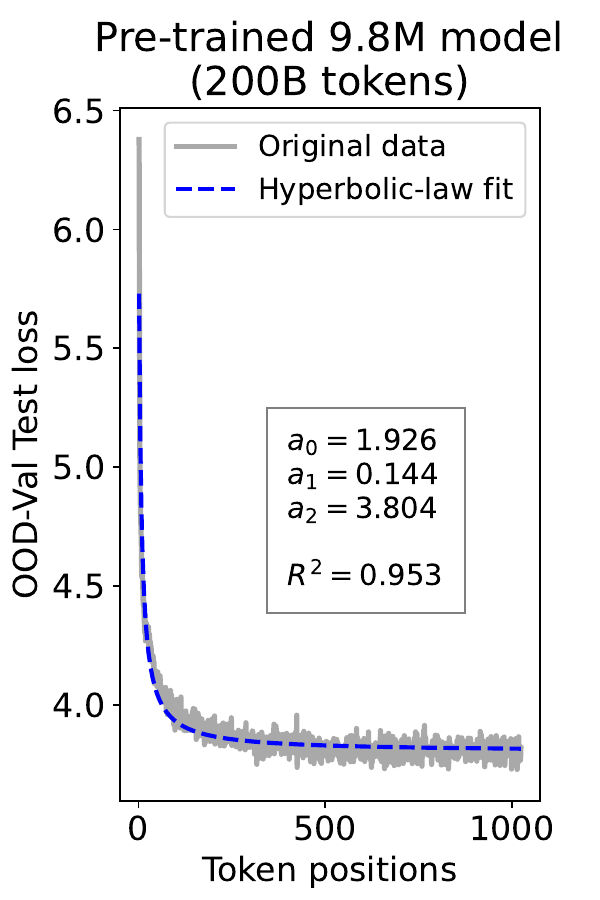}
  \end{subfigure} & 
  \begin{subfigure}{0.22\linewidth}
    \includegraphics[width=1.0\linewidth]{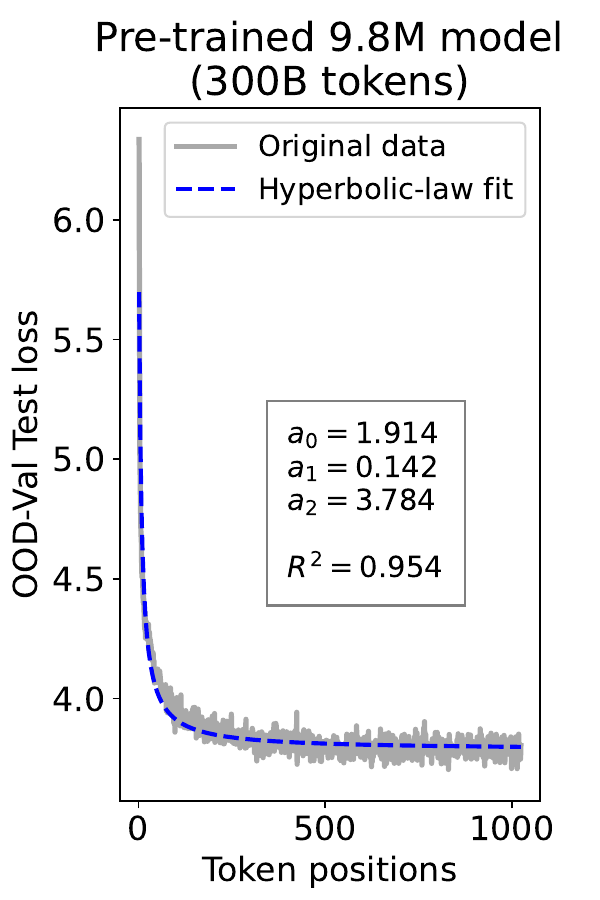}
  \end{subfigure}
  \begin{subfigure}{0.22\linewidth}
    \includegraphics[width=1.0\linewidth]{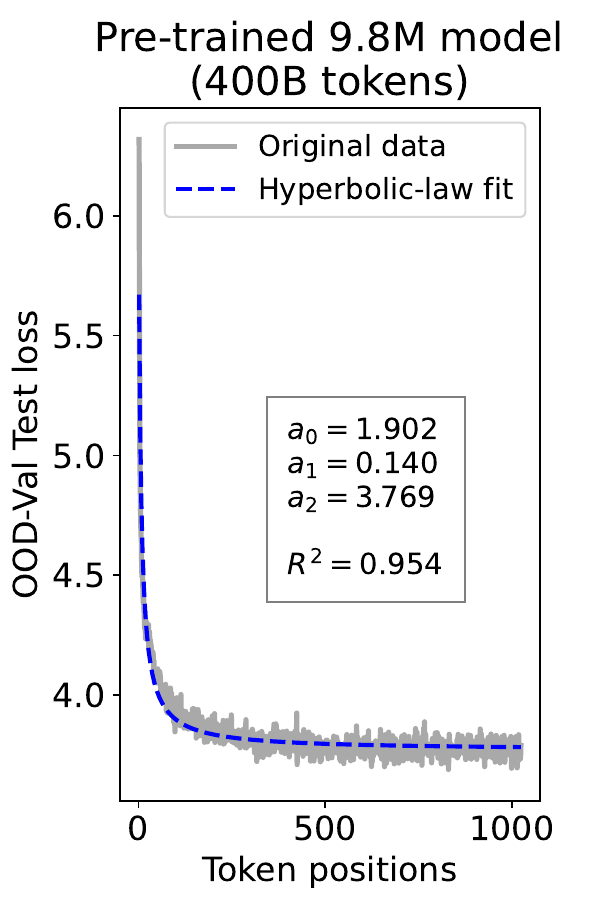}
  \end{subfigure} \\
  \end{tabular}
  \caption{More dynamic hyperbolic-law fitting results for the 9.8M model after training for 100B, 200B, 300B, and 400B tokens on both the ID-Val and OOD-Val.}
  \label{fig:token_loss_complete_9_8m}
\end{figure*}

\begin{figure*}
  \centering
  \begin{tabular}{cccc}

  \begin{subfigure}{0.22\linewidth}
    \includegraphics[width=1.0\linewidth]{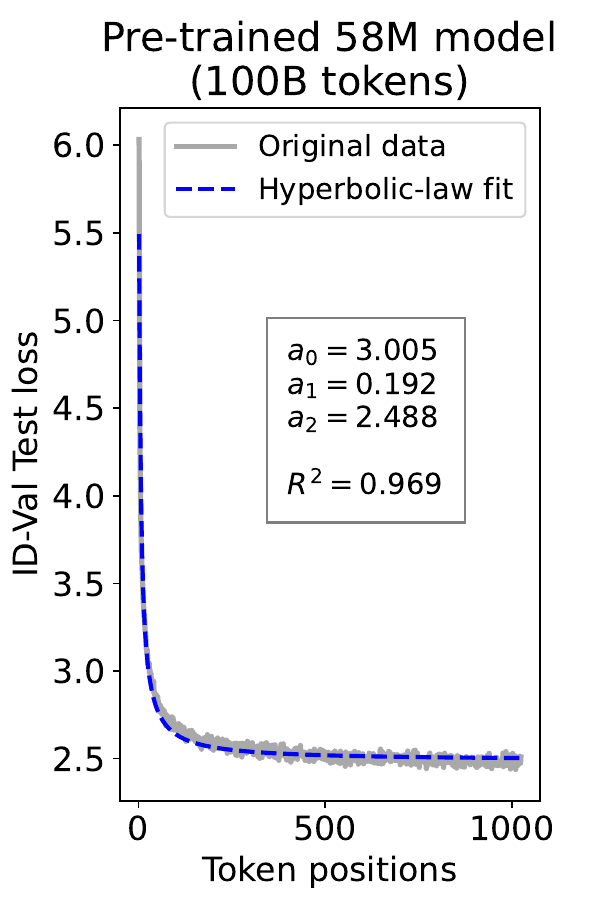}
  \end{subfigure} & 
  \begin{subfigure}{0.22\linewidth}
    \includegraphics[width=1.0\linewidth]{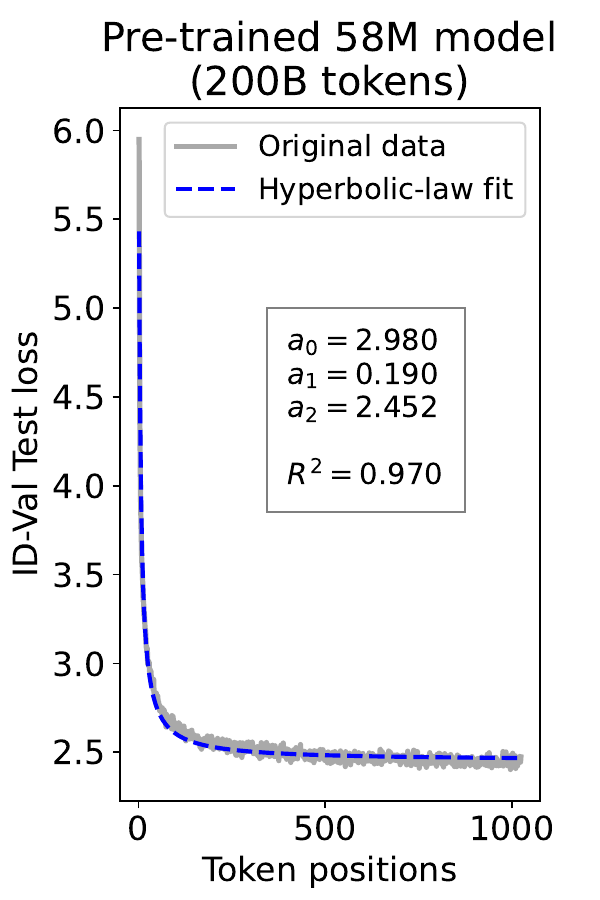}
  \end{subfigure} & 
  \begin{subfigure}{0.22\linewidth}
    \includegraphics[width=1.0\linewidth]{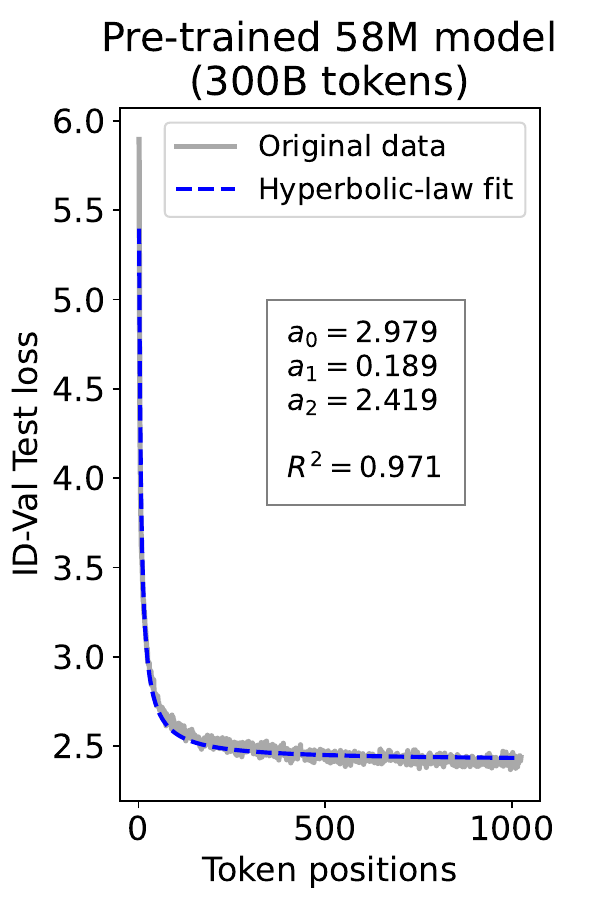}
  \end{subfigure}
  \begin{subfigure}{0.22\linewidth}
    \includegraphics[width=1.0\linewidth]{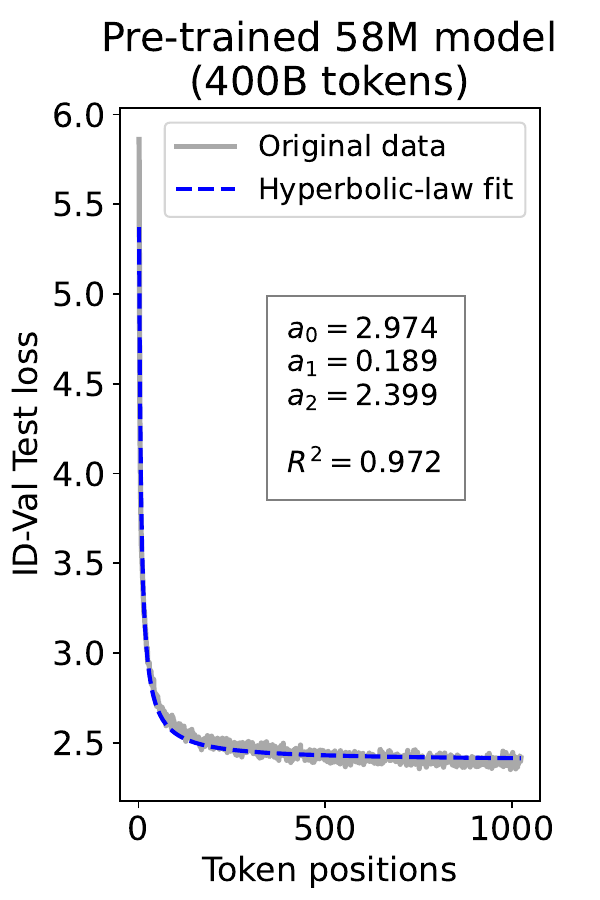}
  \end{subfigure} \\

  \begin{subfigure}{0.22\linewidth}
    \includegraphics[width=1.0\linewidth]{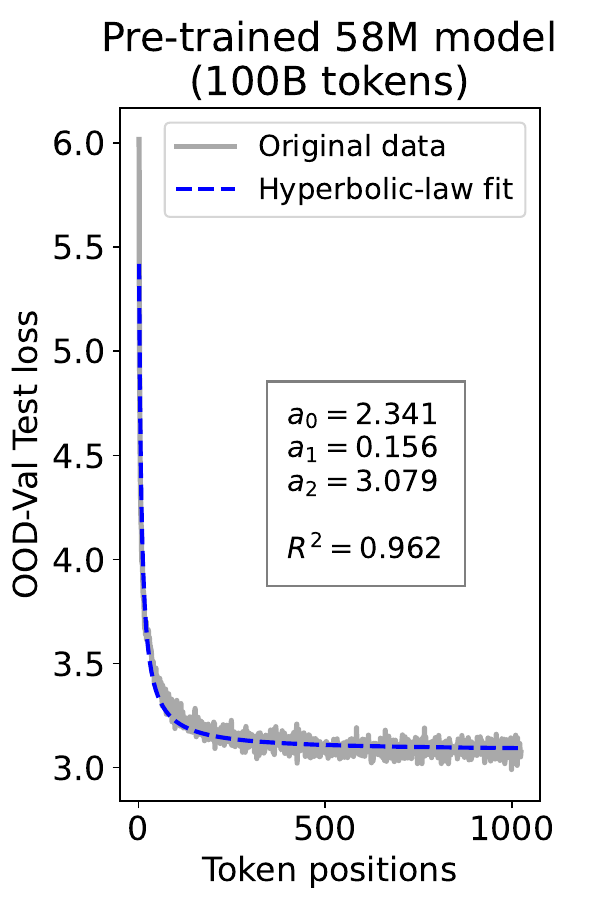}
  \end{subfigure} & 
  \begin{subfigure}{0.22\linewidth}
    \includegraphics[width=1.0\linewidth]{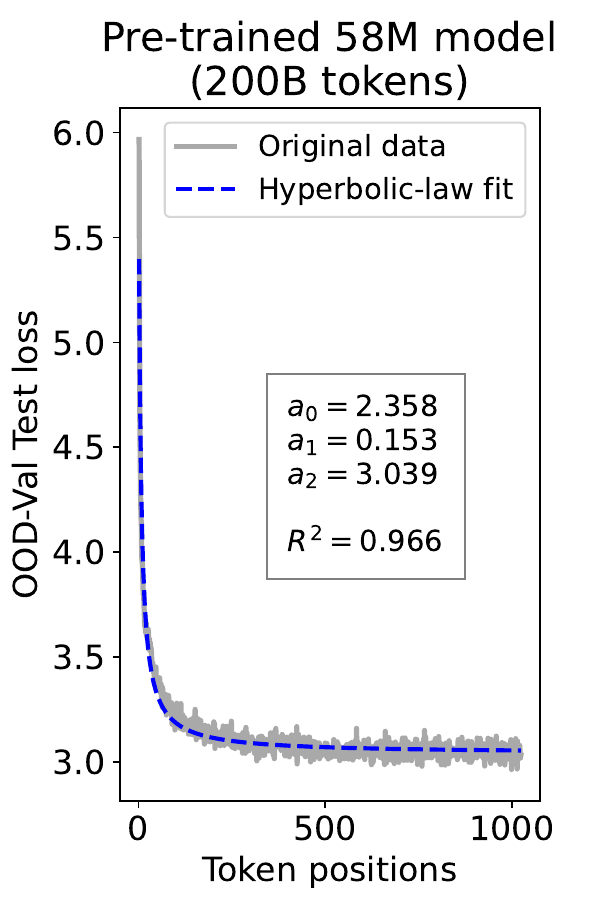}
  \end{subfigure} & 
  \begin{subfigure}{0.22\linewidth}
    \includegraphics[width=1.0\linewidth]{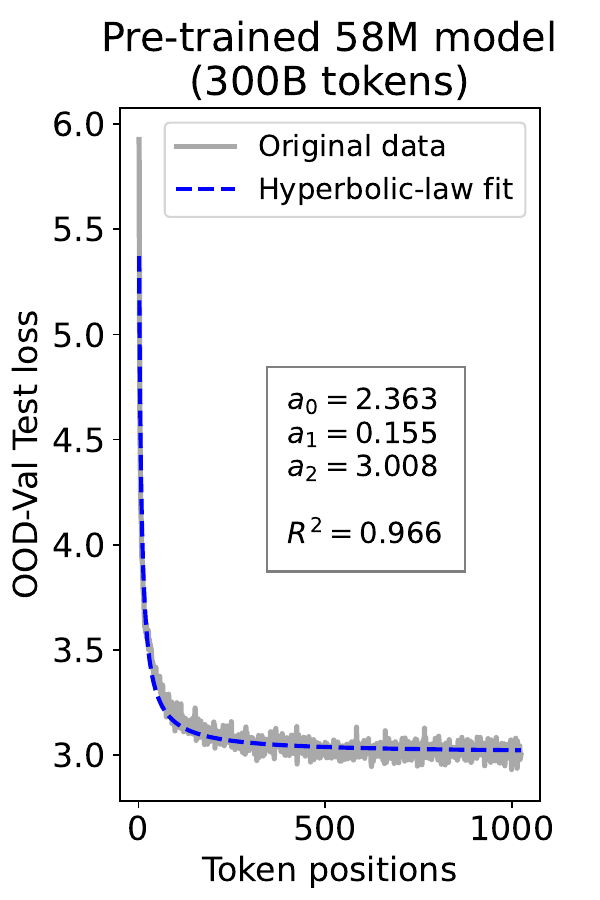}
  \end{subfigure}
  \begin{subfigure}{0.22\linewidth}
    \includegraphics[width=1.0\linewidth]{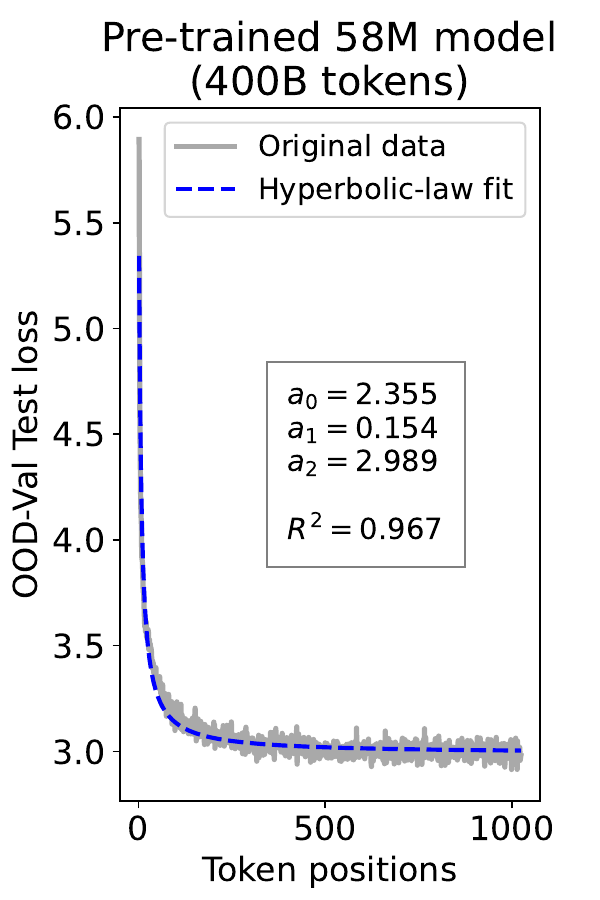}
  \end{subfigure} \\
  \end{tabular}
  \caption{More dynamic hyperbolic-law fitting results for the 58M model after training for 100B, 200B, 300B, and 400B tokens on both the ID-Val and OOD-Val.}
  \label{fig:token_loss_complete_58m}
\end{figure*}

As a supplement to \cref{fig:token_loss}, we provide fitting results of the dynamic hyperbolic-law for both the 9.8M and 58M models after training for 100B, 200B, 300B, and 400B tokens and on the OOD-Val in \cref{fig:token_loss_complete_9_8m} and \cref{fig:token_loss_complete_58m}, respectively. Those results indicate that our dynamic hyperbolic-law consistently achieves accurate fitting results across model sizes and training steps.

\subsection{More Results of Temporal Scaling Law}
\begin{figure*}
  \centering
  \begin{tabular}{ccc}
  
  \begin{subfigure}{0.3\linewidth}
    \includegraphics[width=1.0\linewidth, trim={0cm 0cm 1.3cm 0cm}, clip]{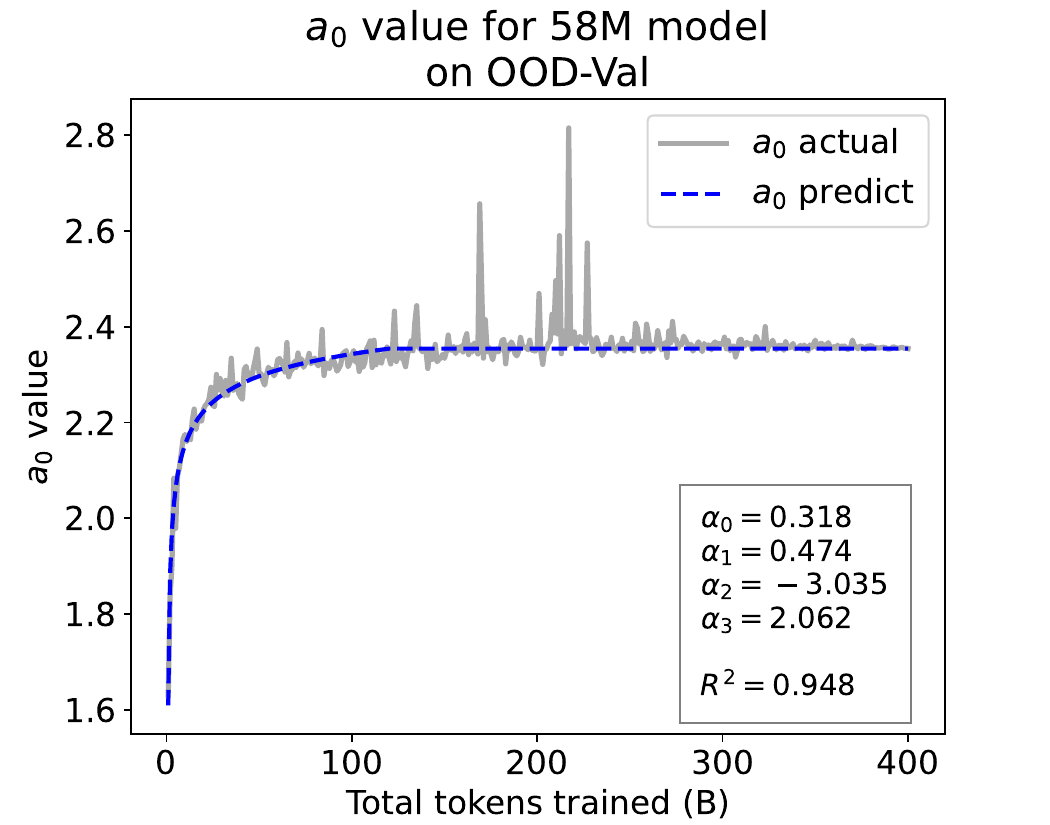}
    \vspace{5pt}
  \end{subfigure} & 
  \begin{subfigure}{0.3\linewidth}
    \includegraphics[width=1.0\linewidth, trim={0cm 0cm 1.3cm 0cm}, clip]{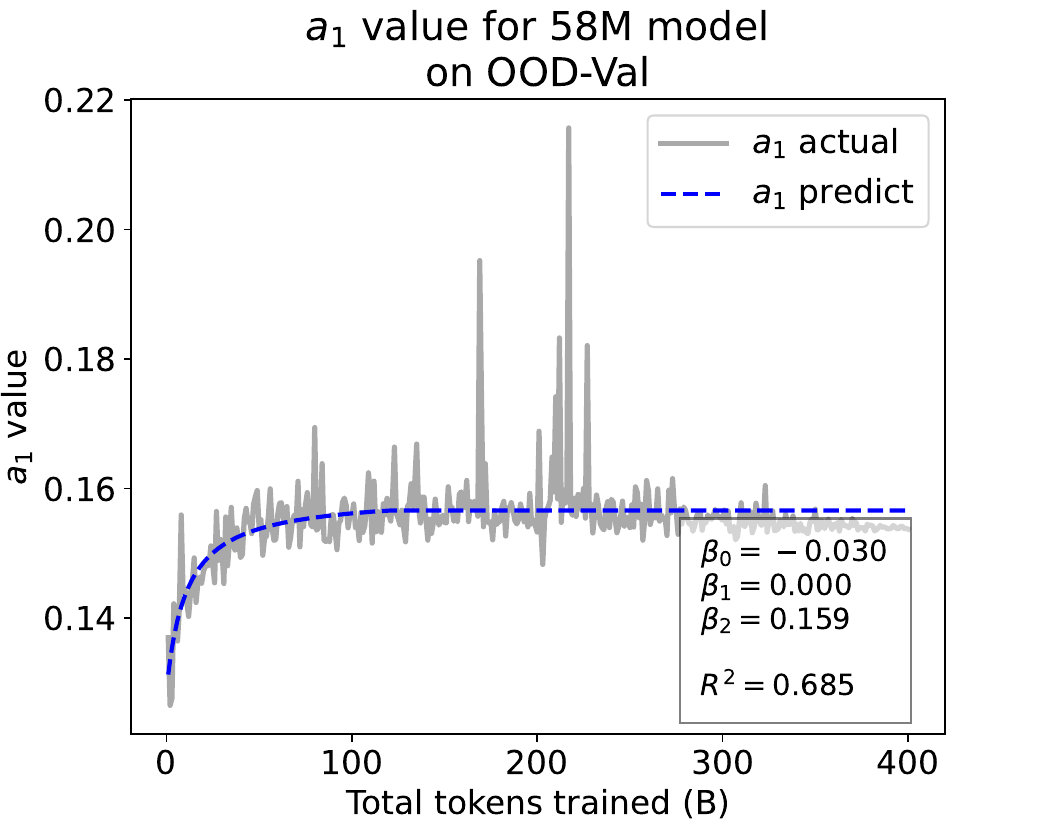}
    \vspace{5pt}
  \end{subfigure} & 
  \begin{subfigure}{0.3\linewidth}
    \includegraphics[width=1.0\linewidth, trim={0cm 0cm 1.3cm 0cm}, clip]{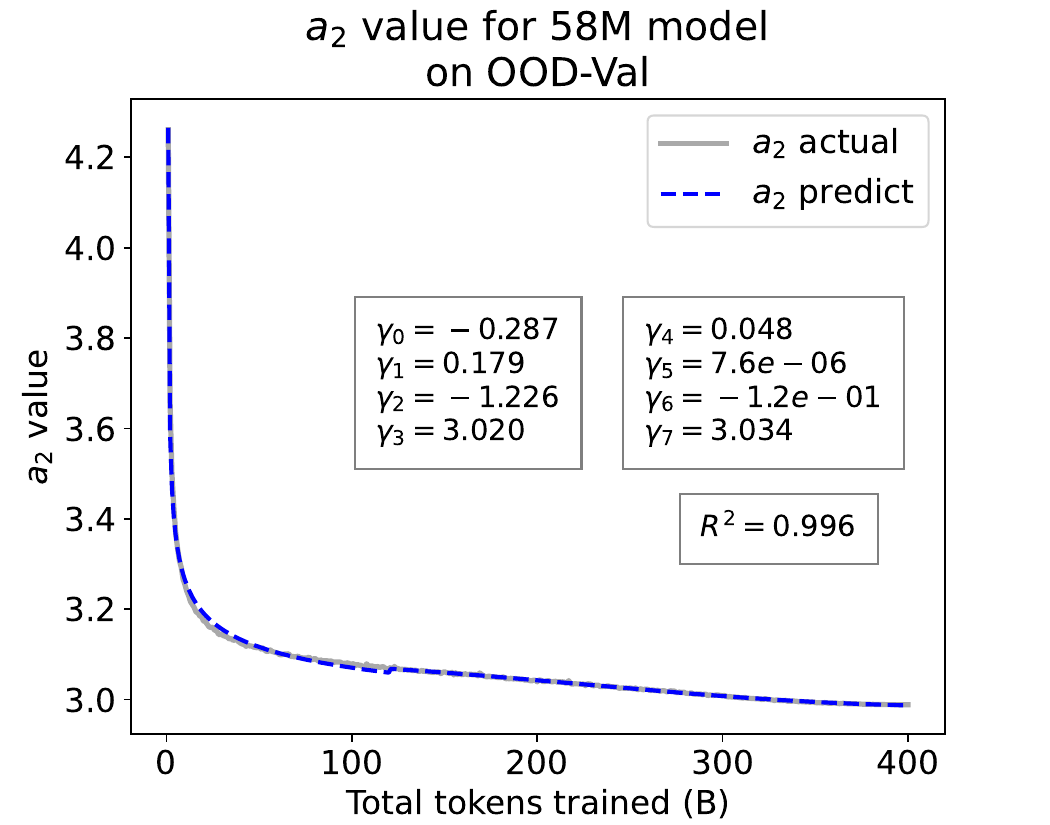}
    \vspace{5pt}
  \end{subfigure} \\
  \end{tabular}
  \caption{Temporal scaling law fitting results for the 58M model on the OOD-Val.}
  \label{fig:fit_temporal_58m_ood}
\end{figure*}

\begin{figure*}
  \centering
  \begin{tabular}{ccc}
  
  \begin{subfigure}{0.3\linewidth}
    \includegraphics[width=1.0\linewidth, trim={0cm 0cm 1.3cm 0cm}, clip]{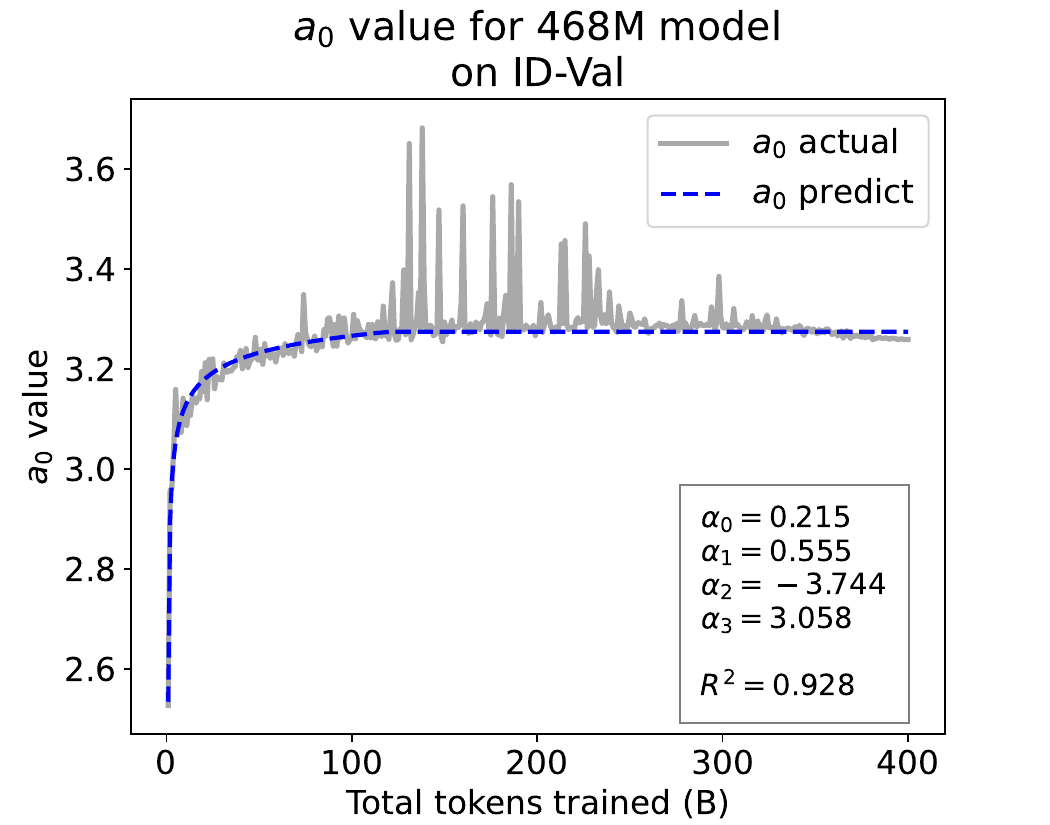}
    \vspace{5pt}
  \end{subfigure} & 
  \begin{subfigure}{0.3\linewidth}
    \includegraphics[width=1.0\linewidth, trim={0cm 0cm 1.3cm 0cm}, clip]{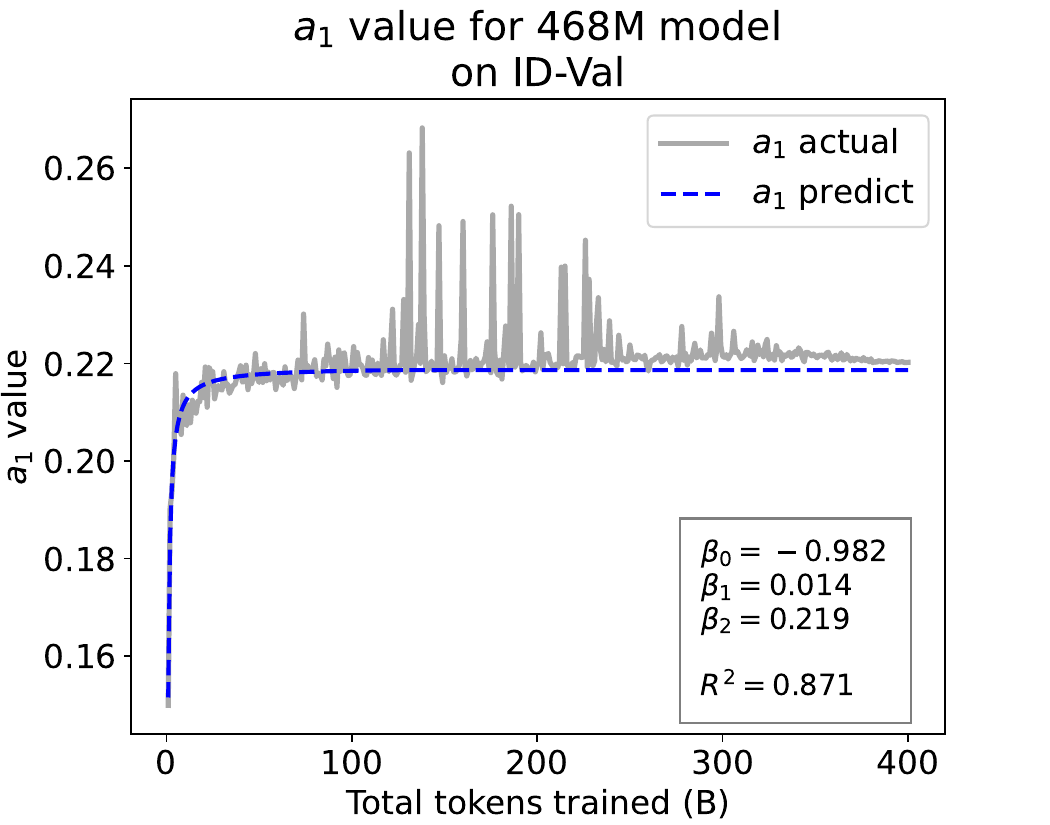}
    \vspace{5pt}
  \end{subfigure} & 
  \begin{subfigure}{0.3\linewidth}
    \includegraphics[width=1.0\linewidth, trim={0cm 0cm 1.3cm 0cm}, clip]{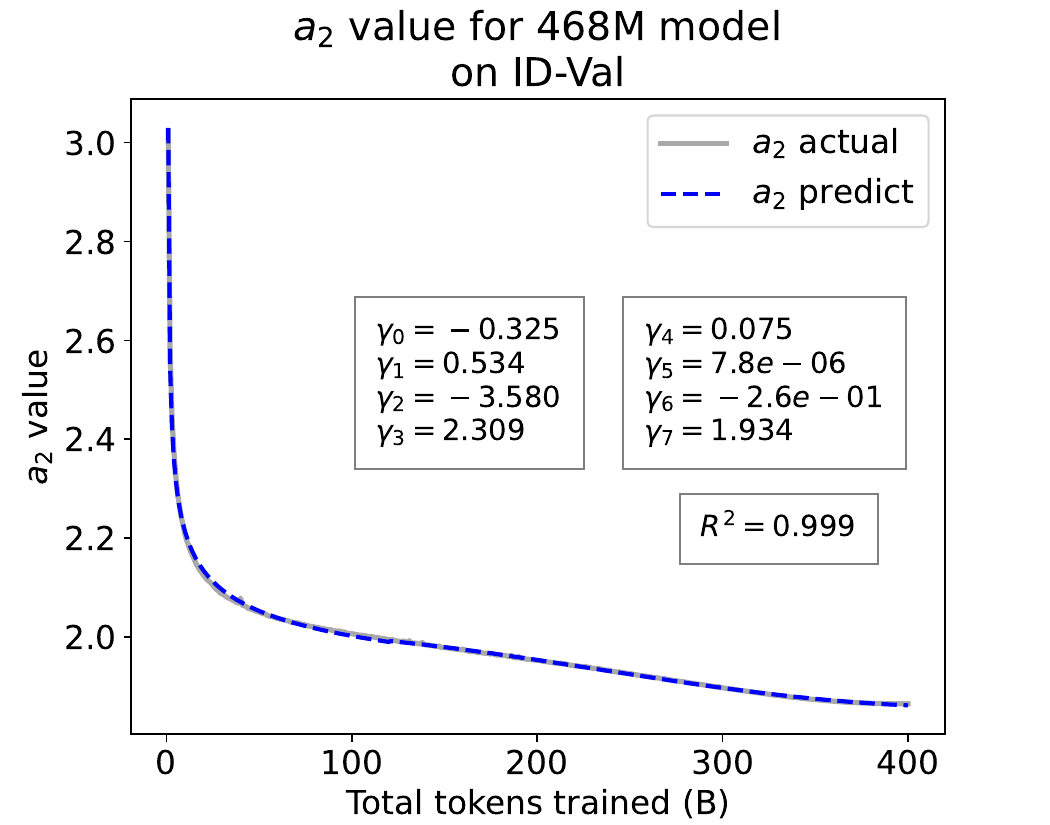}
    \vspace{5pt}
  \end{subfigure} \\
  
  \begin{subfigure}{0.3\linewidth}
    \includegraphics[width=1.0\linewidth, trim={0cm 0cm 1.3cm 0cm}, clip]{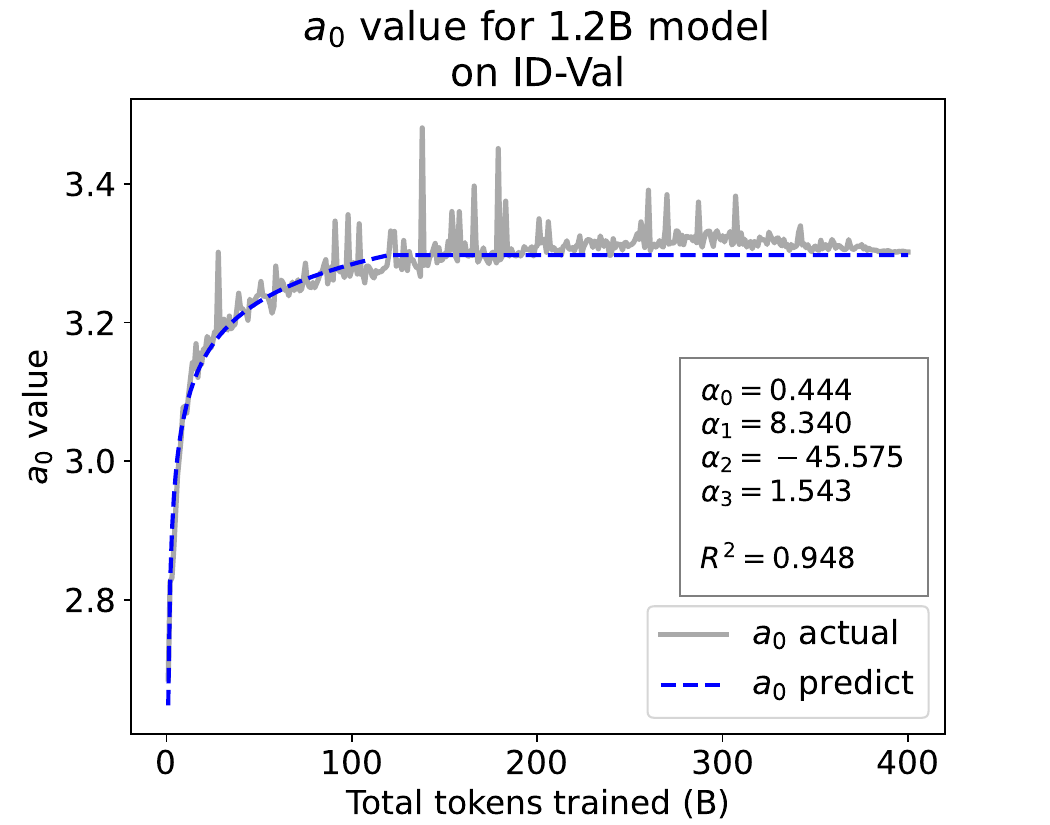}
    \vspace{5pt}
  \end{subfigure} & 
  \begin{subfigure}{0.3\linewidth}
    \includegraphics[width=1.0\linewidth, trim={0cm 0cm 1.3cm 0cm}, clip]{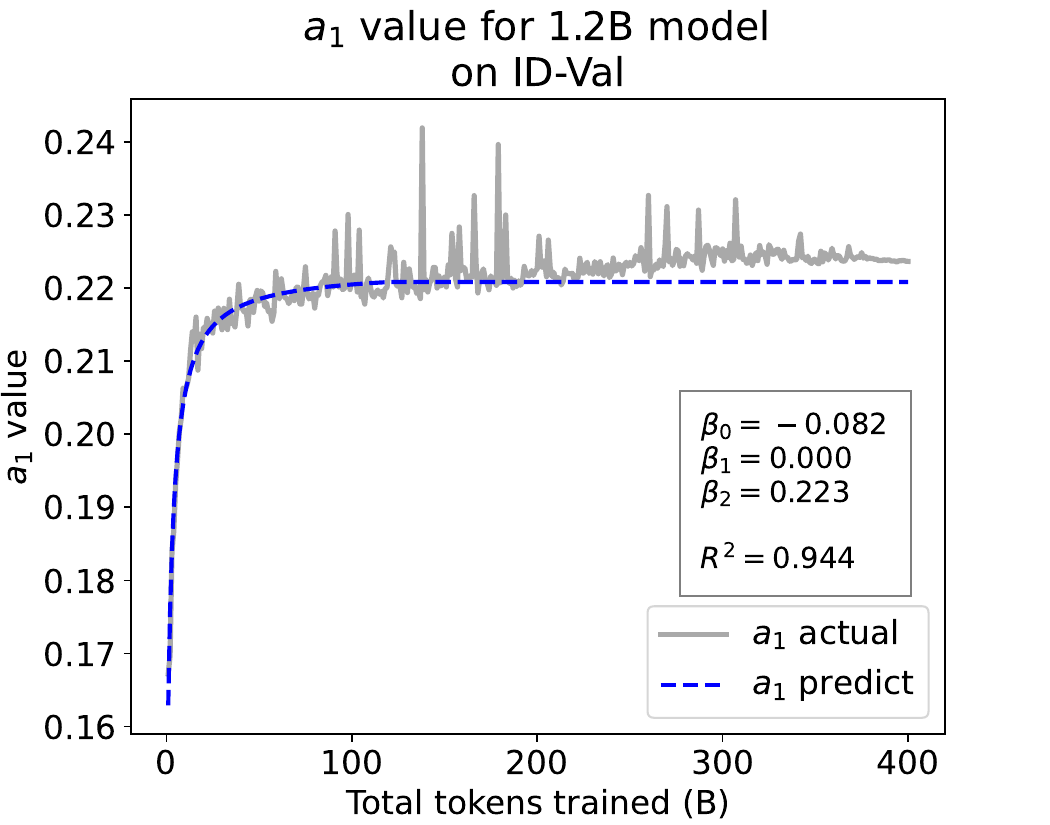}
    \vspace{5pt}
  \end{subfigure} & 
  \begin{subfigure}{0.3\linewidth}
    \includegraphics[width=1.0\linewidth, trim={0cm 0cm 1.3cm 0cm}, clip]{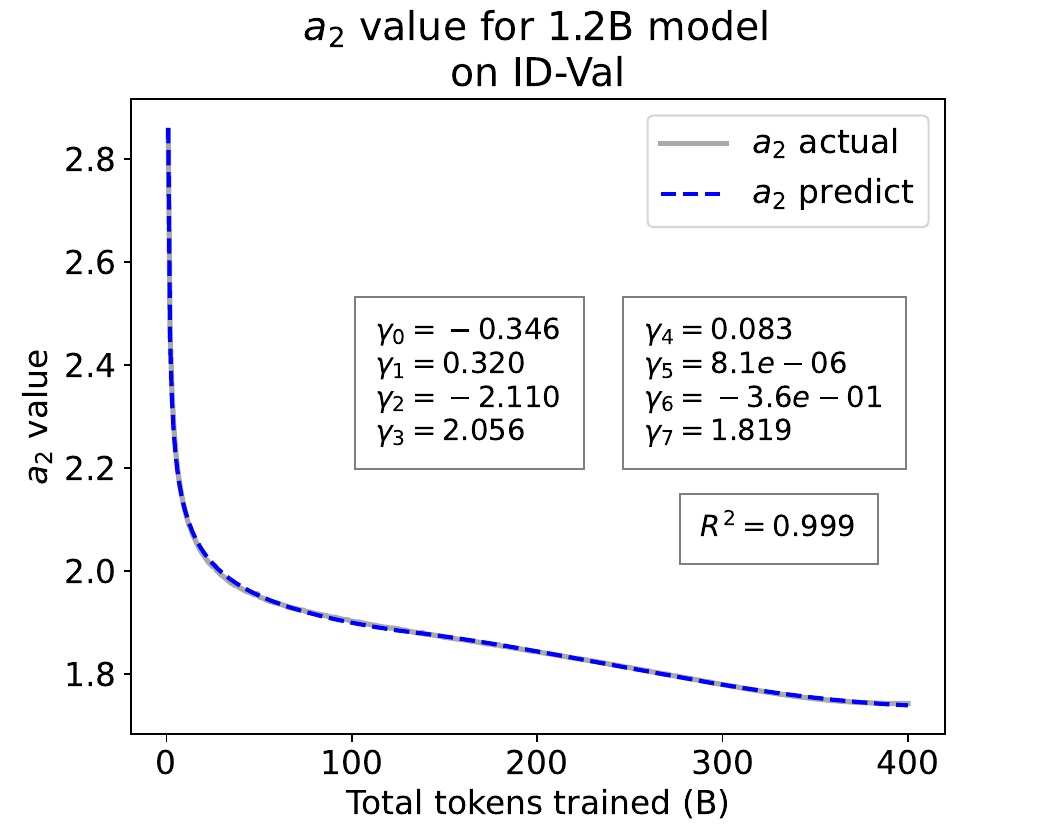}
    \vspace{5pt}
  \end{subfigure} \\

  \begin{subfigure}{0.3\linewidth}
    \includegraphics[width=1.0\linewidth, trim={0cm 0cm 1.3cm 0cm}, clip]{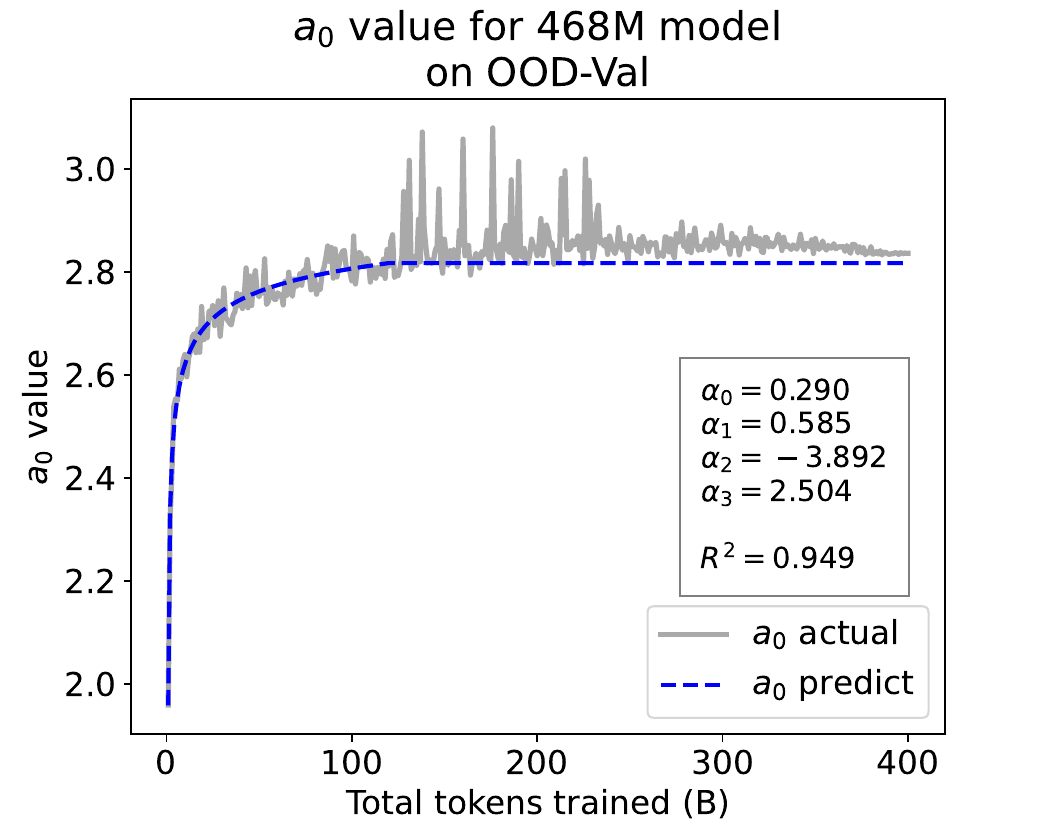}
    \vspace{5pt}
  \end{subfigure} & 
  \begin{subfigure}{0.3\linewidth}
    \includegraphics[width=1.0\linewidth, trim={0cm 0cm 1.3cm 0cm}, clip]{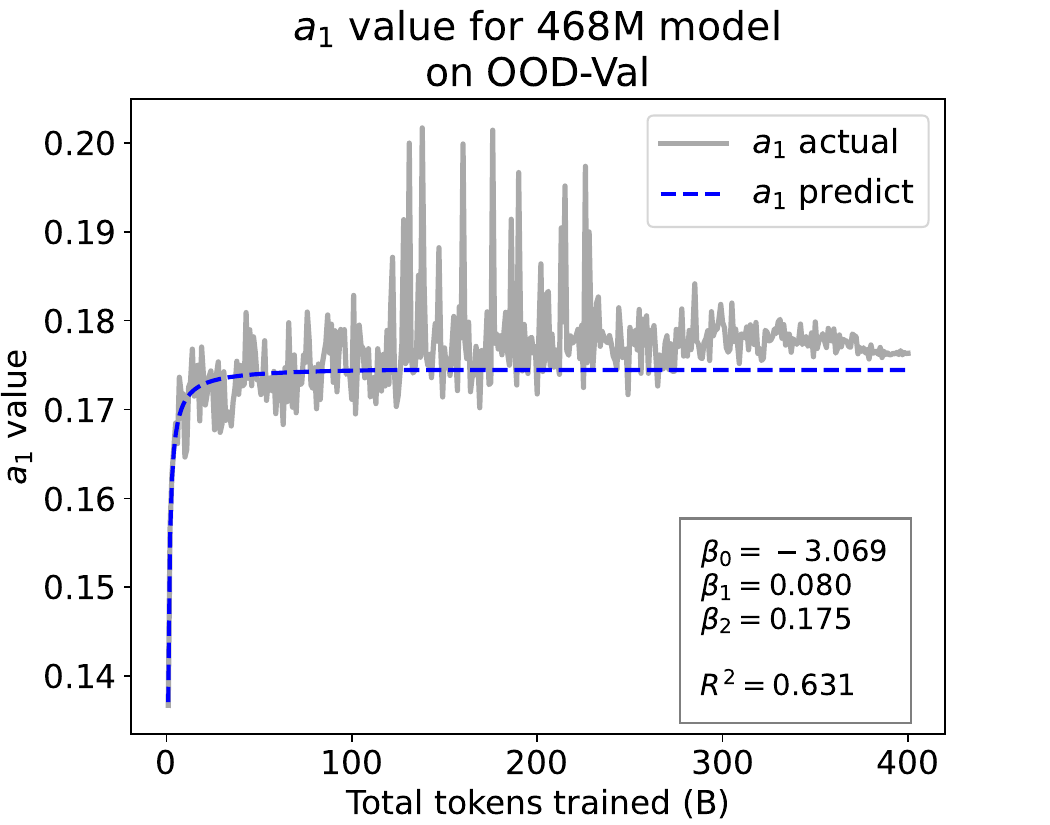}
    \vspace{5pt}
  \end{subfigure} & 
  \begin{subfigure}{0.3\linewidth}
    \includegraphics[width=1.0\linewidth, trim={0cm 0cm 1.3cm 0cm}, clip]{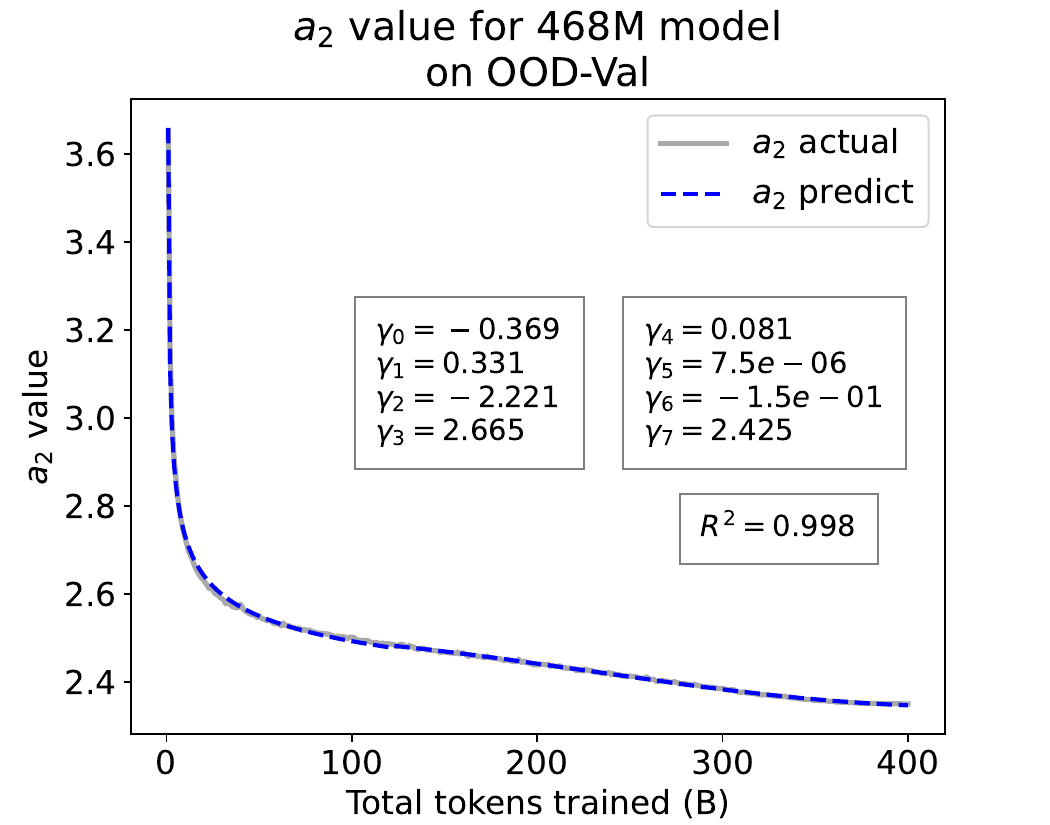}
    \vspace{5pt}
  \end{subfigure} \\

  \begin{subfigure}{0.3\linewidth}
    \includegraphics[width=1.0\linewidth, trim={0cm 0cm 1.3cm 0cm}, clip]{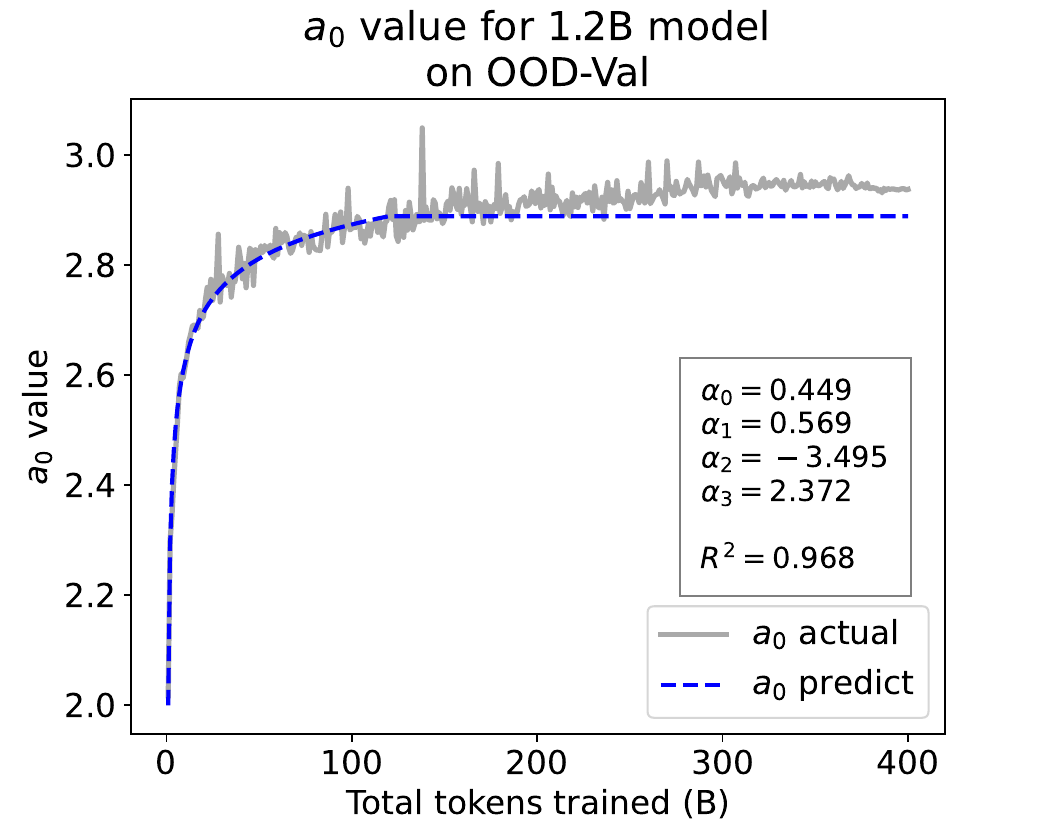}
  \end{subfigure} & 
  \begin{subfigure}{0.3\linewidth}
    \includegraphics[width=1.0\linewidth, trim={0cm 0cm 1.3cm 0cm}, clip]{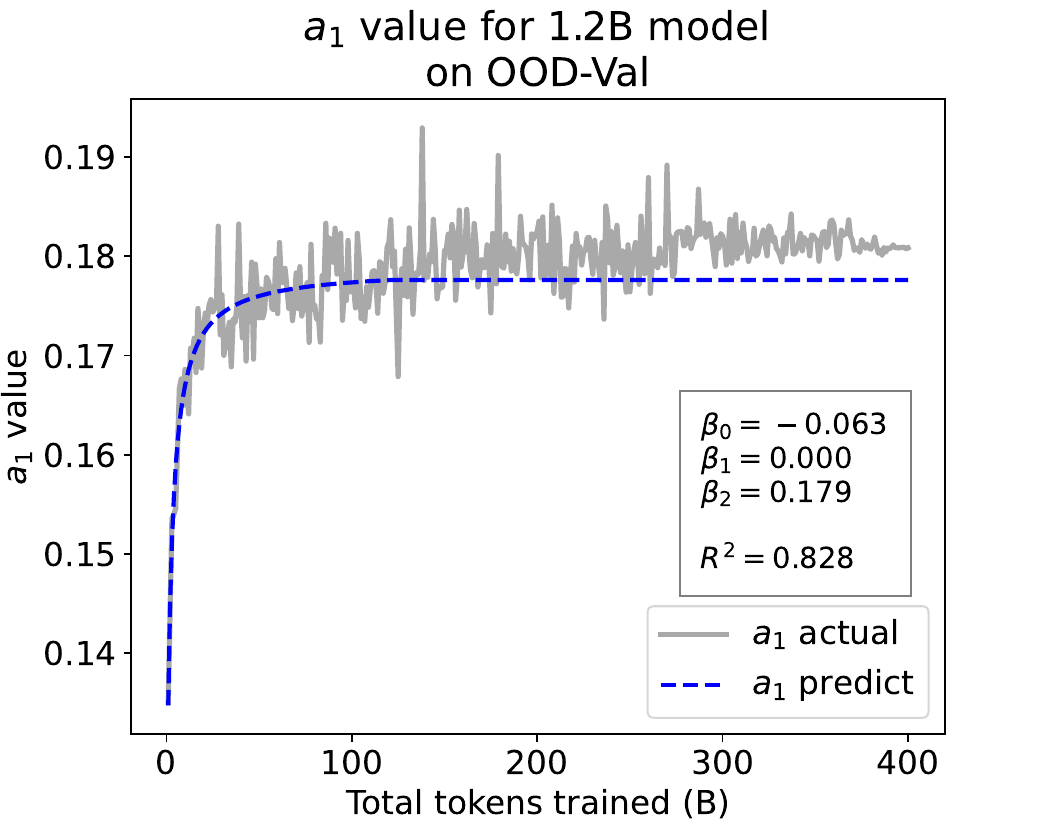}
  \end{subfigure} & 
  \begin{subfigure}{0.3\linewidth}
    \includegraphics[width=1.0\linewidth, trim={0cm 0cm 1.3cm 0cm}, clip]{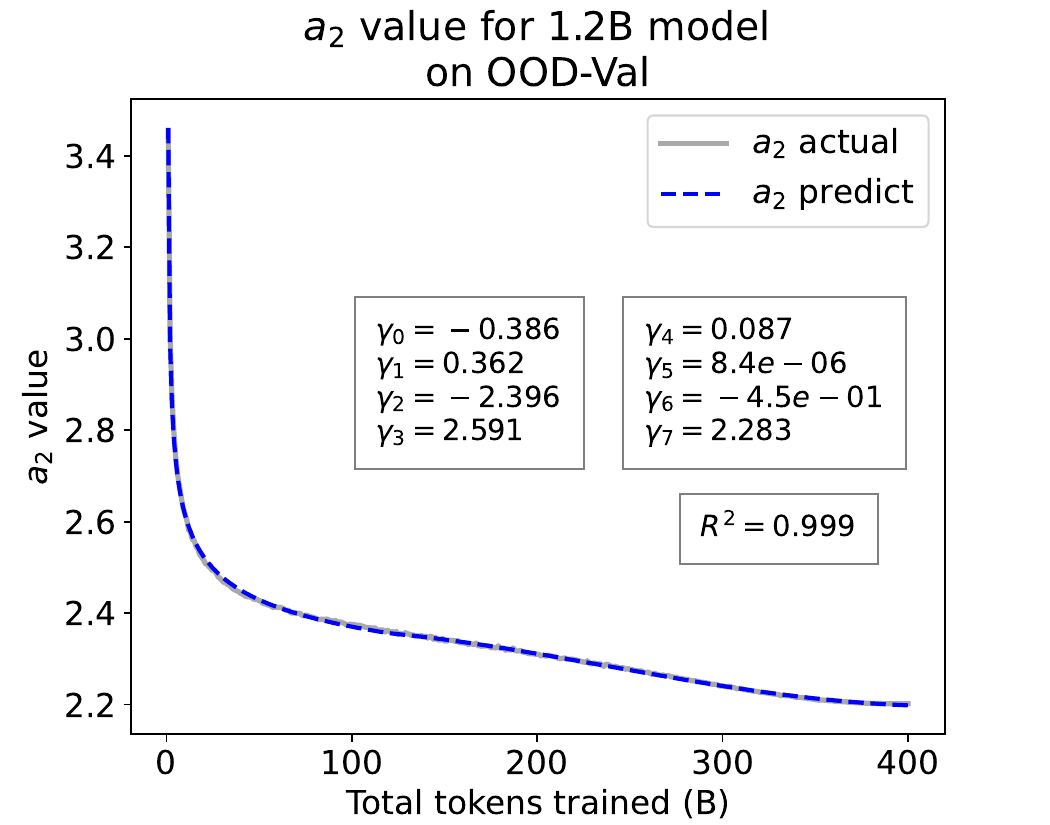}
  \end{subfigure} \\
  \end{tabular}
  \caption{More temporal scaling law fitting results for the larger 468M and the 1.2B models on both the ID-Val and OOD-Val.}
  \label{fig:fit_temporal_larger}
\end{figure*}

\begin{figure*}
  \centering
  \begin{tabular}{ccc}
    
  \begin{subfigure}{0.3\linewidth}
    \includegraphics[width=1.0\linewidth, trim={0cm 0cm 1.3cm 0cm}, clip]{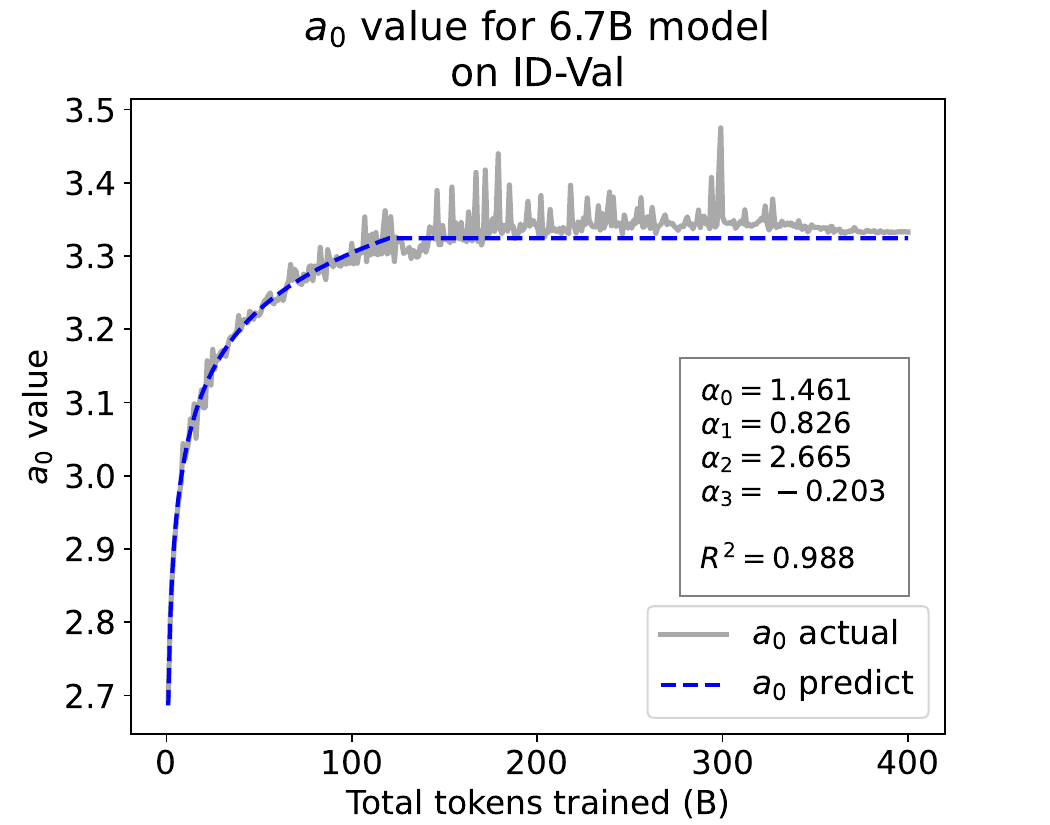}
    \vspace{5pt}
  \end{subfigure} & 
  \begin{subfigure}{0.3\linewidth}
    \includegraphics[width=1.0\linewidth, trim={0cm 0cm 1.3cm 0cm}, clip]{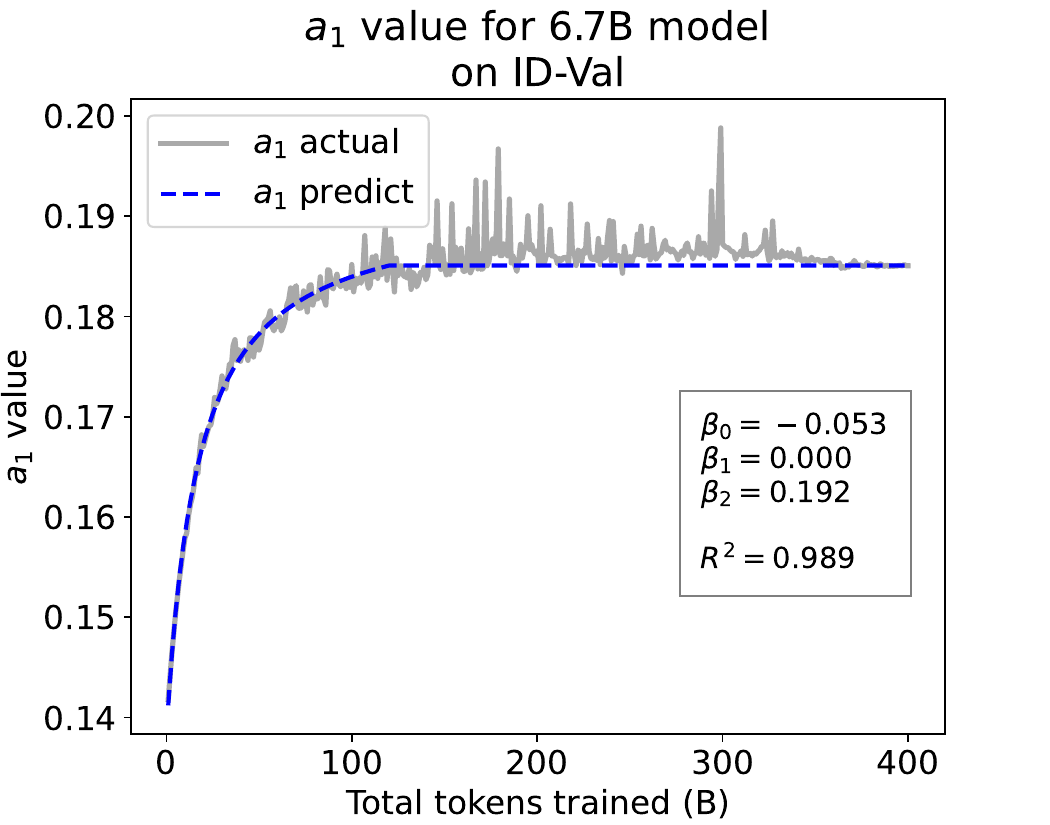}
    \vspace{5pt}
  \end{subfigure} & 
  \begin{subfigure}{0.3\linewidth}
    \includegraphics[width=1.0\linewidth, trim={0cm 0cm 1.3cm 0cm}, clip]{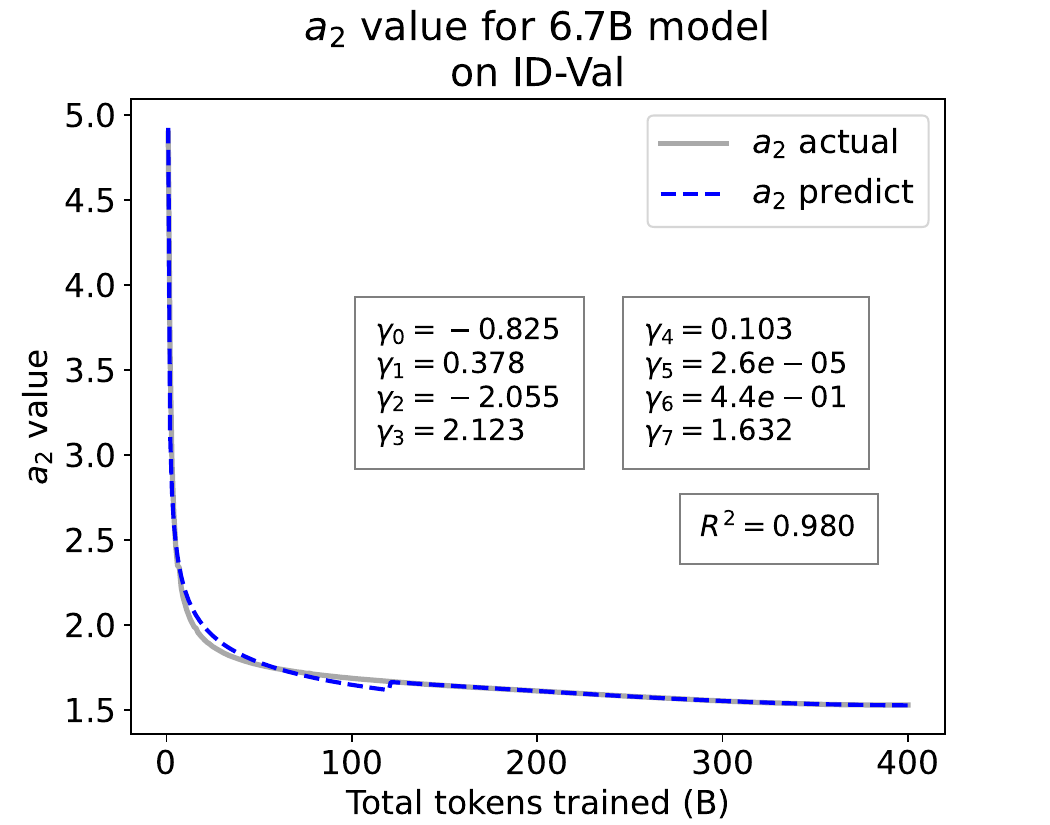}
    \vspace{5pt}
  \end{subfigure} \\

  \begin{subfigure}{0.3\linewidth}
    \includegraphics[width=1.0\linewidth, trim={0cm 0cm 1.3cm 0cm}, clip]{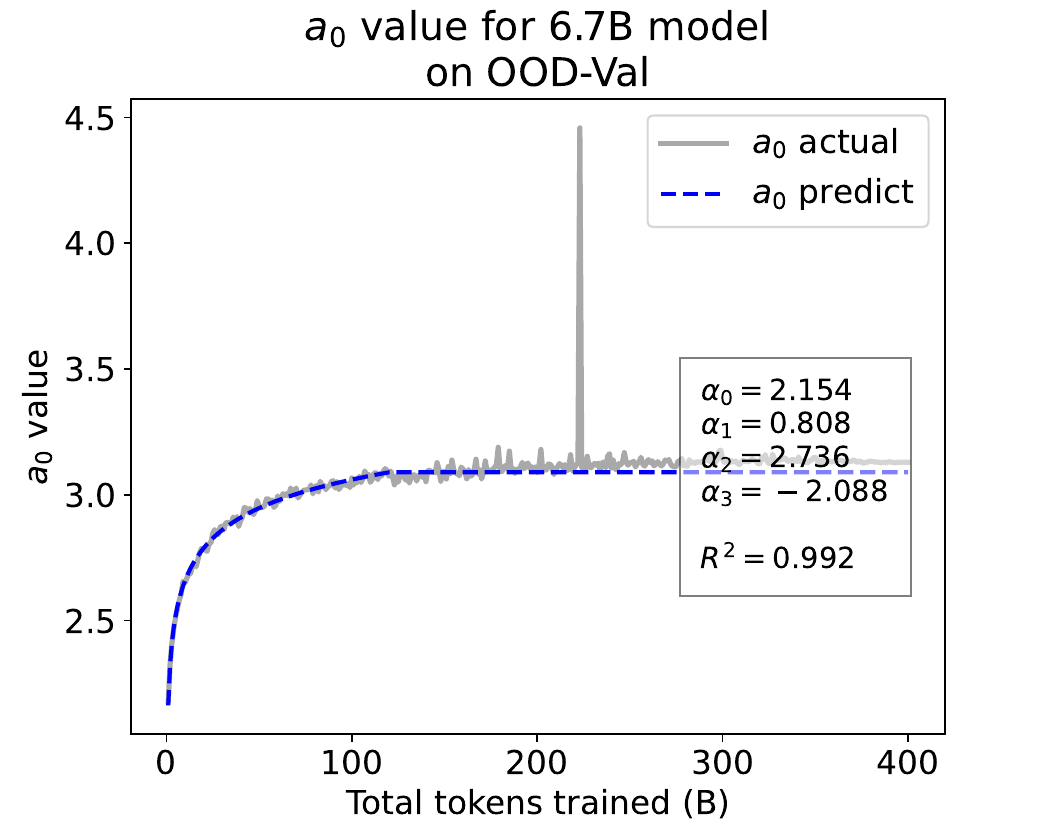}
  \end{subfigure} & 
  \begin{subfigure}{0.3\linewidth}
    \includegraphics[width=1.0\linewidth, trim={0cm 0cm 1.3cm 0cm}, clip]{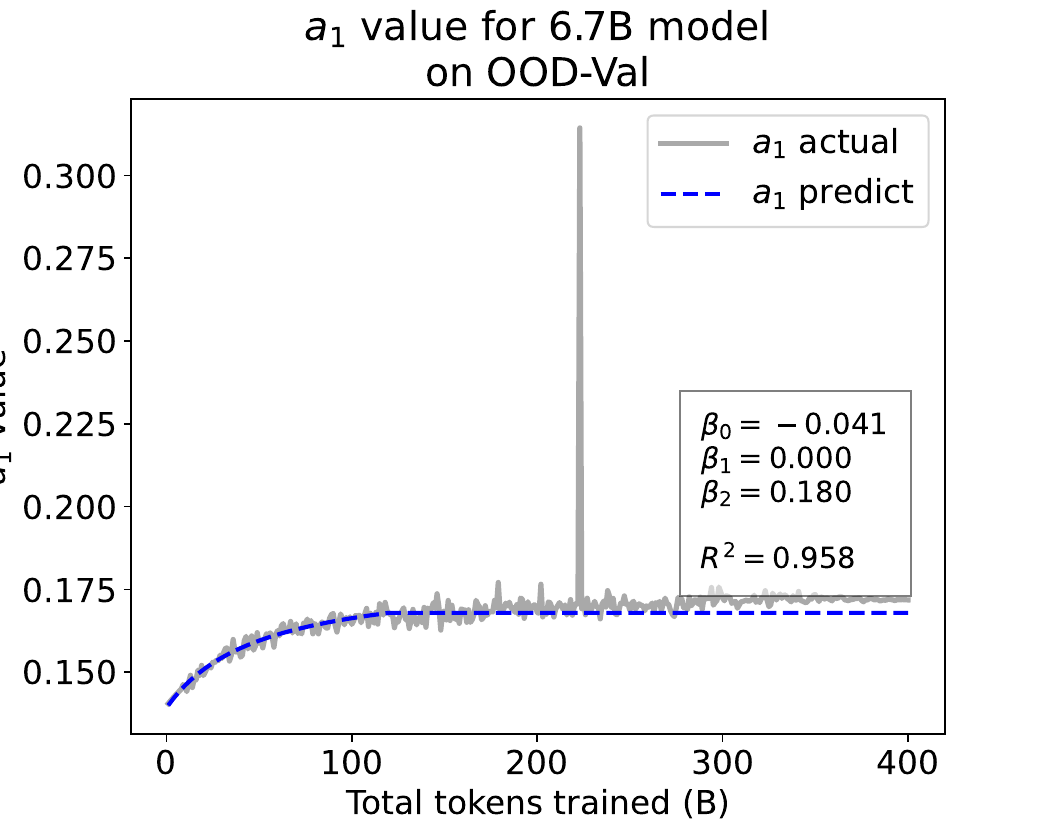}
  \end{subfigure} & 
  \begin{subfigure}{0.3\linewidth}
    \includegraphics[width=1.0\linewidth, trim={0cm 0cm 1.3cm 0cm}, clip]{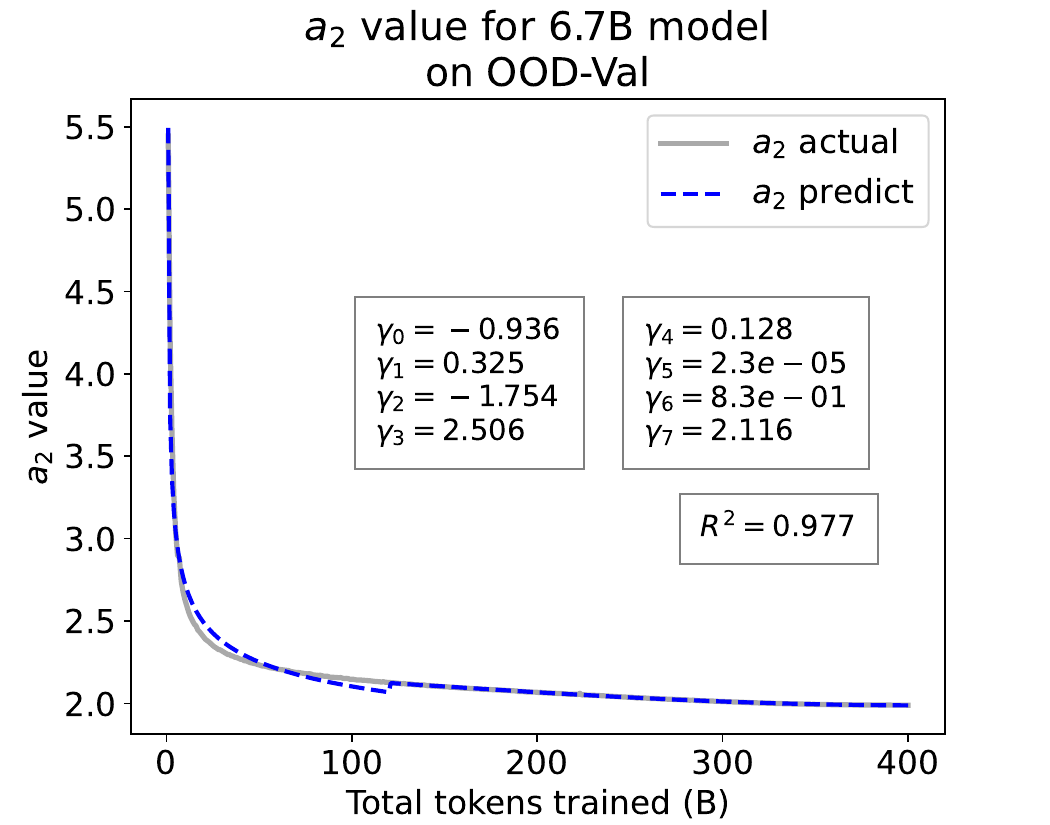}
  \end{subfigure} \\
  \end{tabular}
  \caption{More temporal scaling law fitting results for the 6.7B model on both the ID-Val and OOD-Val.}
  \label{fig:fit_temporal_7B}
\end{figure*}

As a supplement to \cref{fig:temporal_law}, we provide the results on the OOD-Val in \cref{fig:fit_temporal_58m_ood}.
We also provide fitting results on both the ID-Val and the OOD-Val for the larger 468M, 1.2B, and 6.7B model scales in \cref{fig:fit_temporal_larger} and \cref{fig:fit_temporal_7B}. 
Across different settings on model scales and validation set distributions, our temporal scaling law is able to depict the general trend of parameter evolution and reduce the influence of fluctuations as much as possible.


\begin{figure*}
  \centering
  \begin{tabular}{ccc}

  \begin{subfigure}{0.3\linewidth}
    \includegraphics[width=1.0\linewidth, trim={0cm 0cm 1.3cm 0cm}, clip]{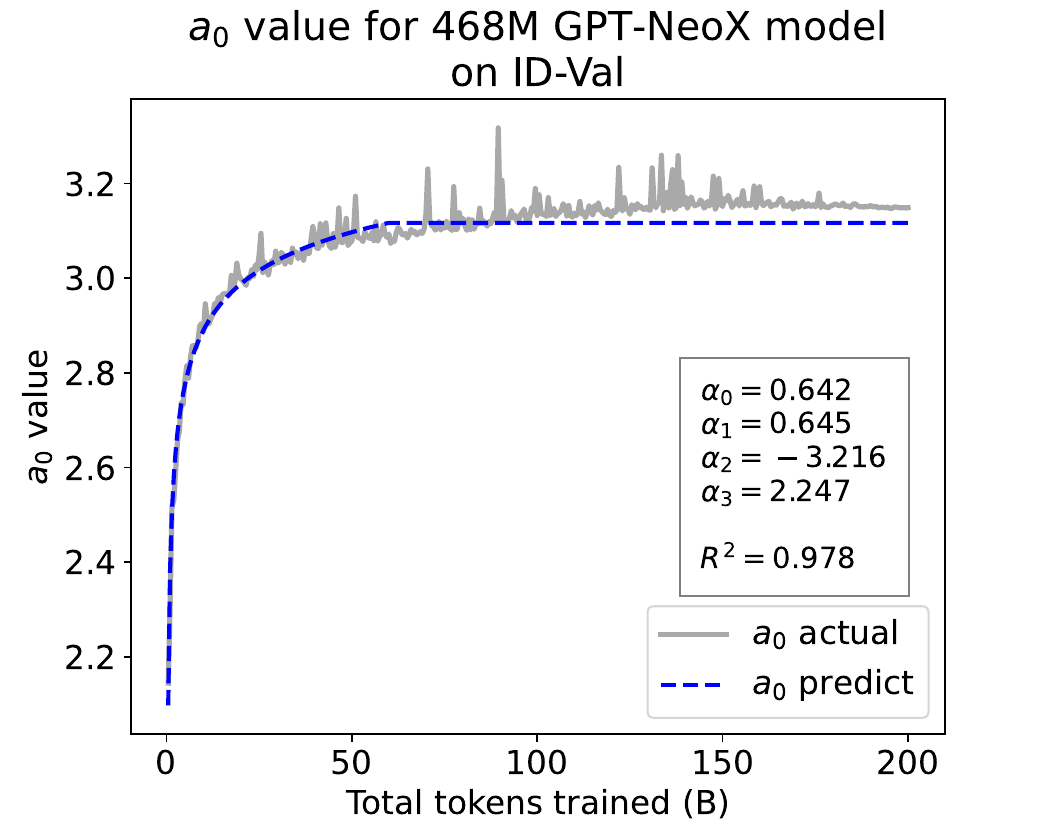}
  \end{subfigure} & 
  \begin{subfigure}{0.3\linewidth}
    \includegraphics[width=1.0\linewidth, trim={0cm 0cm 1.3cm 0cm}, clip]{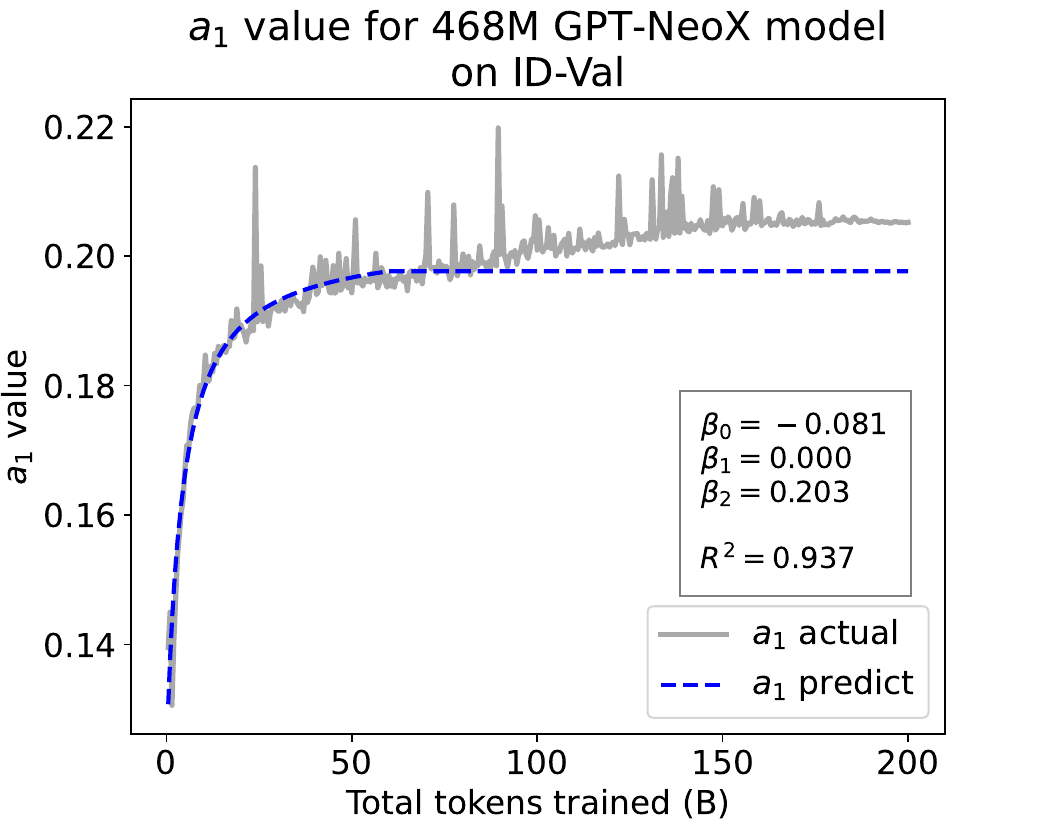}
  \end{subfigure} & 
  \begin{subfigure}{0.3\linewidth}
    \includegraphics[width=1.0\linewidth, trim={0cm 0cm 1.3cm 0cm}, clip]{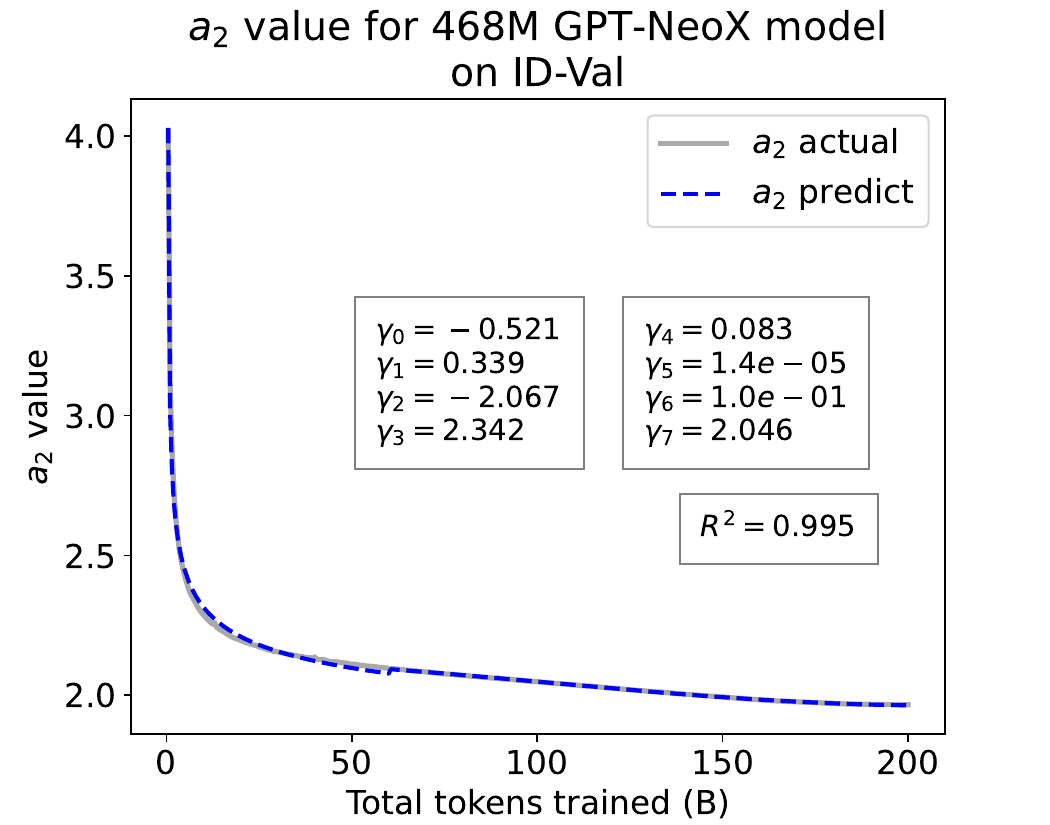}
  \end{subfigure} \\

  \begin{subfigure}{0.3\linewidth}
    \includegraphics[width=1.0\linewidth, trim={0cm 0cm 1.3cm 0cm}, clip]{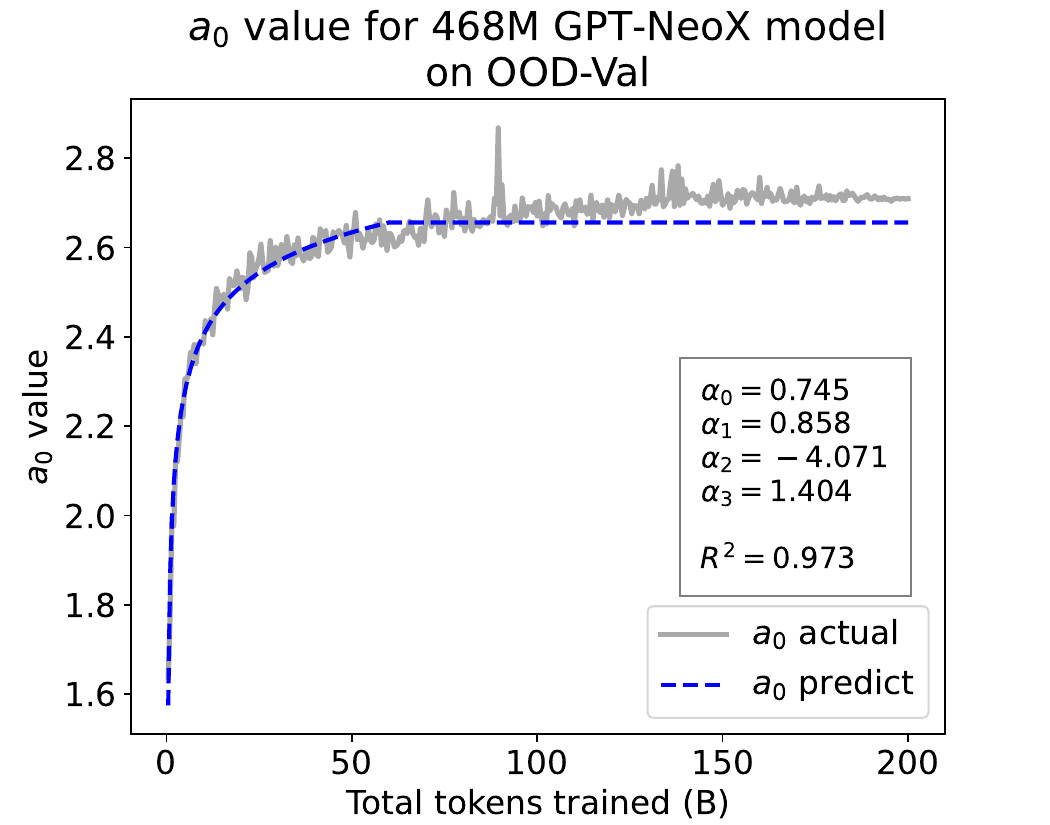}
  \end{subfigure} & 
  \begin{subfigure}{0.3\linewidth}
    \includegraphics[width=1.0\linewidth, trim={0cm 0cm 1.3cm 0cm}, clip]{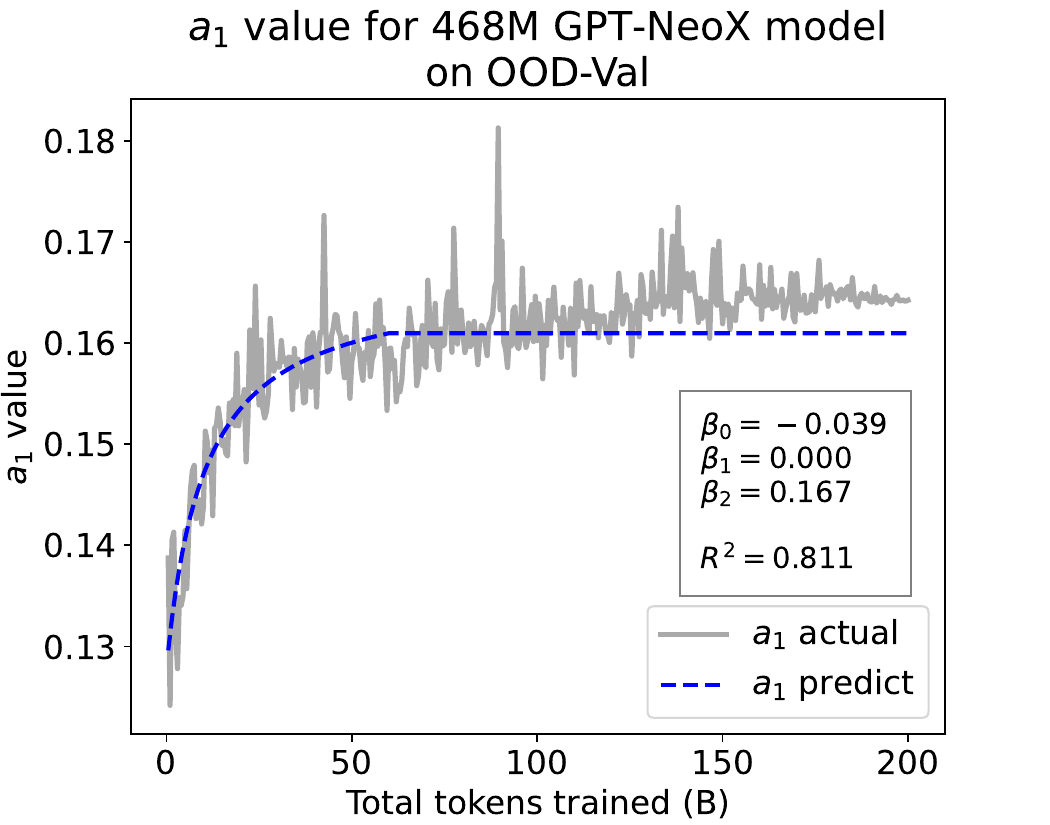}
  \end{subfigure} & 
  \begin{subfigure}{0.3\linewidth}
    \includegraphics[width=1.0\linewidth, trim={0cm 0cm 1.3cm 0cm}, clip]{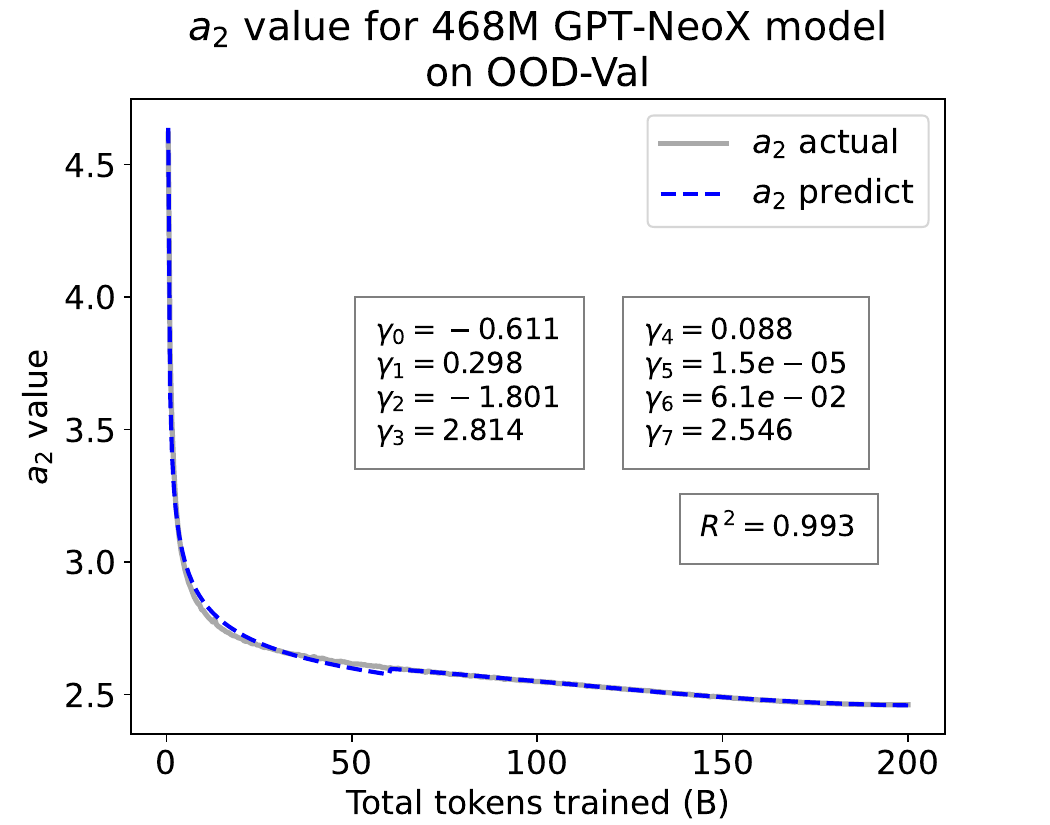}
  \end{subfigure} \\
    
  \begin{subfigure}{0.3\linewidth}
    \includegraphics[width=1.0\linewidth, trim={0cm 0cm 1.3cm 0cm}, clip]{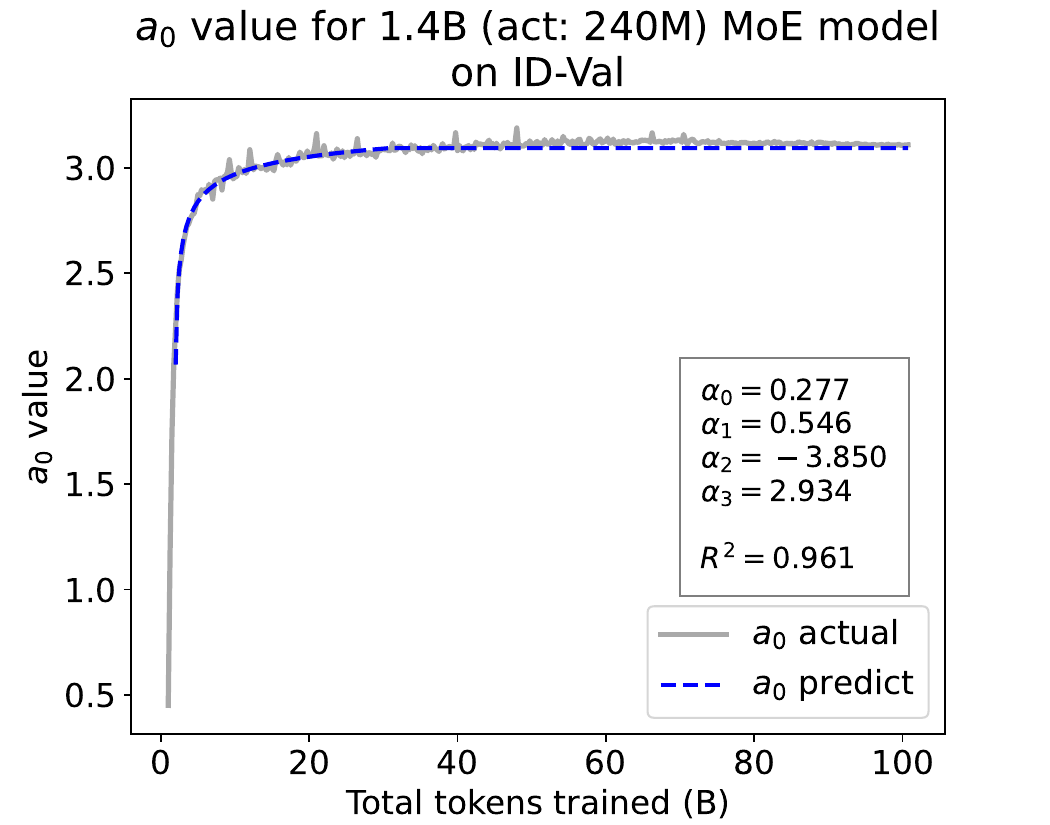}
    \vspace{5pt}
  \end{subfigure} & 
  \begin{subfigure}{0.3\linewidth}
    \includegraphics[width=1.0\linewidth, trim={0cm 0cm 1.3cm 0cm}, clip]{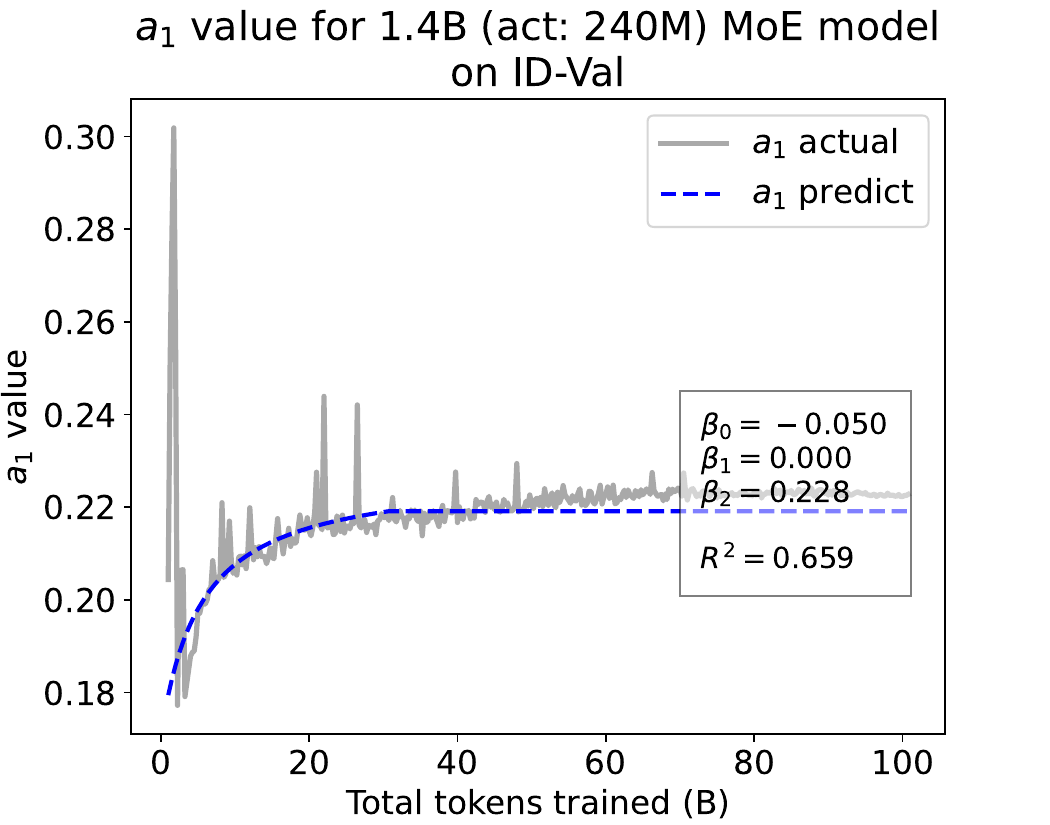}
    \vspace{5pt}
  \end{subfigure} & 
  \begin{subfigure}{0.3\linewidth}
    \includegraphics[width=1.0\linewidth, trim={0cm 0cm 1.3cm 0cm}, clip]{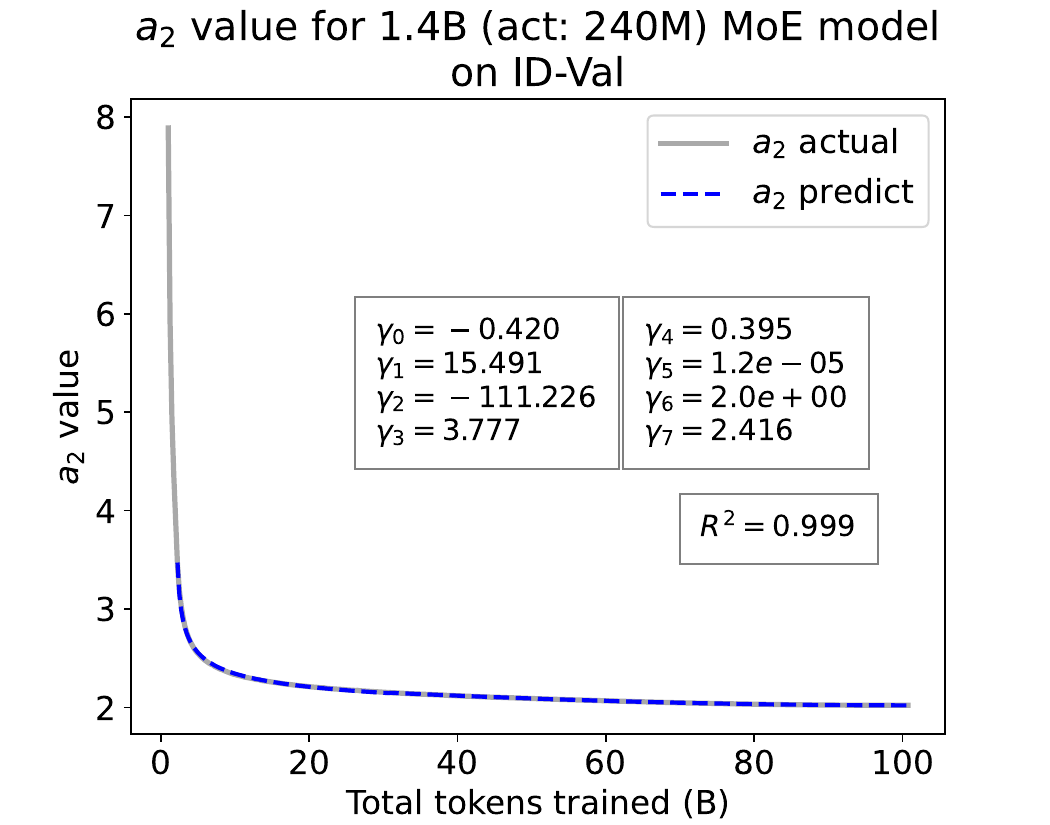}
    \vspace{5pt}
  \end{subfigure} \\

  \begin{subfigure}{0.3\linewidth}
    \includegraphics[width=1.0\linewidth, trim={0cm 0cm 1.3cm 0cm}, clip]{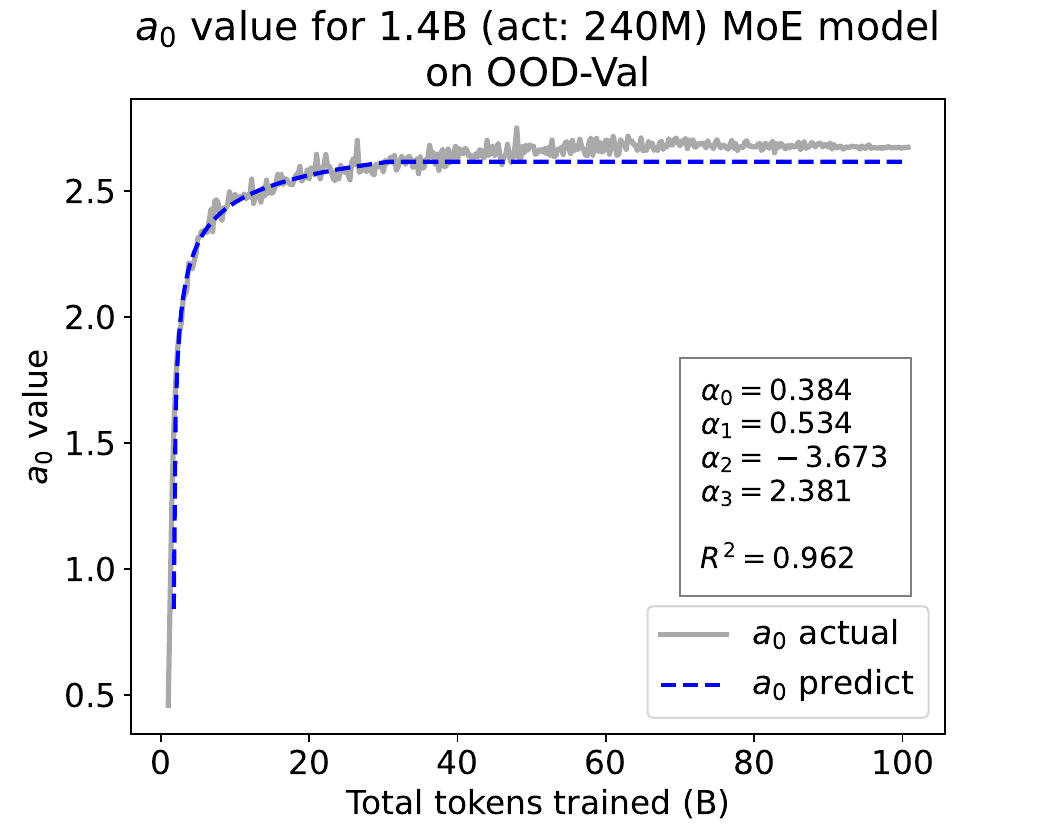}
  \end{subfigure} & 
  \begin{subfigure}{0.3\linewidth}
    \includegraphics[width=1.0\linewidth, trim={0cm 0cm 1.3cm 0cm}, clip]{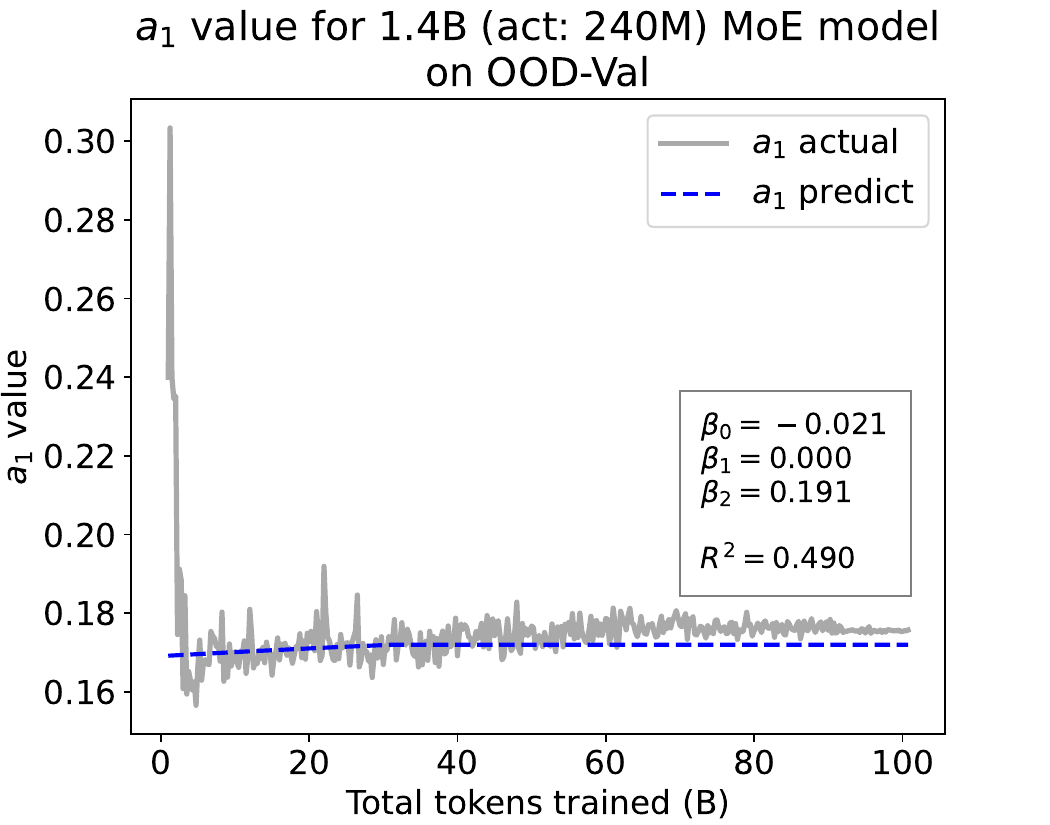}
  \end{subfigure} & 
  \begin{subfigure}{0.3\linewidth}
    \includegraphics[width=1.0\linewidth, trim={0cm 0cm 1.3cm 0cm}, clip]{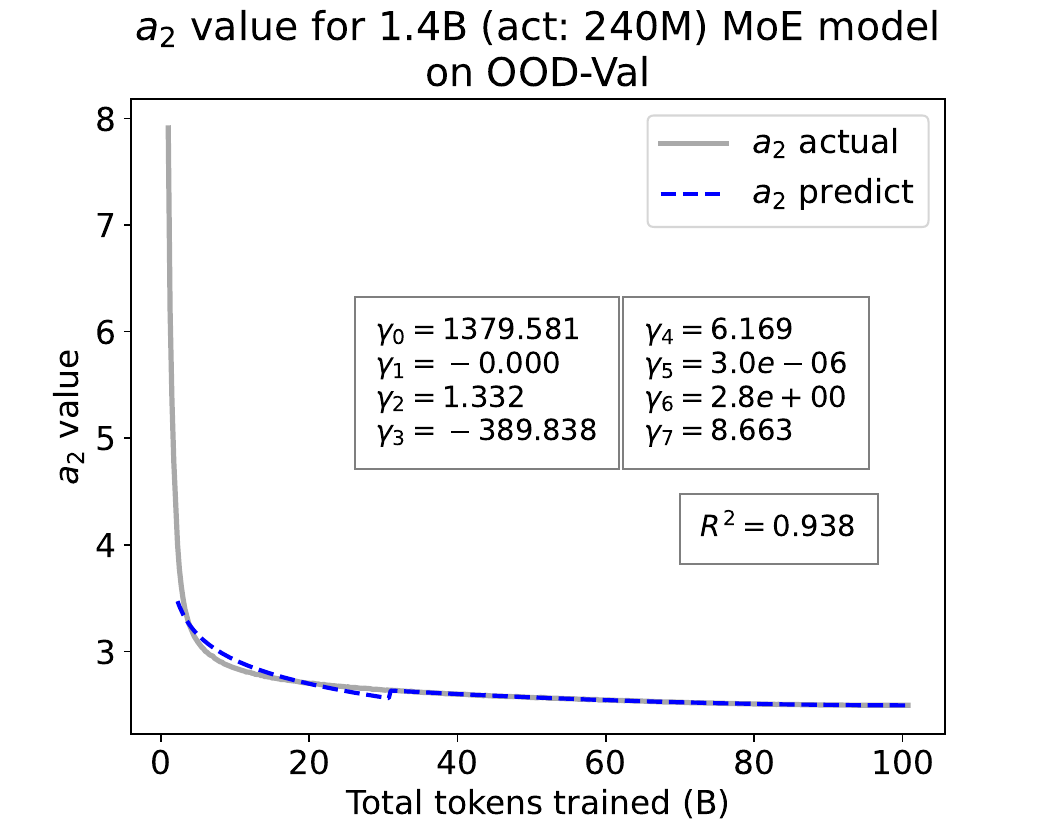}
  \end{subfigure} \\
  \end{tabular}
  \caption{More temporal scaling law fitting results for the 468M GPT-NeoX model and the 1.4B MoE model on both the ID-Val and OOD-Val. \textit{Note that the fitting for $a_1$ on the MoE model has smaller $R^2$ values than the fit for other parameters. This is due to the outliers in the beginning of the graph, as we applied a smaller number of total training tokens for the model, and the training was not stablized in the beginning. Nevertheless, with the outlier filting mechanism described in \cref{sec:3_3_temporal_law} and \cref{alg:predict_pipeline}, the temporal scaling law successfully portraits the overall trend of $a_1$. }}
  \label{fig:fit_temporal_generalize}
\end{figure*}

\begin{figure*}
  \centering
  \begin{tabular}{ccc}

  \begin{subfigure}{0.31\linewidth}
    \includegraphics[width=1.0\linewidth, trim={0cm 0cm 1.3cm 0cm}, clip]{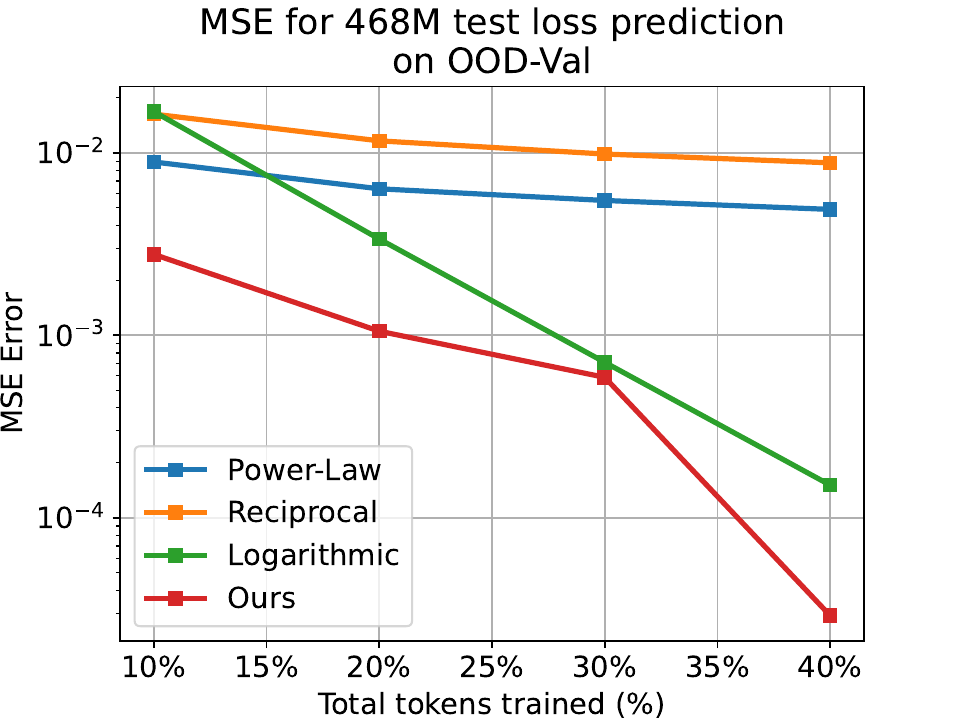}
  \end{subfigure} &
  \begin{subfigure}{0.31\linewidth}
    \includegraphics[width=1.0\linewidth, trim={0cm 0cm 1.3cm 0cm}, clip]{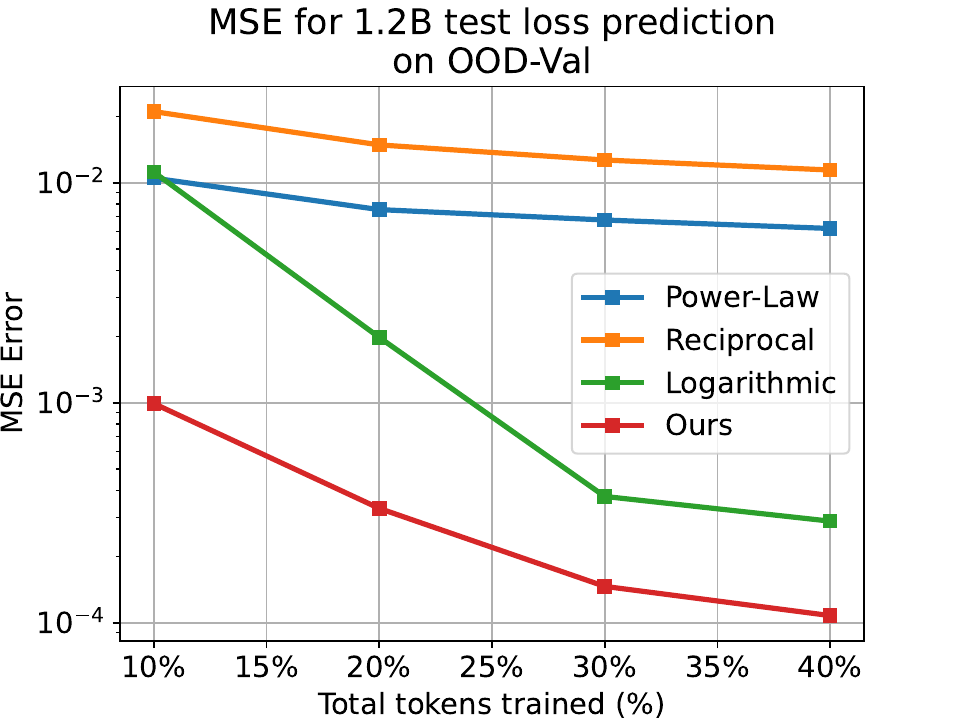}
  \end{subfigure} &
  \begin{subfigure}{0.31\linewidth}
    \includegraphics[width=1.0\linewidth, trim={0cm 0cm 1.3cm 0cm}, clip]{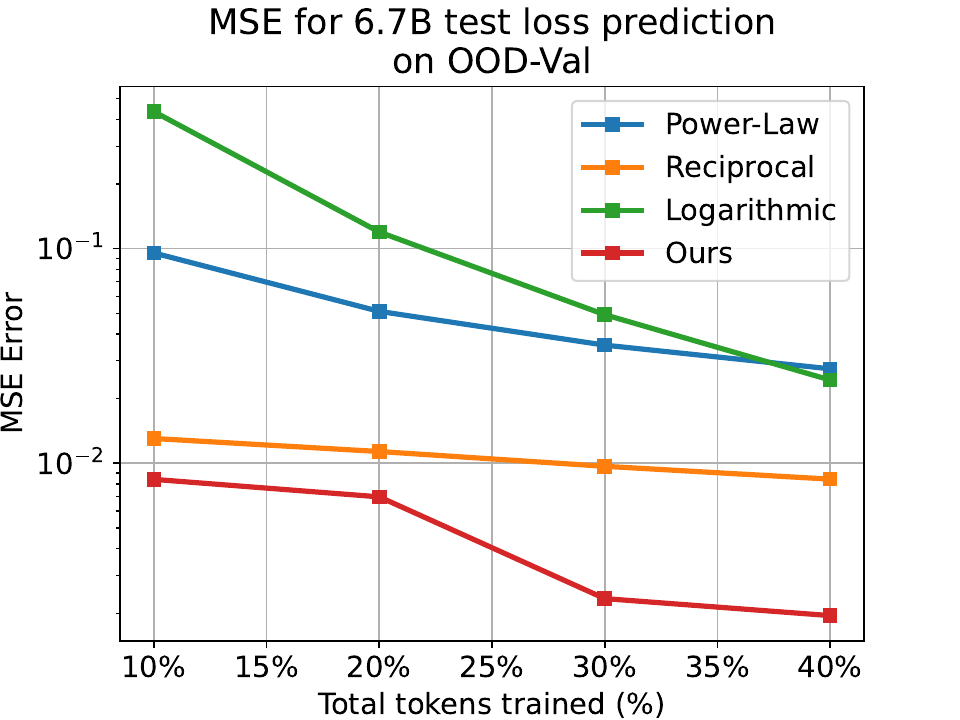}
  \end{subfigure} \\

  \end{tabular}
  \caption{MSE results for predicting the subsequent test loss via the temporal scaling law after completing different proportions of the training process. Note that the y-axis representing the MSE error is in \underline{\textbf{log scale}}.}
  \label{fig:fit_exp_result_ood}
\end{figure*}

\begin{table}[!t]
\centering
\scalebox{0.7}{\begin{tabular}
{l|l}
\toprule
 Strategy Name & Description \\
\midrule
Default Practice & Average loss on all token positions. \\
Head Suppression & Weight loss on the foremost 10\% tokens by 0.5x. \\
Body Suppression & Weight loss on the central 80\% tokens by 0.5x. \\
Tail Suppression & Weight loss on the last 10\% tokens by 0.5x. \\
\bottomrule
\end{tabular}}
\caption{Details for different weighting strategies.}
\label{tab:weight_strategies}
\end{table}

\begin{table}[!t]
\centering
\scalebox{0.7}{\begin{tabular}
{l|cccc}
\toprule
 & Default Practice & Head & Body & Tail \\
\midrule
\textit{468M} & \textbf{8.62} & \textbf{8.62} & 8.66 & 8.63 \\
\textit{1.2B} & \textbf{7.52} & \textbf{7.52} & 7.56 & 7.53 \\
\bottomrule
\end{tabular}}
\caption{ID-Val perplexity of models pretrained on different weighting strategies. ``Head'' represents ``Head suppression'', etc. \textbf{Bold} represents the best result.}
\label{tab:weight_token_res_mini}
\end{table}


\begin{table*}[!t]
\centering
\scalebox{0.68}{
\begin{tabular}
{ll|p{1.1cm}<{\centering}p{1.65cm}<{\centering}p{1.65cm}<{\centering}p{1.65cm}<{\centering}p{1.65cm}<{\centering}p{1.65cm}<{\centering}p{1.7cm}<{\centering}}
\toprule
 & & BoolQ & HellaSwag & OpenBookQA & PIQA & SIQA & StoryCloze & Winogrande \\
 \midrule
\multirow{4}{*}{\textit{468M}} & Default Practice & \underline{52.28} & \underline{45.39} & \textbf{31.87} & \underline{68.64} & \underline{44.04} & 64.24 & 55.33 \\
& Head Suppression & 52.05 & \textbf{45.60} & \underline{31.80} & 68.50 & \textbf{44.13} & \textbf{64.71} & 55.38 \\
& Body Suppression & 51.98 & 45.12 & 31.07 & \textbf{68.66} & 43.76 & 64.41 & \textbf{55.64} \\
& Tail Suppression & \textbf{52.91} & 45.28 & 31.67 & 68.05 & 44.03 & \underline{64.64} & \underline{55.62} \\
  \midrule
\multirow{4}{*}{\textit{1.2B}} & Default Practice & \underline{61.42} & \underline{54.07} & 34.20 & \underline{72.00} & \textbf{45.87} & \textbf{68.63} & \textbf{58.64} \\
& Head Suppression & 59.96 & \textbf{54.49} & 34.33 & 71.75 & 45.36 & 68.39 & \underline{58.30} \\
& Body Suppression & 60.57 & 53.75 & \underline{34.80} & \textbf{72.05} & 44.75 & 68.20 & 57.72 \\
& Tail Suppression & \textbf{61.57} & 54.02 & \textbf{35.53} & 71.49 & \underline{45.63} & \underline{68.61} & 58.27 \\
\bottomrule
\end{tabular}
}
\caption{Average performance (0/1/5-shot) on common sense reasoning benchmarks of models pre-trained under different weighting strategies. \textbf{Bold} and \underline{underline} represent the best and the second-best averaged results, correspondingly.}
\label{tab:weight_token_res}
\end{table*}

\section{Generalization to More Model Structures}
\label{sec:append_structure}
To demonstrate the generalizability of the Temporal Scaling Law to other model structures, we pre-trained a 468M model based on the \textit{GPTNeoXForCausalLM} \cite{GPT-NeoX} architecture and a 1.4B MoE model with 240M activated parameters based on the LLaMA architecture. 

\textbf{Experimental Setup.}
The 468M \textit{GPTNeoXForCausalLM} model shares the same hyperparameters with the 468M LLaMA-based model listed in \cref{tab:model_hyperparameters}. The 1.4B MoE model is built based on the LLaMA architecture with 16 total experts and adopts a Top-2 activation strategy, following the common setting in various MoE research \cite{su2024maskmoe,su2024cartesianmoe,topprouting}. 
Both models are pre-trained on the Pile dataset used in the main article. Specifically, the 468M \textit{GPTNeoXForCausalLM} model is pre-trained with 200B tokens, and the 1.4B MoE model is pre-trained with 100B tokens. Detailed settings are listed in \cref{tab:model_hyperparameters_generalize}.

\textbf{Experiment Results.}
We present the fitting results on both ID-Val and OOD-Val of the two pre-trained models in \cref{fig:fit_temporal_generalize}. The proposed temporal scaling law successfully genralizes to those architectures, capturing the primary patterns of parameter evolution for the GPT-NeoX and the MoE models, well demonstrating its generalizability.

\section{Generalization to More Learning Schedules}

In this article, we mainly focus on the cosine decay scheduler, which is adopted by most mainstream LLM structures \cite{DBLP:journals/corr/abs-2302-13971,DBLP:journals/corr/abs-2307-09288,DBLP:journals/corr/abs-2309-16609}. 
To provide insights for other schedulers, we have managed to conduct a preliminary experiment with the 468M model using a simple linear learning rate scheduler. Our results indicate that for this setup, replacing the fitting function for the $a_2$ parameter with $a_2^N=\gamma_4\cdot N+\gamma_5\ (N>N_{sep})$ yields a high overall fit with $R^2=0.9947$. This demonstrates a strong correlation between the temporal evolution of $a_2$ and the choice of learning rate scheduler, which is also observed in recent works \cite{luo2025multi}. 

\section{Test Loss Prediction Results on OOD-Val}
\label{sec:pred_ood}
In \cref{fig:fit_exp_result}, we presented the test loss prediction results on the ID-Val for the scaled-up larger models. Additionally, we present the prediction results on the OOD-Val in \cref{fig:fit_exp_result_ood}. Same with results on the ID-Val, our temporal scaling law consistently generates reliable results over the comparing baselines.

\section{More Details for Use Case \#1: Hyperparameter Selection}
\label{sec:more_case_1}
Due to the page limit, we omitted some details for the use case \#1 in \cref{sec:4_hyperparameter}. They are thoroughly described below for reproducibility.

\textbf{58M Model Architecture.}
The architecture of the 58M model is identical to the 70M parameter model outlined in \cite{DBLP:conf/icml/BidermanSABOHKP23}. The differences between model parameters can be ascribed to different vocabulary sizes and different activation functions (i.e., SwiGLU v.s. GeLU).

\textbf{Proportion Candidate Generation.}
In the retrieval stage, we generate a group of proportion candidates by conducting grid search on domain weights in the Pile dataset \cite{DBLP:journals/corr/abs-2101-00027}.
Specifically, we first calculate the original domain weight according to the original epoch settings noted by \cite{DBLP:journals/corr/abs-2101-00027} on our applied 32k tokenizer. 
Note that \textit{this is also the domain weight applied in \cref{sec:3_2_exp_setup}}. For the grid search, we adopt a simple but popular pipeline by selecting a domain, modifying its domain weight by a random value $s\in [-0.05,0.05]$, and normalizing the domain weight in other domains. 

\textbf{Candidate Selection on the 58M Model.}
In the retrieval stage, we choose the Top-5 data mixture proportions on the 58M model for the rerank stage and locate the Top-1 proportion on the 58M model for final comparisons. We choose the Top-5 and Top-1 proportions by calculating the average benchmark performance of the corresponding 58M model under 0-shot, 1-shot, and 5-shot settings.

\section{More Details for Use Case \#2: Revisit the Training Strategy}
\label{sec:more_case_2}
In \cref{sec:5_token_position_loss}, we have stressed that a fundamental bias in learning difficulty based on token position exists. Specifically, head tokens (with shorter context) are generally harder to predict than tail tokens (with longer context), due to the increased uncertainty caused by limited contextual information available for earlier tokens in a sequence. Surprisingly, our temporal scaling law suggests that LLMs learn equally on different token positions after an early training period (as shown by \cref{eq:learning_bias}), despite the learning difficulty bias.

\begin{figure*}
  \centering
  \begin{tabular}{cc}
  
  \begin{subfigure}{0.46\linewidth}
    \includegraphics[width=1.0\linewidth, trim={0cm 0cm 1.3cm 0cm}, clip]{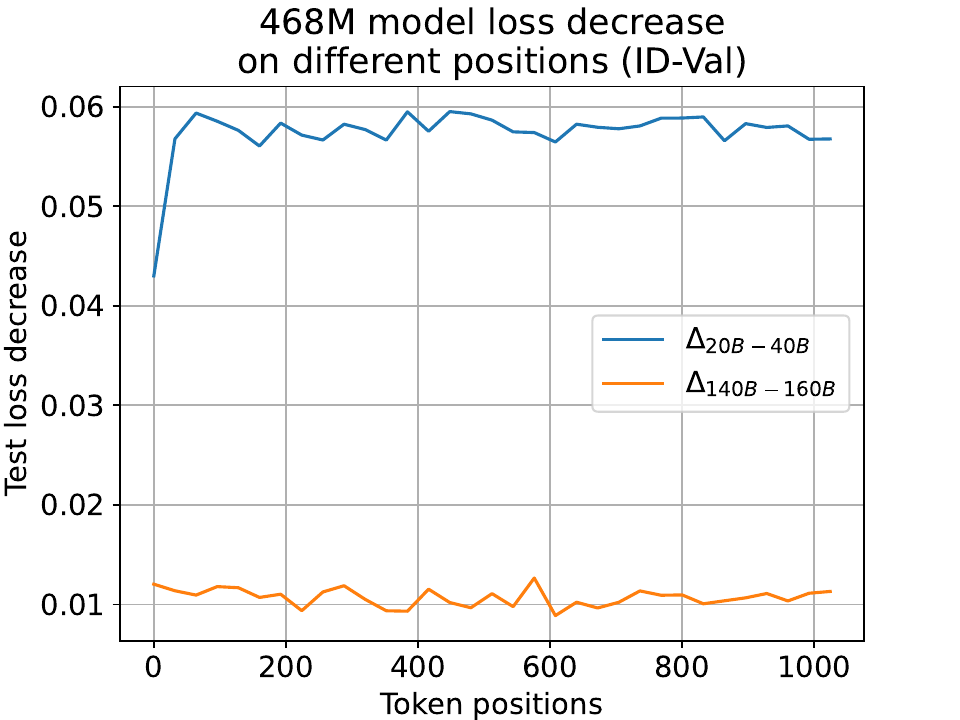}
  \end{subfigure} & 
  \begin{subfigure}{0.46\linewidth}
    \includegraphics[width=1.0\linewidth, trim={0cm 0cm 1.3cm 0cm}, clip]{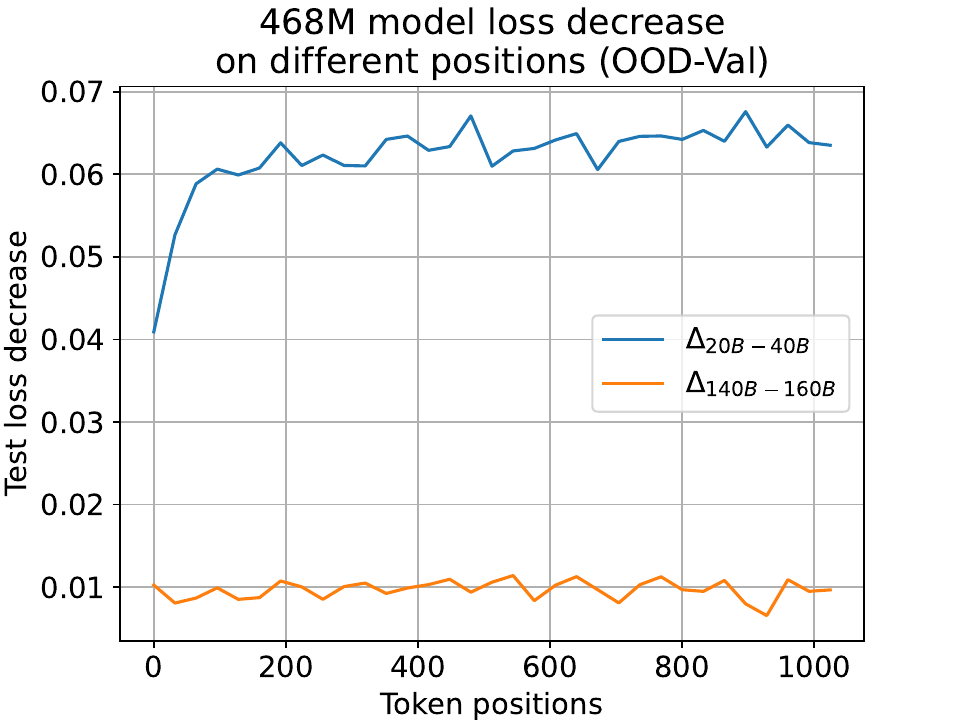}
  \end{subfigure} \\

  \begin{subfigure}{0.46\linewidth}
    \includegraphics[width=1.0\linewidth, trim={0cm 0cm 1.3cm 0cm}, clip]{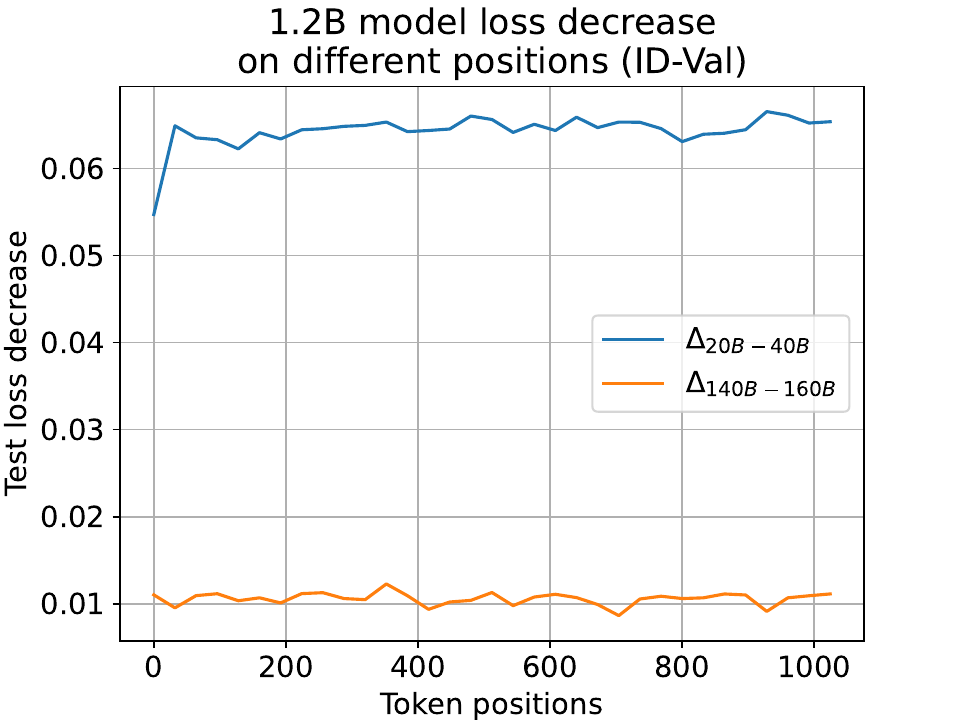}
  \end{subfigure} & 
  \begin{subfigure}{0.46\linewidth}
    \includegraphics[width=1.0\linewidth, trim={0cm 0cm 1.3cm 0cm}, clip]{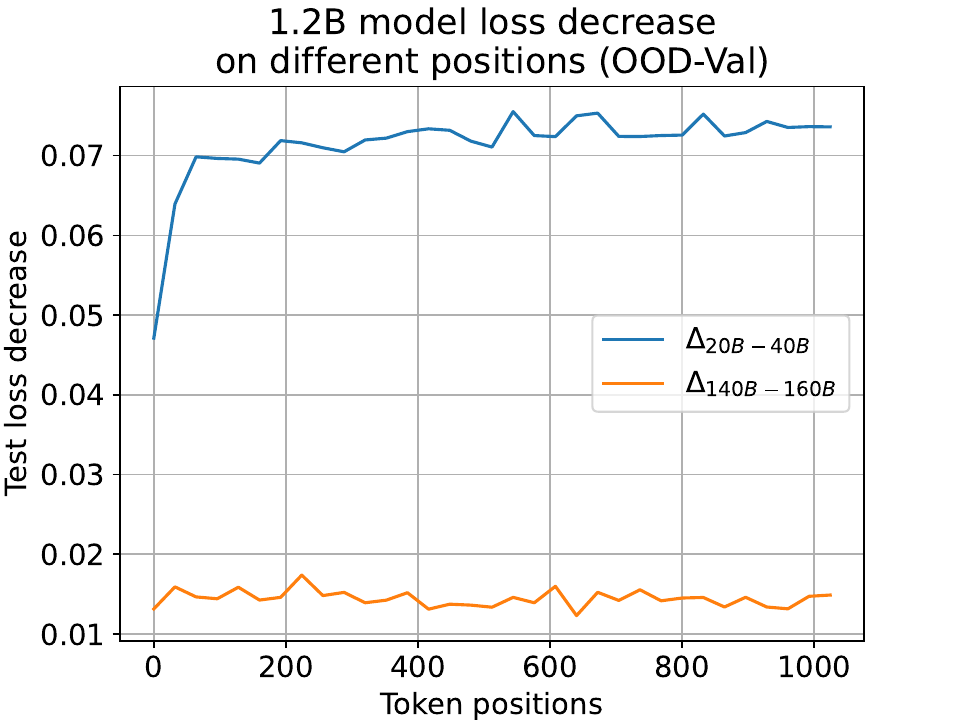}
  \end{subfigure} \\
  
  \end{tabular}
  \caption{Test loss decrease on different token positions in the given training period for the 468M and the 1.2B models. ``$\Delta_{20B-40B}$'' means the decrease of loss at each token position from ``being trained with 20B tokens'' to ``being trained with 40B tokens'' etc. After an early training period, the loss decrease tends to be uniform across all token positions (i.e., $\Delta_{140B-160B}$).}
  \label{fig:token_loss_diff}
\end{figure*}

\textbf{Observation for Actual Loss Decrease on Positions.} 
We use the larger 468M and the 1.2B models to validate our suggestion.
To validate this observation, we plot the actual loss decrease pattern observed during training along different token positions in~\cref{fig:token_loss_diff}. As shown in~\cref{fig:token_loss_diff}, in the early training period of ``from 20B to 40B tokens'', the head tokens suffer from less loss decrease due to a higher learning difficulty. However, after the early training period, in a later period of ''from 140B to 160B tokens'' that the separation point conditions in \cref{eq:sep_point} are already fulfilled, test loss decrease among all token positions tends to be uniform across all settings.
Therefore, we can infer from this observation that the learning dynamics derived from our temporal scaling law are authentically presented in LLM pre-training.

\textbf{Insights for Weighting Strategies.} 
We hypothesize that the default strategy for training generative language models, in which losses on tokens in all positions are simply averaged, is an effective solution. 
To validate the hypothesis, we conduct LLM pretraining on the larger 468M and the 1.2B models with three position-based weighting strategies: Head suppression, Body suppression, and Tail suppression, applied only after the $N_{sep}$ point. 
The implementation details of these strategies are described in \cref{tab:weight_strategies}. 
Note that for the Suppression strategies, we normalize the weighted losses to ensure the average weight for each token position is 1.0, and thus make the corresponding average loss comparable with the default practice.
We apply the pre-training settings as in \cref{sec:3_2_exp_setup} for comparing different position-based weighting strategies. 
As shown in \cref{tab:weight_token_res_mini}, despite being weighted on different token positions, weighting different tokens by position during LLM pretraining probably cannot yield better results than the default practice, though different positions possess fundamental bias in learning difficulty.


To further validate that the default practice actually trains a competitive model compared to the weighting strategies, we test the model performance on the common sense reasoning benchmarks described in \cref{sec:4_hyperparameter}, and report average model performance on 0-shot, 1-shot, and 5-shot settings. As shown in \cref{tab:weight_token_res}, all position-based weighting strategies acquire comparable or slightly inferior average results to the default practice, in which no weighting strategies are attached. On individual tasks, the default practice even achieves top-2 accuracies among 11 of 14 settings, surpassing all weighting variants. This indicates that weighting different tokens by position during LLM pre-training probably cannot yield better results than the default training strategy, further demonstrating that it is unnecessary to re-weight tokens by their positions during LLM pre-training.

\end{document}